\documentclass[10pt,twocolumn,letterpaper]{article}

\usepackage{cvpr}
\usepackage{caption} 
\usepackage{times}
\usepackage{epsfig}
\usepackage{graphicx}
\usepackage{amsmath}
\usepackage{amssymb}
\usepackage[usenames,dvipsnames]{color}
\usepackage{flushend}

\usepackage{multirow}
\usepackage{booktabs}
\usepackage{lib}
\usepackage{tabularx}
\usepackage{colortbl}
\usepackage{epstopdf}

\newcommand{\vertical}[1]{\rotatebox[origin=l]{90}{#1}}
\newcommand{\arbelaez}[0]{Arbel\'{a}ez\xspace}
\newcommand{\rgbd}[0]{RGB-D\xspace}
\newcommand{\rgb}[0]{RGB\xspace}

\newcommand{\nyu}[0]{NYUD2\xspace}

\newcommand{\regionAP}[0]{$AP^r$\xspace}
\newcommand{\modelAP}[0]{$AP^m$\xspace}
\newcommand{\boxAP}[0]{$AP^b$\xspace}

\newcommand{\cnn}[0]{CNN\xspace}

\newcommand{\hha}[0]{HHA\xspace}

\newcommand{\notextbf}[1]{#1}
\newcommand{\insertA}[2]{\IfFileExists{#2}{\includegraphics[width=#1\textwidth]{#2}}{\includegraphics[width=#1\textwidth]{figures/blank.jpg}}}
\newcommand{\insertB}[2]{\IfFileExists{#2}{\includegraphics[height=#1\textwidth]{#2}}{\includegraphics[height=#1\textwidth]{figures/blank.jpg}}}
\usepackage[pagebackref=true,breaklinks=true,letterpaper=true,colorlinks,bookmarks=false]{hyperref}

\cvprfinalcopy % *** Uncomment this line for the final submission

\newcolumntype{C}{>{\centering\arraybackslash}p{40ex}}

\ifcvprfinal\fi
\begin{document}

%%%%%%%%% TITLE
\title{Inferring 3D Object Pose in \rgbd Images}

\author{
% \begin{tabular}{C@{\hspace{8mm}}C@{\hspace{8mm}}}
\begin{tabular}{CC}
Saurabh Gupta & Pablo \arbelaez \\
UC Berkeley & Universidad de los Andes, Colombia \\
{\tt\small sgupta@eecs.berkeley.edu} & {\tt\small pa.arbelaez@uniandes.edu.co} \\ \\
Ross Girshick & Jitendra Malik\\
Microsoft Research & UC Berkeley\\
{\tt\small rbg@microsoft.com} & {\tt\small malik@eecs.berkeley.edu}
\end{tabular}
}

\maketitle
%\thispagestyle{empty}

%%%%%%%%% ABSTRACT
\begin{abstract} The goal of this work is to replace objects in an \rgbd scene
with corresponding 3D models from a library. We approach this problem by first
detecting and segmenting object instances in the scene using the approach from
Gupta \etal \cite{guptaECCV14}. We use a convolutional neural network (\cnn) to
predict the pose of the object. This \cnn is trained using pixel normals in
images containing rendered synthetic objects. When tested on real data, it
outperforms alternative algorithms trained on real data.  We then use this
coarse pose estimate along with the inferred pixel support to align a small
number of prototypical models to the data, and place the model that fits the
best into the scene. We observe a 48\% relative improvement in performance at
the task of 3D detection over the current state-of-the-art \cite{songECCV14},
while being an order of magnitude faster at the same time.
\end{abstract}

%%%%%%%%% BODY TEXT
\section{Introduction}
Consider \figref{teaser}. Understanding such an indoor image ultimately
requires replacing all the objects present in the scene by three dimensional models.
Traditionally, computer vision researchers have studied the problems of object
detection, semantic and instance segmentation, fine-grained categorization and
pose estimation.  However, none of those outputs by itself is enough for, \eg, a
robot to interact with this cluttered environment. This work strives to achieve such a
level of scene understanding in \rgbd images.

%This work encompasses all of these sub-problems in computer vision and strives
%to achieve an understanding of the image rich enough to allow a robot to
%interact with the world. \todo{Pablo, is this too bold to start with?}

The output of our proposed system is visualized in \figref{teaser}. Our 
approach is able to successfully retrieve relevant models and align them with
the data. Such an output does not only address traditional problems of detection,
segmentation, pose estimation and fine-grained recognition, but it goes beyond. The
explicit correspondence with a 3D CAD model allows a representation which, from
a robotics perspective, can be used directly for trajectory
optimization, motion planning, and grasp estimation among other tasks. In this
setting, a coarse output such as a bounding box at 50\% overlap around the
instance, or a segmentation mask marking pixels belonging to the object, or a
fine-grained distinction between an office chair and a dinning room table, or a
coarse viewpoint estimate of front facing versus side facing are insufficient.

\begin{figure} \centering \insertA{0.48}{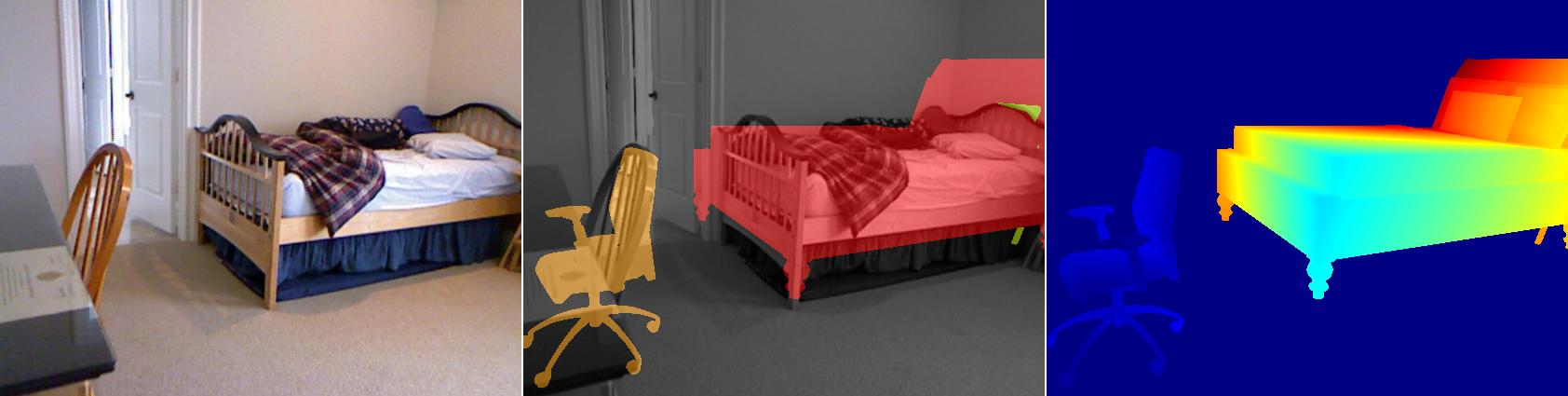} \\
\insertA{0.48}{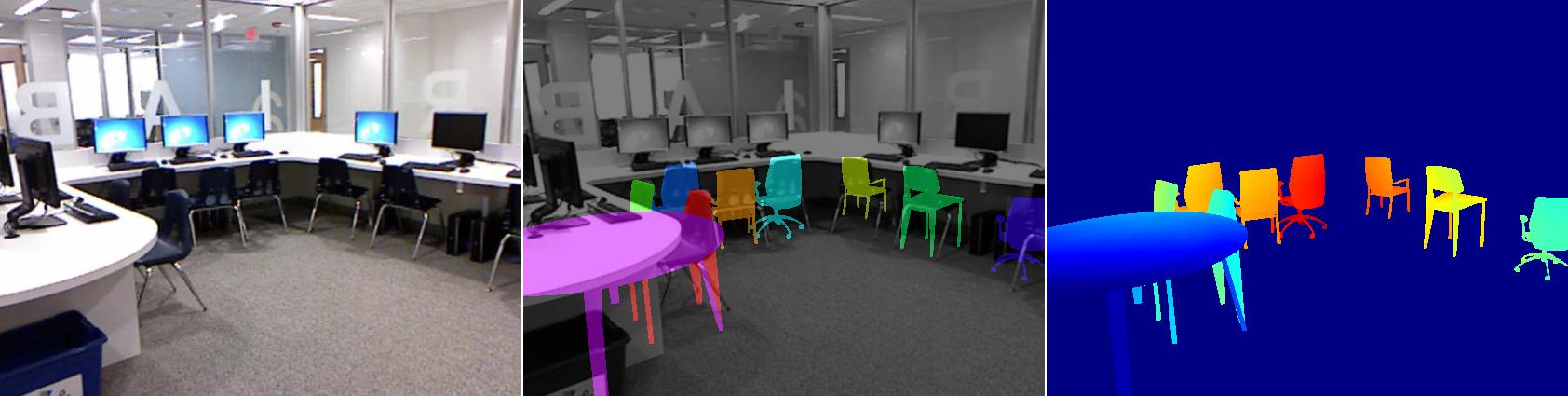} \caption{\textbf{Output of our system}:
Input \rgbd image and output a 3D model associated with objects in the
scene.} \figlabel{teaser} \end{figure}

\figref{overview} describes the pipeline of our approach. We first use the
output of the state-of-the-art detection and segmentation system
\cite{guptaECCV14}, and infer the pose of the object using a neural network. We
train this \cnn on synthetic data and use normal images instead of depth images
as input. We show that this \cnn trained on synthetic data works better than
the one trained on real data. We then use the top two inferred pose hypothesis
to initialize a search over a small set of 3D model, their scales and exact
placement. We use iterative closest point (ICP) for doing this and show that
when initialized properly this works well even when working at the level of
object categories rather than exact instances for which ICP has traditionally
been used. In doing so we only use 2D annotations on the image and are able to
generate a 3D representation of the scene richer than one that was annotated. 

Our final output is a 3D model that has been aligned to the objects present in
the image. The richness and quality of the output from our system is
illustrated when we compare to current state-of-the-art methods for 3D
detection. A natural side-product of our output is a 3D bounding box for each
object in the scene. When we use this 3D bounding box for 3D detection we
observe that we are able to outperform the current state-of-the-art method
\cite{songECCV14} by 19\% absolute AP points (48\% relative), while at the same
time being at least an order of magnitude faster.

\begin{figure} \centering \insertA{0.50}{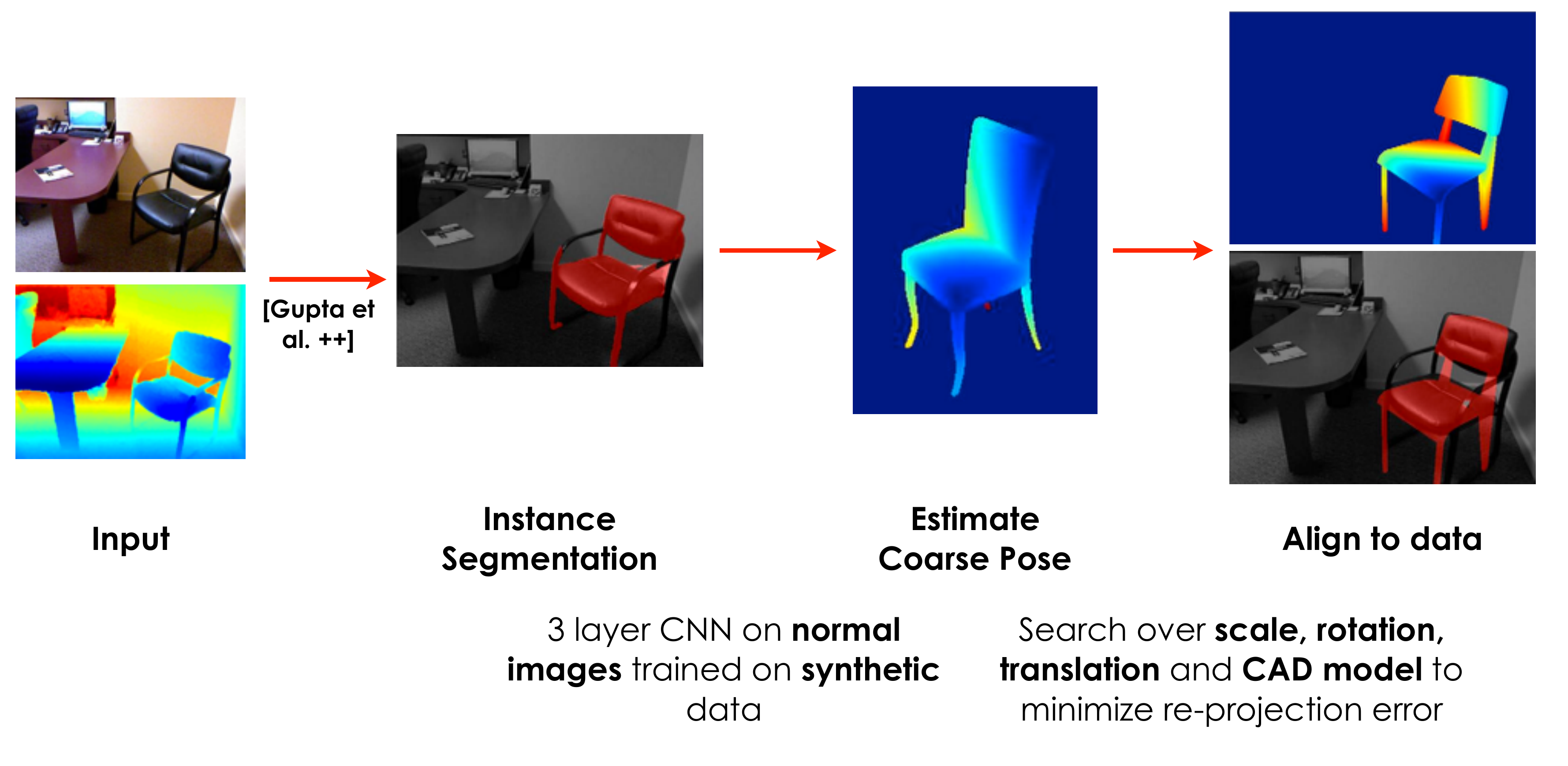}
\caption{\textbf{Overview of approach}: We start with object detection and
instance segmentation output from Gupta \etal \cite{guptaECCV14}. We first infer
the pose of the object using a convolutional neural network, and then search for
the best fitting model that explains the data.} \figlabel{overview} \end{figure}

\section{Related Work} 
A large body of work in computer vision has focused on
the problem of object detection, where the final output is a bounding box around
the object, overlapping with the actual extent of the object by more than
50\%
\cite{lsvm-pami,viola01,Dalal05,girshickCVPR14,Rowley95}.
There has also been substantial work on labeling each pixel in the image with a semantic
label \eg \cite{o2p,arbelaez2012semantic}. Recent work from Hariharan \etal
\cite{hariharanECCV14}, Tighe \etal \cite{tigheCVPR14} brings these two lines 
of work together by inferring the pixel support of object instances.

There have been corresponding works for \rgbd images studying the problem of
object detection
\cite{b3do,kimCVPR13,laiICRA11,laiICRA12,tangACCV13,guptaECCV14,songECCV14,laiICRA14,linICCV13,boIJRR14},
semantic segmentation
\cite{o2p-d,guptaCVPR13,koppulaNIPS11,renCVPR12,silbermanECCV12,guptaECCV14,silbermanECCV14},
and more recently instance segmentation \cite{guptaECCV14,silbermanECCV14}.
Since our approach builds on an object detection system, we discuss this body of
research in more detail.  \cite{guptaIJCV14,b3do,kimCVPR13,tangACCV13} propose
modification to deformable part models \cite{lsvm-pami} to adapt them to
the \rgbd domain. Gupta \etal \cite{guptaECCV14} also reason in the 2D image space and
propose a geocentric embedding for depth images into horizontal disparity,
height above ground and angle with gravity to learn features on bottom-up
bounding box proposals using a \cnn. They also produce an instance
segmentation where they label pixels belonging to the object for each detection.
\cite{linICCV13} also operate in a similar paradigm of reasoning with bottom-up
region proposals, but focus on modeling object-object and object-scene context.

We note that, although all of these outputs are useful representations, but each
of them is far from an understanding of the world that would enable a robot to
interact with it. 

Of course we are not the first one to raise this argument. There has been a lot
of research on 3D scene understanding from a single \rgb image
\cite{hedauECCV10,satkinIJCV14}, and 3D object analysis
\cite{limECCV14,aubryCVPR14,hejratiNIPS12,ziaPAMI13,schwingICCV13}. Given the
challenging nature of the problem, most of these works either study unoccluded
clean instances, or fail under clutter. In this paper, we study the problem
in the context of the challenging \nyu dataset and analyze how \rgbd data can be
effectively leveraged for this task. 

The most relevant research to our work comes from Song and Xiao
\cite{songECCV14} and Guo and Hoiem \cite{guoThesis}. Song and Xiao
\cite{songECCV14} reason in 3D and train exemplar SVMs using synthetic data and
slide these exemplars in 3D space to search for objects thus naturally dealing
with occlusion, and study the tasks of both 2D and 3D detection. Their 
approach is inspiring but computationally expensive (25 minutes per image
per category). \cite{songECCV14} also show examples where their model is able
to place a good fitting exemplar to data, but they do not empirically study the
problem of estimating good 3D models that fit the data. We differ from their
philosophy and propose to do 2D reasoning to effectively prune out large
parts of the search space, and then do detailed 3D reasoning with the top few
winning candidates. As a result, our final system is significantly faster
(taking about two minutes per image).
% \todo{most of this time gets spent in
% computing bottom-up bounding boxes and regions and there are faster alternatives
% to doing these which we don't use here} 
We also show that lifting from a 2D
representation to a 3D representation is possible and show that naively fitting
a box around the detected region outperforms the model from
\cite{songECCV14}.

Guo and Hoeim \cite{guoThesis} start with a bottom-up segmentation, retrieve
nearest neighbors from the training set, and align the retrieved
candidate with the data. In contrast, we use category knowledge in the
form of top-down object detectors and inform the search procedure about
the orientation of the object. Moreover, our algorithm does not rely on detailed
annotations (which take about 5 minutes for each scene) \cite{guoICCV13} of the
form used in \cite{guoThesis}. We also propose a category-level metric to
evaluate the rich and detailed output from such algorithms. Finally,
\cite{slamCVPR13,shaoTOG12} among many others, work on the same problem but
either consider known instances of objects, or rely on user interaction.

\section{Estimating Coarse Pose}
\seclabel{coarse-pose}
In this section, we propose a convolutional neural network to estimate the
coarse pose of rigid objects from a depth image. 

We first observe that reliable annotations for such a detailed task are extremely
challenging to obtain \cite{guoICCV13}. At the same time, a large amount of
synthetic data can be obtained by rendering a library of 3D models in different
poses.  Thus, it is desirable to be able to train the algorithm on synthetic data
instead of real data. 

Secondly, the representation of the input is important. Gupta \etal
\cite{guptaECCV14} proposed an \hha embedding for a depth image for feature
learning in a \cnn, and demonstrated that this embedding is superior to just
using the depth image by itself. We observe that, while this embedding is
appropriate for detecting objects (a chair is defined by a horizontal surface at
some height), it effectively removes critical information about the pose of the
object. As an example, consider a chair rotating about the vertical axis. As the chair
rotates, its pose changes, but both its height above ground and its angle with
gravity remain fairly constant. It is therefore important to choose an
appropriate depth embedding for the task we address.

Thirdly, occlusion is a predominant phenomenon in indoor environments (consider
for instance chairs in a conference room). It will be desirable for an algorithm
for pose estimation to be robust to occlusion where parts of the object may
not be visible. 

Lastly, depth images have more regularities than color images
(because of the absence of texture), and we want to generalize from synthetic
data to real data, hence the network should not have a very large capacity.

With these motivations in mind, we propose to learn our coarse pose estimator from
synthetic data which consists of aligned 3D CAD models for different categories.

Assume $C(k,n,s)$ is a convolutional layer with kernel size $k\times k$, $n$
filters and a stride of $s$, $P_{\{max,ave\}}(k,s)$ a max or average pooling
layer of kernel size $k\times k$ and stride $s$, $N$ a local response
normalization layer, $RL$ a rectified linear unit, and $D(r)$ a dropout layer
with dropout ratio $r$. Our network has the following architecture:
$C(7,96,4)-RL-P_{max}(3,2)-D(0.5)-N-C(5,128,2)-RL-P_{max}(3,2)-N-C(3,|(N_{pose}+1)N_{class}|,1)-RL-P_{ave}(14,1)$

We chose a smaller fully convolutional network to account for the fact that
depth images have more regularities. In the same vein, the classification layer
at the end is an average pooling over the neurons of the previous layer. We
deliberately did not use fully connected layers and introduced a dropout layer
after $conv1$ in order to make the network robust to occlusion. 

We observe that the HHA encoding proposed in
\cite{guptaECCV14} explicitly removes the azimuth direction by considering only
the disparity, angle with gravity and height above ground. Thus, we propose to
use the normal image as input to the network. We use 3 channel normal images
where the three channels encode $N_x$, $N_y$ and $N_z$ as the angle the
normal vector makes with the three geocentric directions estimated using the
algorithm from \cite{guptaCVPR13}. We scale this to be in degrees and shift it
to center at 128 instead of 90. 

We train this network for classification using a soft-max loss and share the
lower layers of the network among different categories. We also adopt the
geocentric constraint and assume that the object rests on a surface and hence
must be placed flat on the ground.  Thus, we only have to determine the azimuth
of the object in the geocentric coordinate frame. We bin this azimuth into
$N_{posebin}$ bins and train the network to predict the bin for each example. 

We use 3D models from ModelNet \cite{wuARXIV14} to train the network. In
particular, we use the subset of models as part of the training set and work
with the 10 categories for which the models from each category are aligned to
have a canonical pose (bathtub, bed, chair, desk, dresser, monitor, night-stand,
sofa, table, toilet). We sample 50 models for each category and render 10
different poses for each model placed on a horizontal floor at locations and
scales as estimated from the \nyu dataset \cite{silbermanECCV12}. We place one
object per scene, and sample boxes with more than 70\% overlap with the ground
truth box as training examples. We crop and warp the bounding box in the same way
as Girshick \etal \cite{girshickCVPR14}. Note that warping the normals preserves
the angles that are represented (as opposed to warping a depth image or a HHA
image \cite{guptaECCV14} which will change the orientation of surfaces being
represented).

At test time, we simply forward propagate the image through the network and take
the output pose bin as the predicted pose estimate. Given that the following
stage requires a good initialization, we work with the top $k (= 2)$ modes of
prediction, rather than a single prediction.

\section{Model Alignment} In this section we describe how we place the object in
the scene.  We start from the instance segmentation output from
\cite{guptaECCV14}, and infer the coarse pose of the object using the neural
network introduced in \secref{coarse-pose}. With this rough estimate of the
pixel support of the object and a coarse estimate of its pose, we solve an
alignment problem to obtain an optimal placement for the object in the scene.

\subsection{Model Search} Note that our pose estimator only gives us an
orientation for the model. It does not inform about the size of the object, or
about which model would fit the object best. 
Thus this stage has to search over scale and models and infer the exact rotation
$R$ and translation $t$ that aligns the model best with the data. We search over
scale and models, optimizing for a rotation $R$ and a translation $t$ that best
explain the data.

To search over scale, we gather category level statistics from the 3D bounding
box annotations from \cite{guoICCV13}. In particular, we use the area of the
bounding box in the top view, and estimate the mean of this area and its standard
deviation, and take $N_{scale}$ stratified samples from $\mathcal{N}(\mu_{area},
\sigma_{area})$. Such statistics do not require annotations and can also be
obtained from online furniture catalogues. To search over scale, we
isotropically scale each model to have this area in the top-view.

To search over models, we manually pick a small number $N_{models}$ (about 5) of
3D models for each category. Care was taken to pick distinct models, but this
could also be done in a data-driven manner (by picking models which explain
better the data at hand).

Finally, we optimize over $R$ and $t$. We do this iteratively using iterative
closest point (ICP) \cite{rusinkiewicz2001efficient}, which we modify by 
constraining the rotation estimate to be consistent with the gravity direction. We
initialize $R$ using the pose estimate obtained from \secref{coarse-pose}, and
the inferred direction of gravity \cite{guptaCVPR13}. We initialize translation
$t$ by using the median of the world co-ordinates of the points in the
segmentation mask, to set $t_x$ and $t_z$, and set $t_y$ such that the model is
resting on the floor. (This constraint helps with heavily occluded objects \eg
chairs for which often only the back is visible). The following subsection
describes the model alignment procedure.

\subsection{Model Alignment} The input to the model alignment algorithm is a
depth image $D$, a segmentation mask $S$, a 3D model $M$ at a given fixed scale
$s$ and an initial estimate of the transformation (a rotation matrix $R_0$ and
a translation vector $t_0$) for the model. The output of the algorithm is a
rotation $R$ and a transformation $t$, such that the 3D model $M$ rendered with
transformations $R$ and $t$ explains as many points as possible in the
segmentation mask $S$. We solve this problem approximately by the following
procedure which we repeat for $N$ iterations. 

\begin{enumerate} 
\item \textbf{Render model}: Use the current estimate of the
transformation parameters $(s, R, t)$ to render the model $M$ to obtain a depth
image of the model. Project points from the depth image that belong to the
segmentation mask $S$, and the points from the rendered model's depth image to
3D space, to obtain two point sets $P_{object}$ and $P_{model}$.  
\item \textbf{Re-estimate model transformation parameters}: Run ICP to align
points in $P_{object}$ to points in $P_{model}$. We form correspondence by
associating each point in $P_{object}$ with the closest point in $P_{model}$,
which prevents associations for occluded points in the object. We also reject
the worst 20\% of the matches based on the distance.  This allows the
association to be robust in the presence of over-shoot in the segmentation mask
$S$. Lastly, while estimating the updates of the transformation ${(R, t)}$, we
enforce an additional constraint that the rotation matrix $R$ must rotate the
object only about the direction of gravity.
\end{enumerate}

\subsection{Model Selection} 
Now we need to select the fitted model that best explains the data among
$N_{scale}N_{model}$ candidates. We pose this selection as a learning
problem and compute a set of features to capture the quality of the fit to the
data. We compute the following features: number and fraction of pixels of the
rendered model that are occluded, which are explained by the data, fraction and
number of pixels of the input instance segmentation which are explained by the
model, intersection over union overlap of the instance segmentation with mask of
the model explained by the data, and mask of the model which is unoccluded. We
learn a linear classifier on these features to pick the best fitting model. This
classifier is trained with positives coming from rendered models which have
more than 50\% overlap with a ground truth region.

\section{Experiments} 

\begin{figure*}
\begin{tabular}{ccccc}
\insertA{0.17}{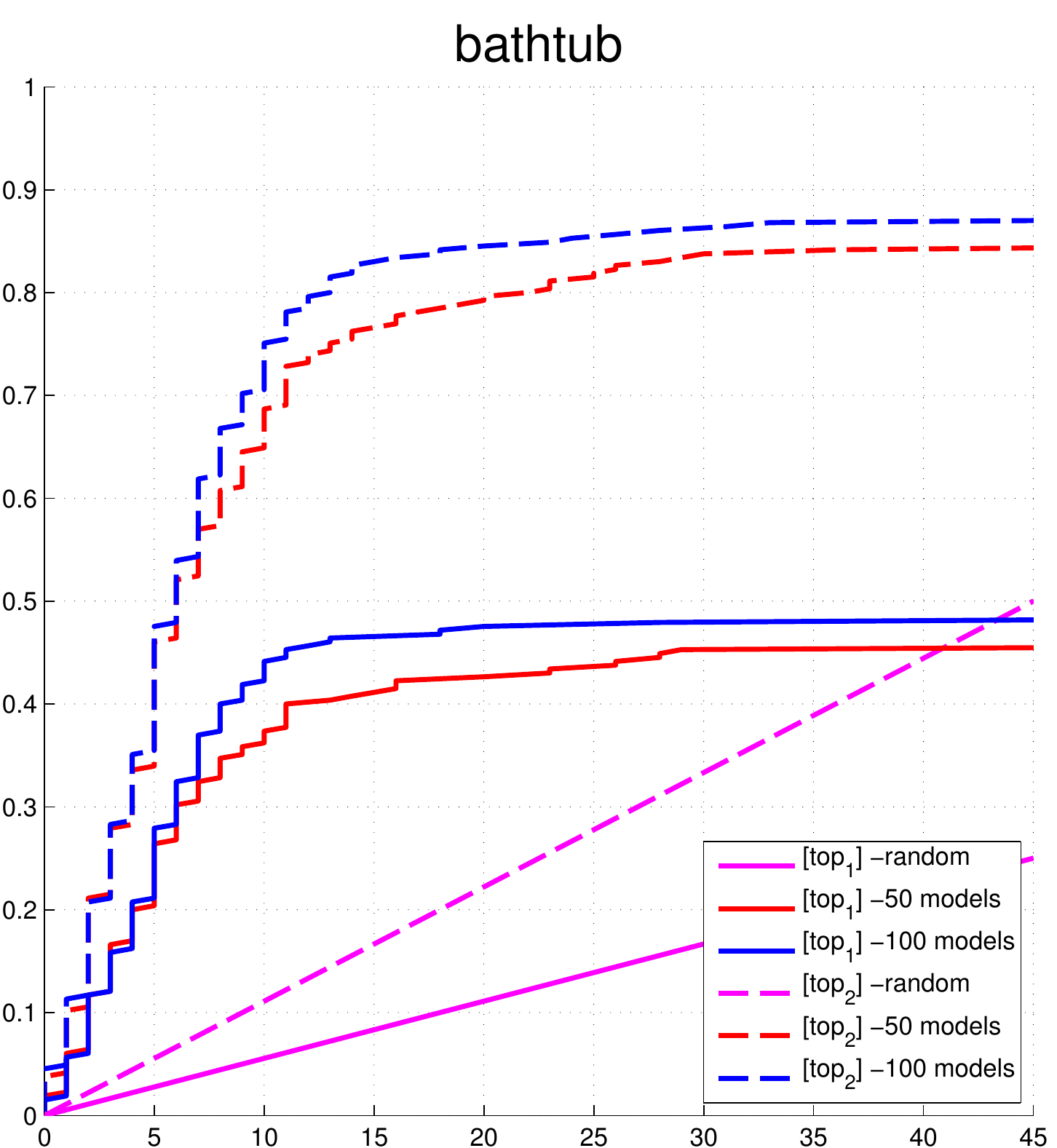} &
\insertA{0.17}{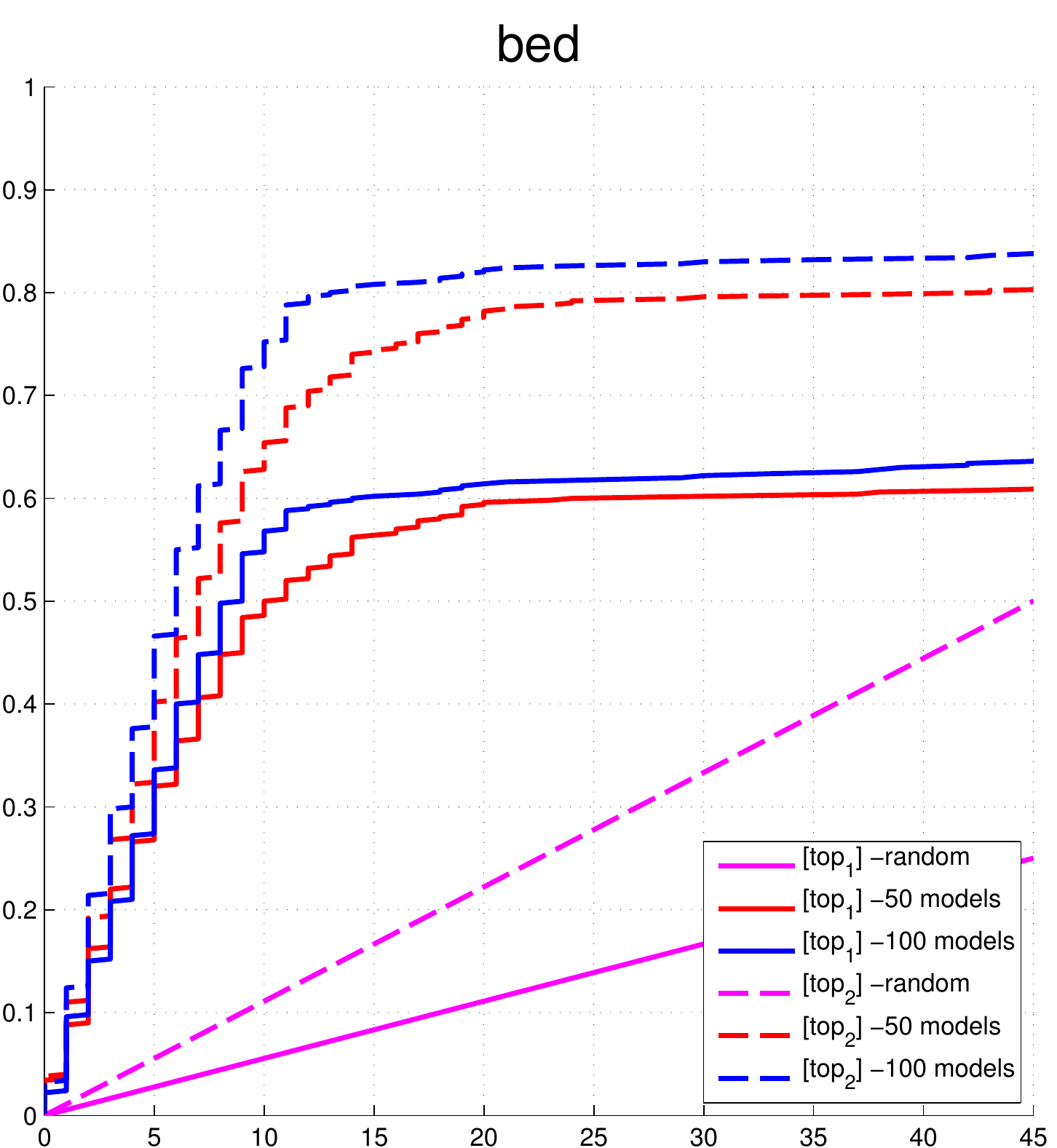} &
\insertA{0.17}{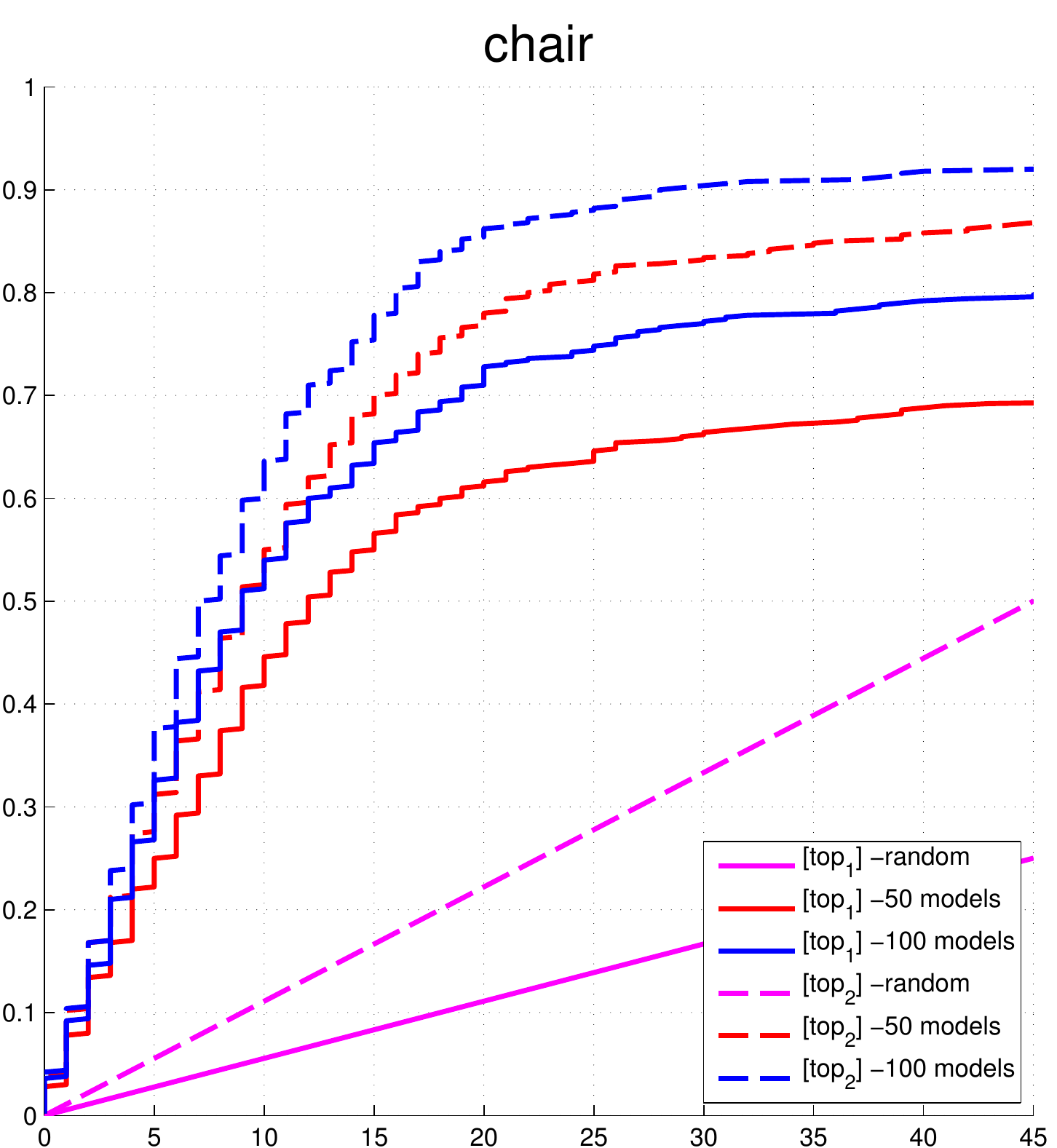} &
\insertA{0.17}{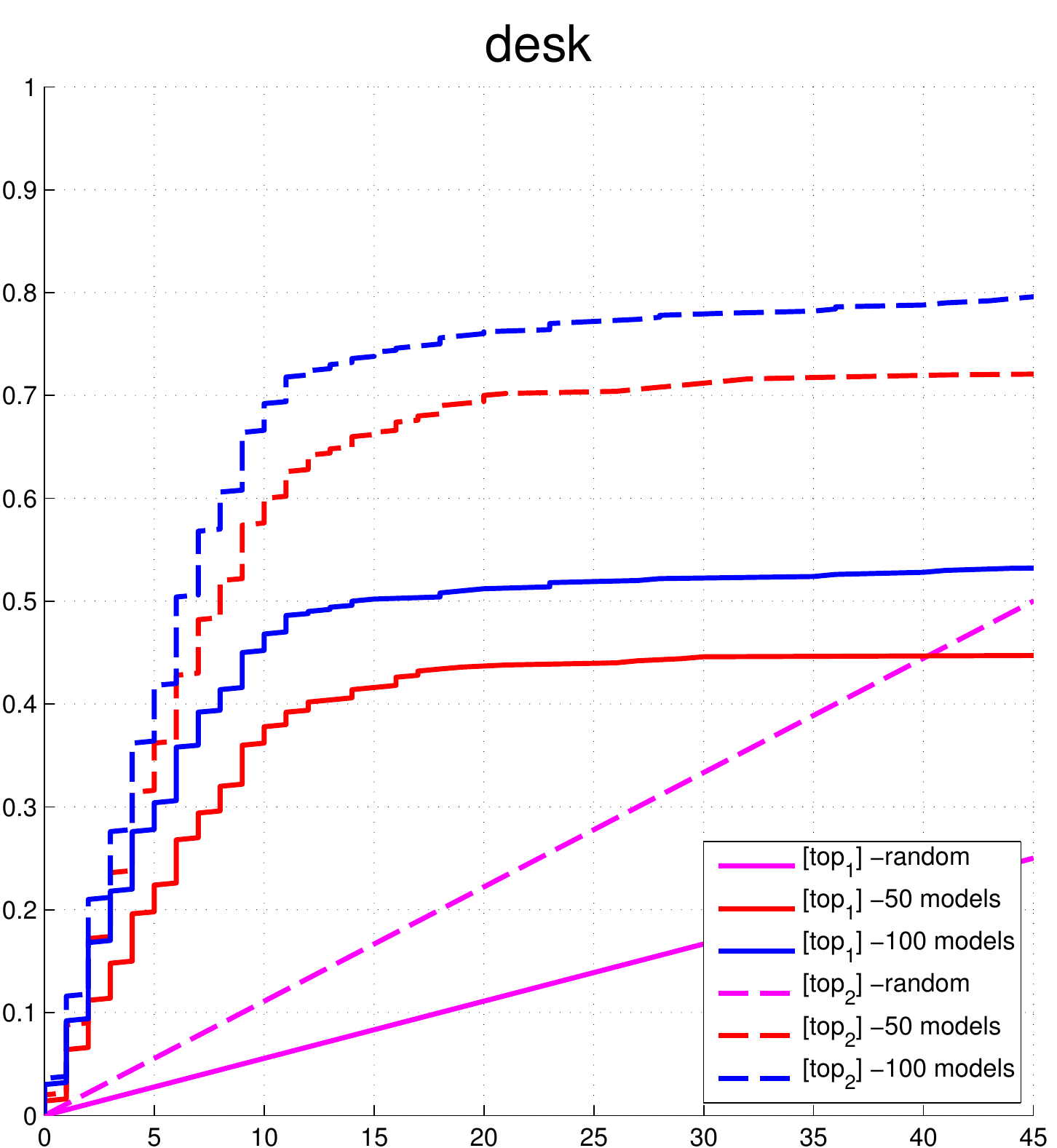} &
\insertA{0.17}{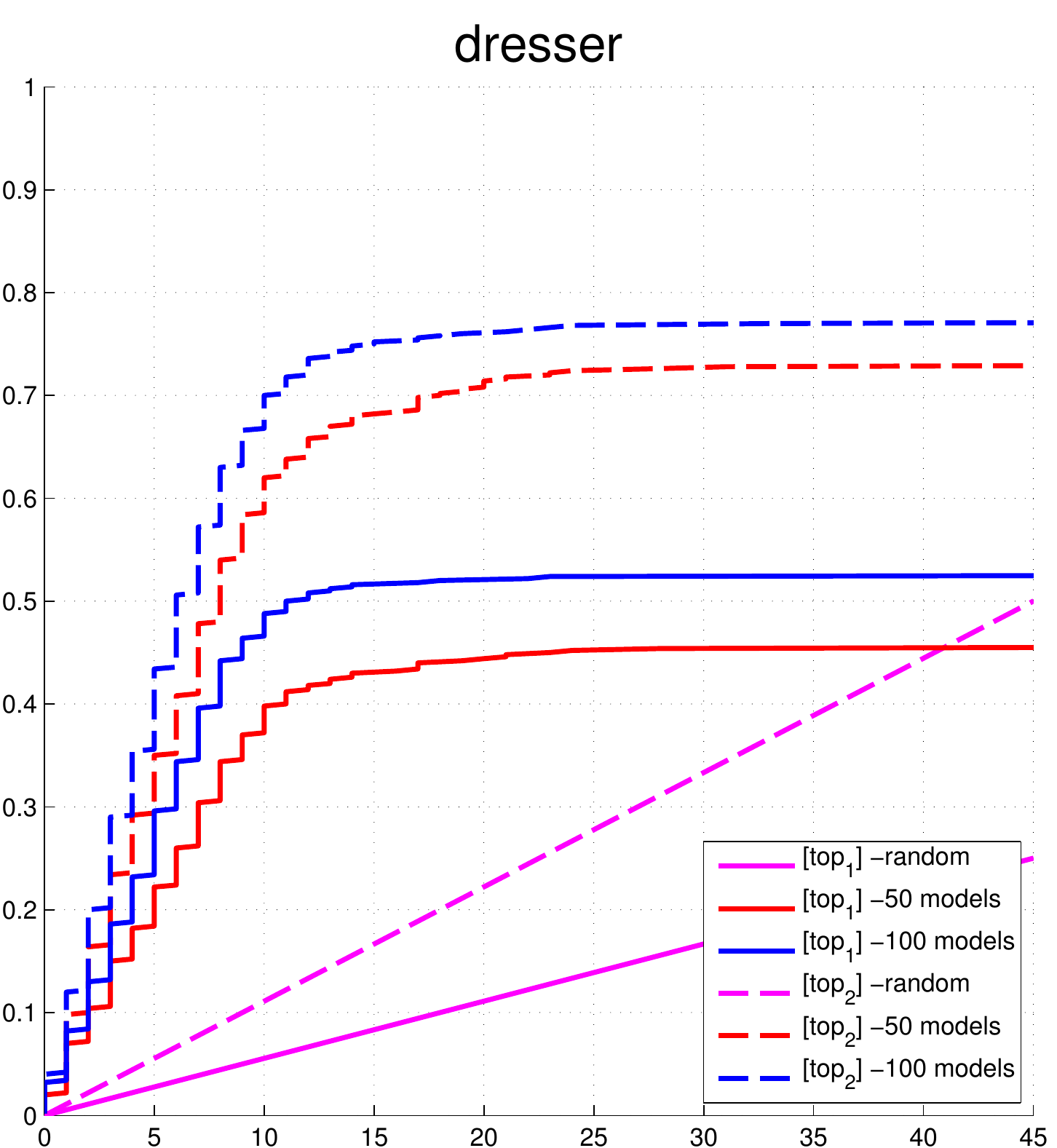} \\ 
\insertA{0.17}{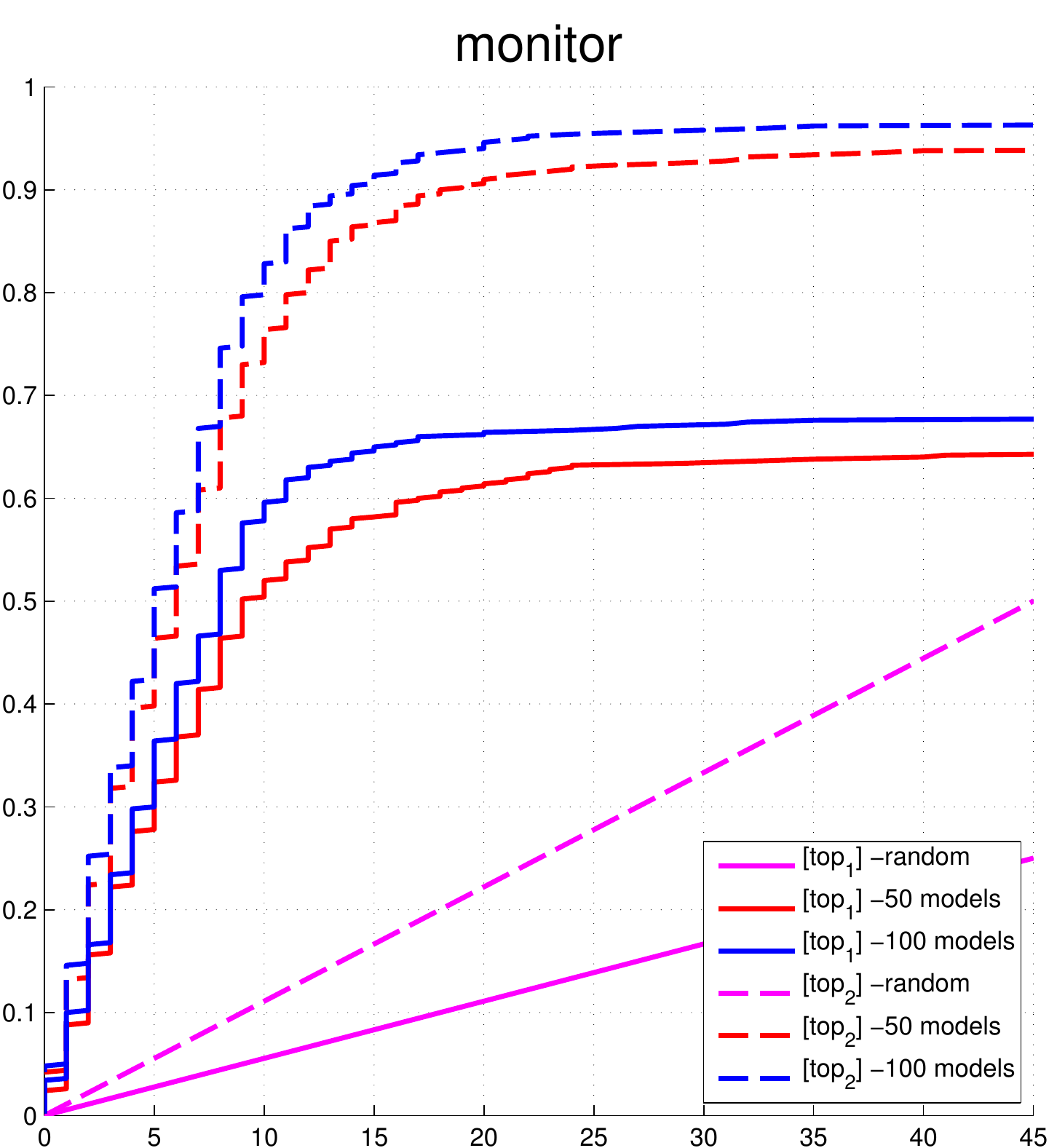} &
\insertA{0.17}{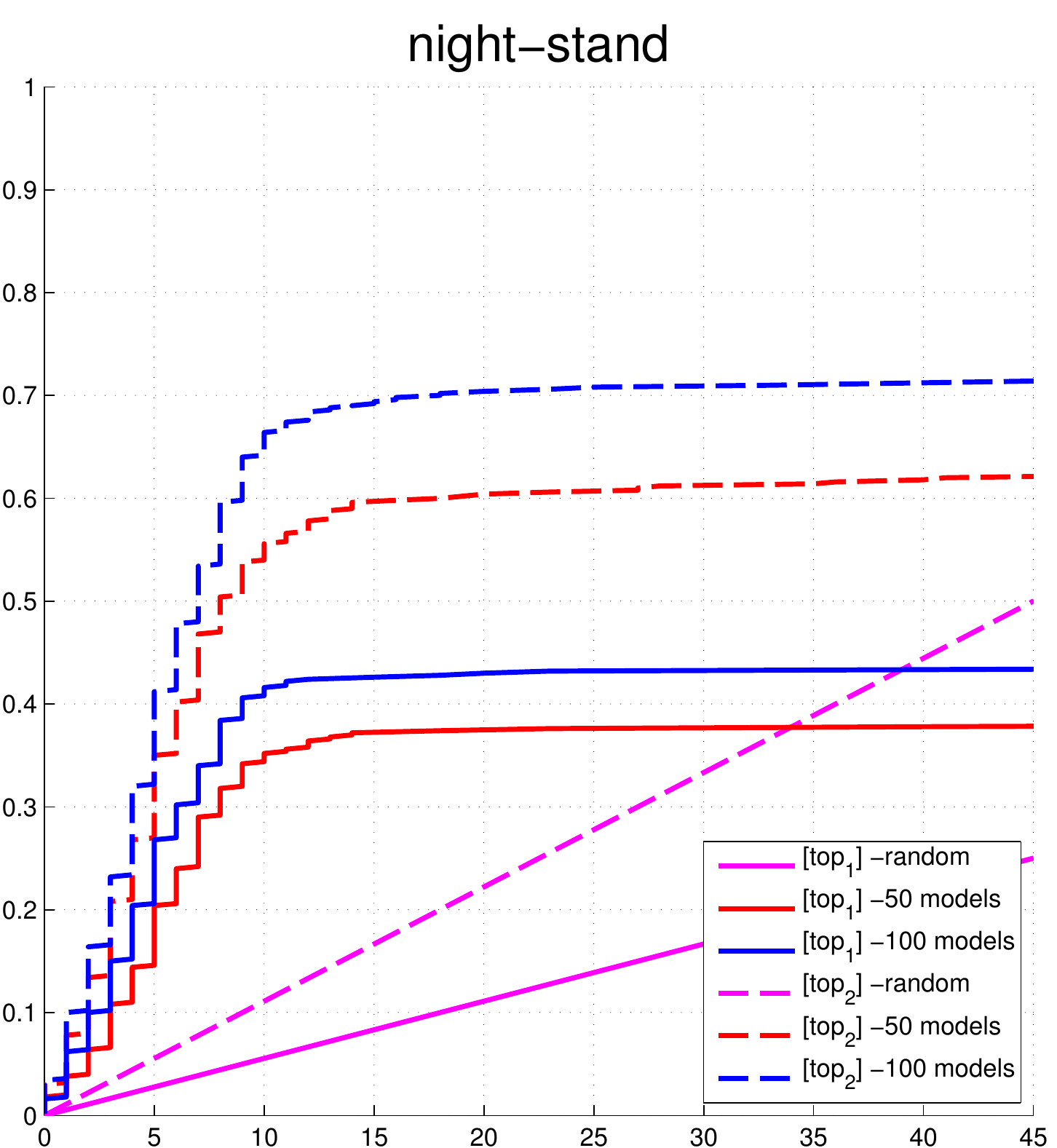} &
\insertA{0.17}{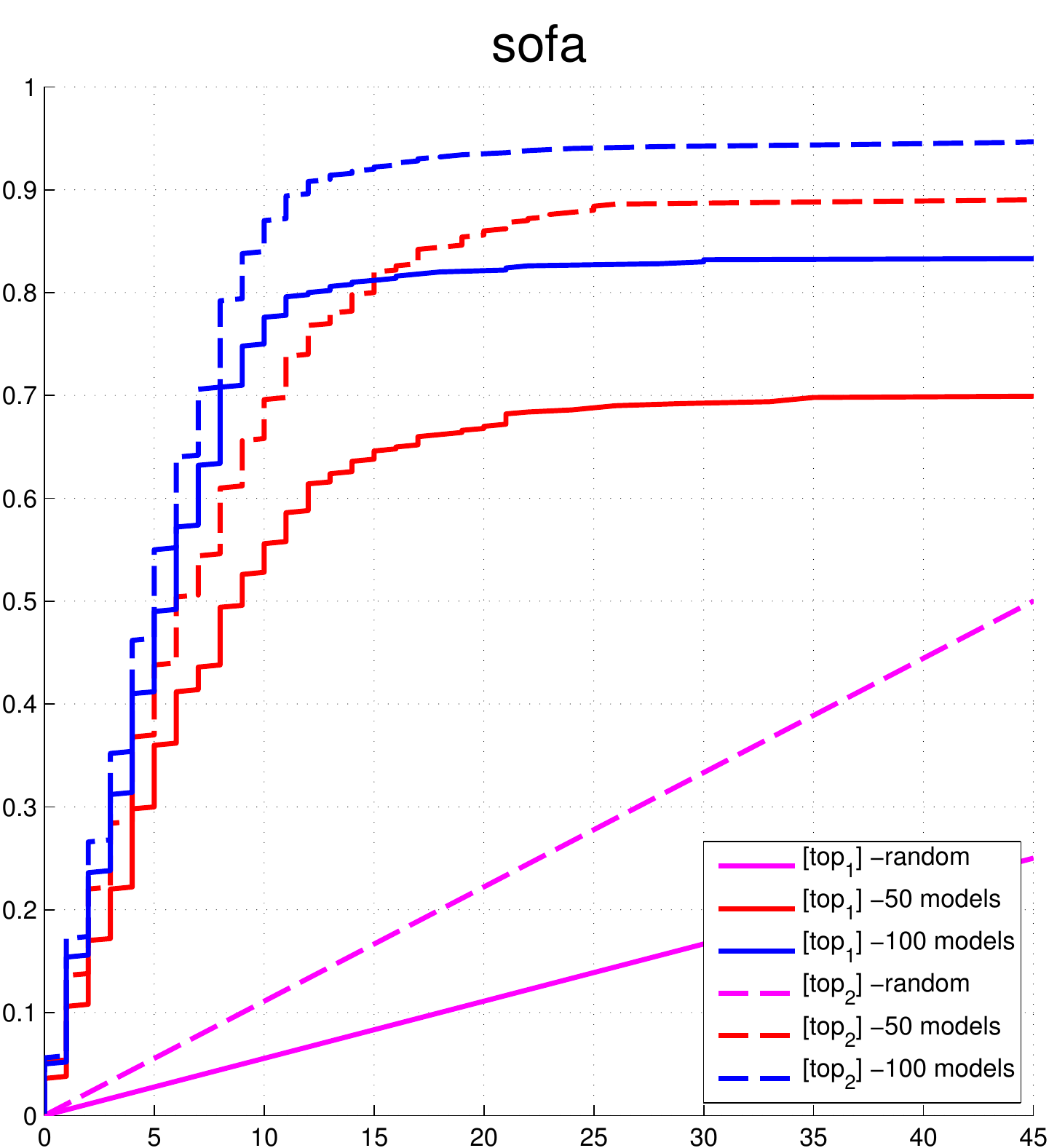} &
\insertA{0.17}{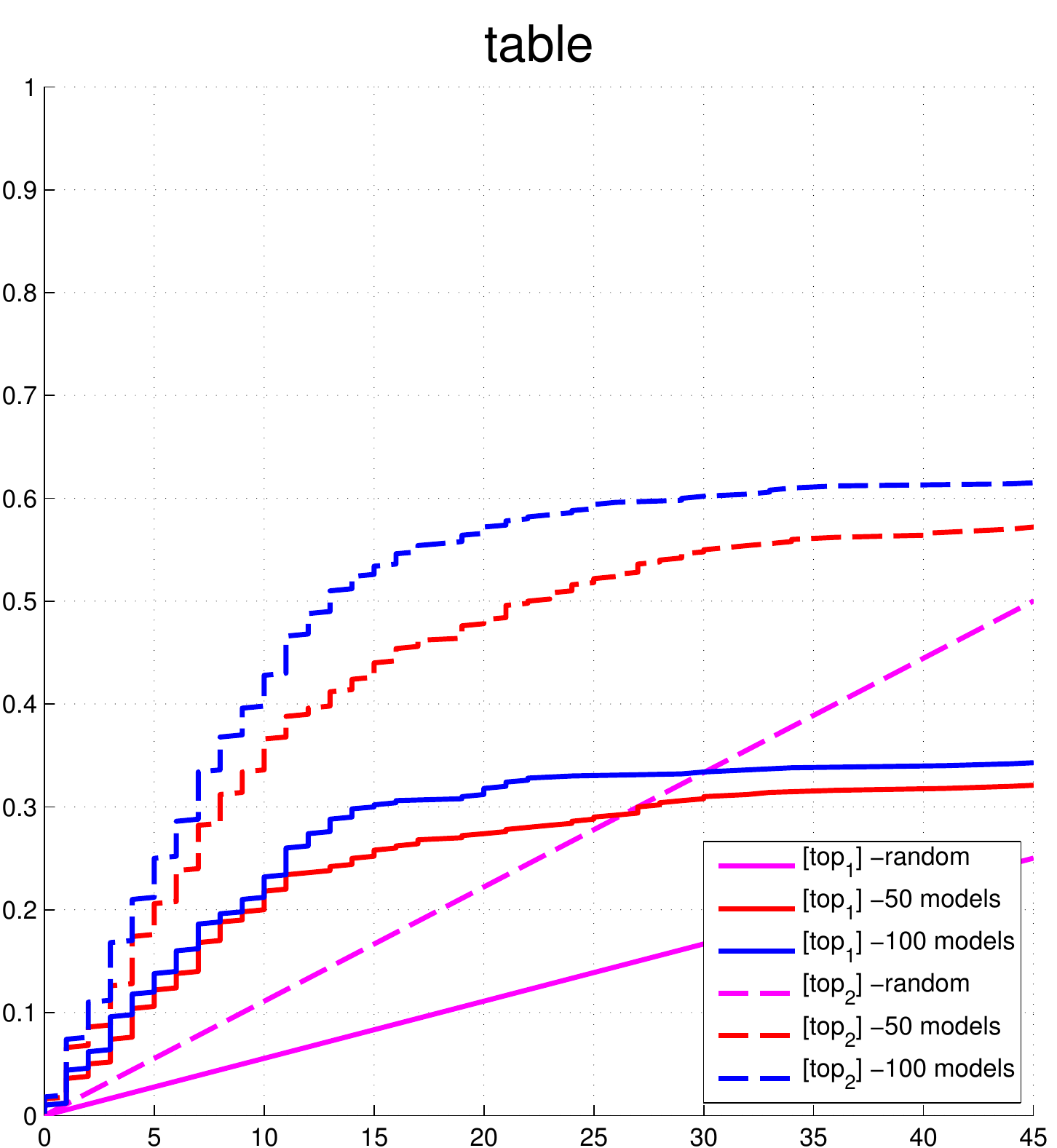} &
\insertA{0.17}{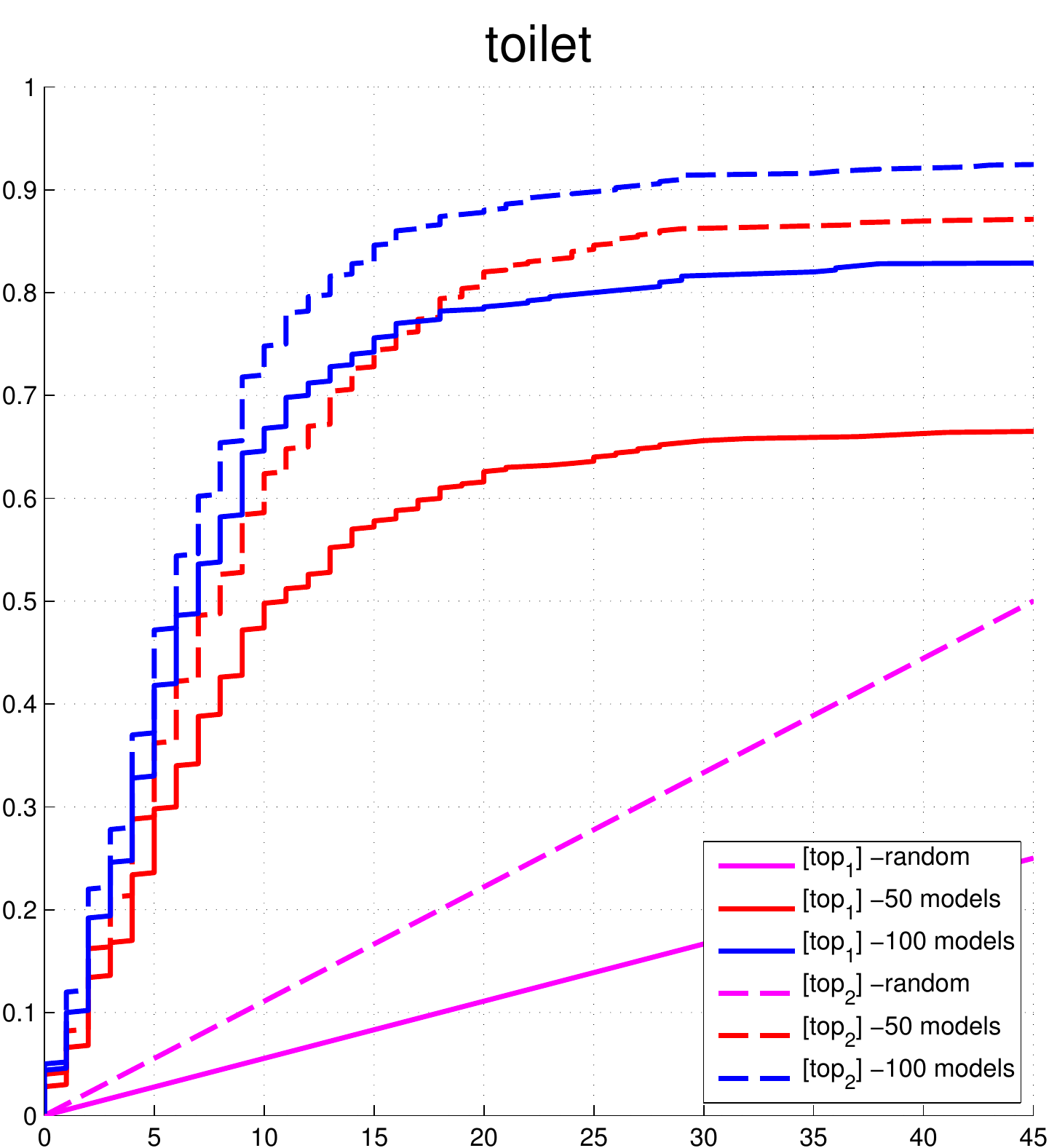} 
\end{tabular}
\caption{Performance on a synthetic test set. We plot accuracy (fraction of
instances for which we are able to predict pose within a $\delta_{\theta}$
angle) as a function of $\delta_{\theta}$. We experiment with using 50 or 100
models for each category for training, and also look at how the performance
changes when using best of $top_1$ or $top_2$ modes of the output.}
\figlabel{coarse-pose-synthetic}
\end{figure*}

\begin{figure*}
\centering
\begin{tabular}{c|c}
\insertA{0.48}{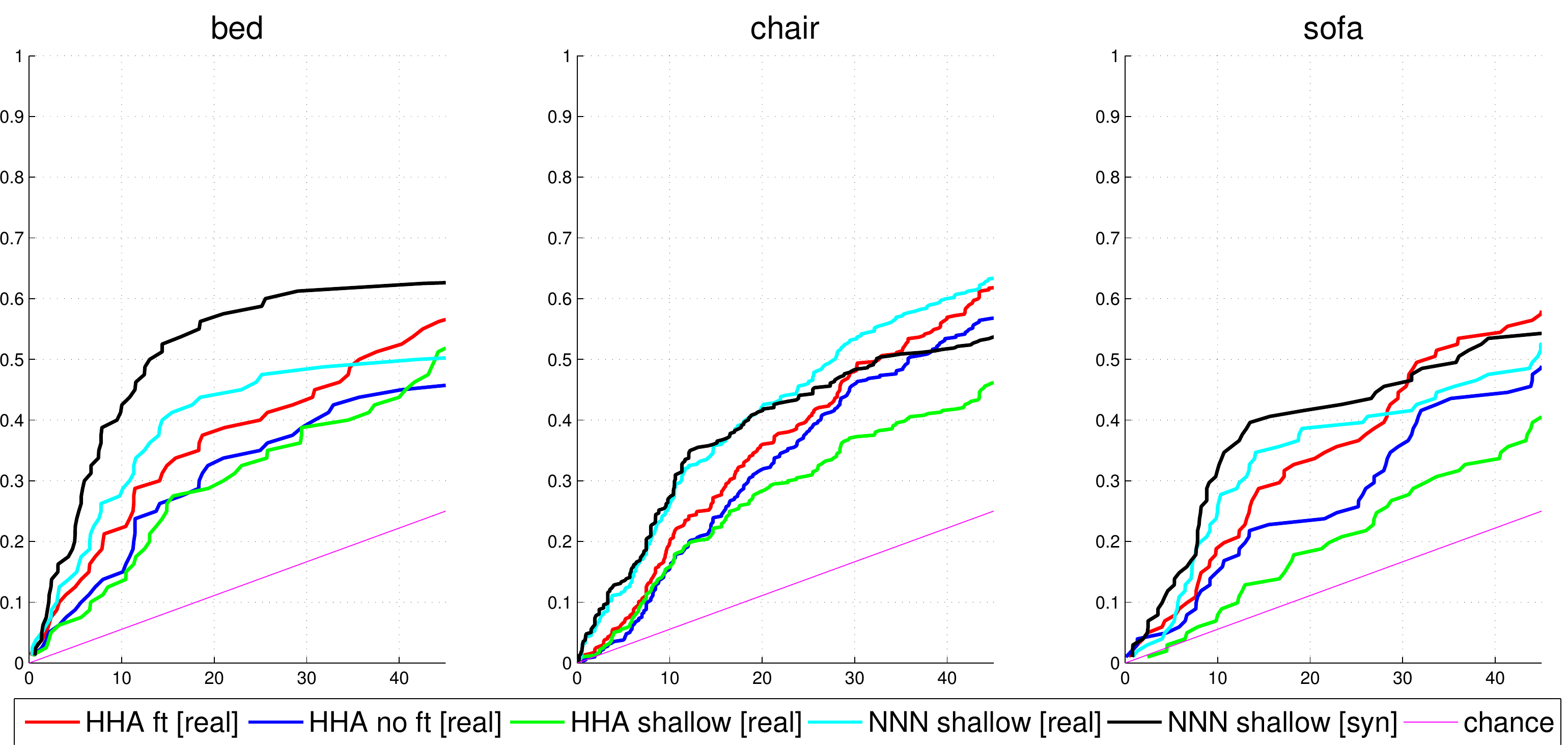} & 
\insertA{0.48}{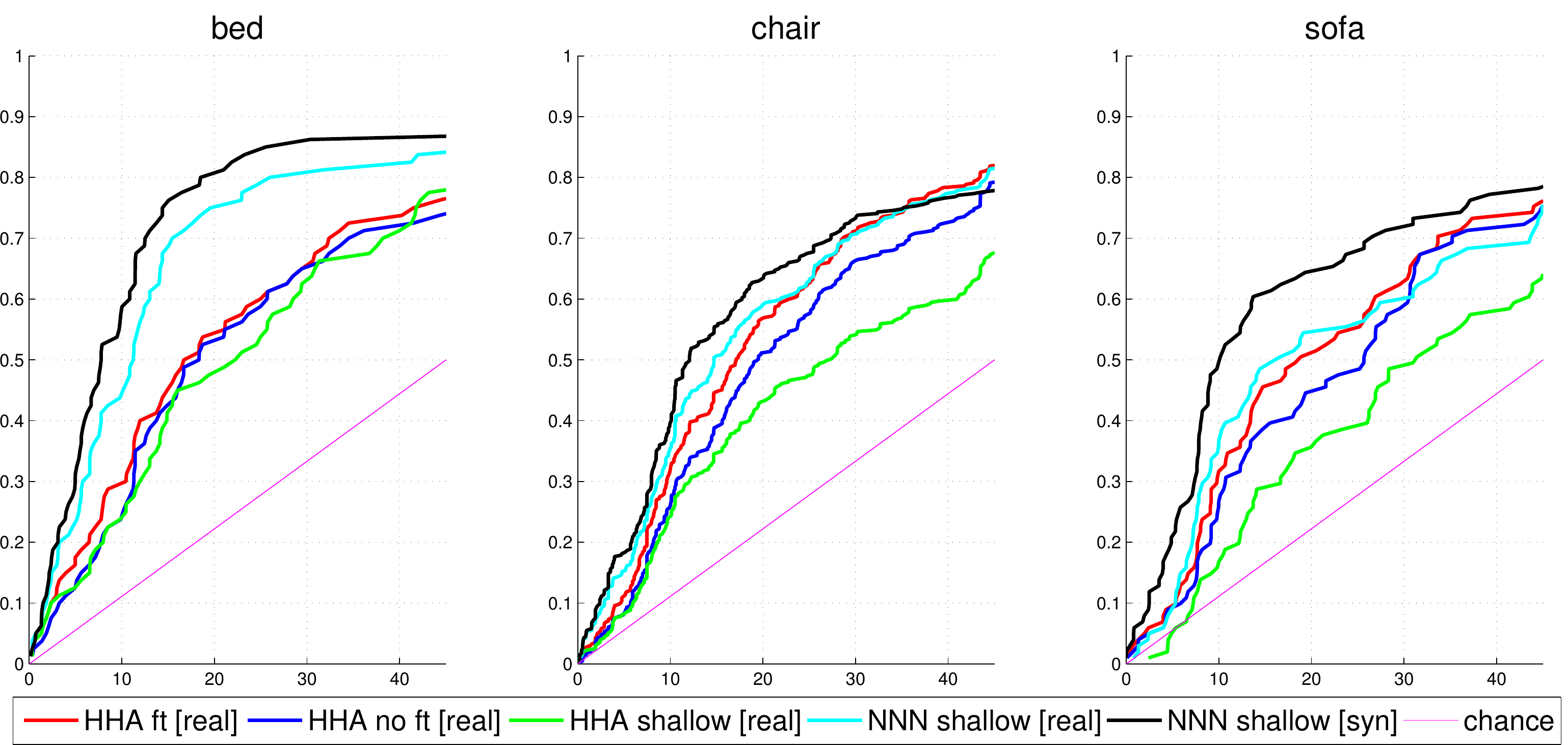}
\end{tabular}
\caption{Performance on a real val set. We plot accuracy (fraction of
instances for which we are able to predict pose within a $\delta_{\theta}$
angle) as a function of $\delta_{\theta}$. The left three plots show $\text{top}_{1}$
accuracy and the right three plots $\text{top}_{2}$ accuracy. Note that real
in the legend refers to model trained on real data, syn refers to the model
trained on synthetic data and NNN stands for normal image.} 
\figlabel{coarse-pose-real}
\end{figure*}

We evaluate our approach on the \nyu dataset from Silberman \etal
\cite{silbermanECCV12} and use the standard train set of 795 images and test set
with 654 images. We split the 795 training images into 381 train and 414
validation images. For synthetic data we use the collection of aligned models
made available by Wu \etal \cite{wuARXIV14}.

\subsection{Coarse Pose Estimation} Here we describe our experiments to evaluate
our coarse pose estimator. We present two evaluations, one on synthetic data and
another one on real data.

\renewcommand{\arraystretch}{1.4}
\begin{table*}
\caption{\textbf{\emph{Test} set results for detection and instance
segmentation on \nyu}: First line reports \boxAP (bounding box detection $AP$)
performance using features from just the bounding box and second line reports
\boxAP when using features from the region mask in addition to features from
the bounding box. Third and fourth lines report the corresponding performance
when using the full trainval set to finetune (instead of only using the train
set).  Subsequent lines report \regionAP (region detection $AP$
\cite{hariharanECCV14}). Using features from the region in addition to features
from the box (row 6) improves performance over the refinement method used in
\cite{guptaECCV14} (row 5). Finally, finetuning over the trainval set boosts
performance further.}
\begin{center}
\setlength{\tabcolsep}{4.0pt}
\scalebox{0.7}{
\begin{tabular}{c|l|c|c|ccccccccccccccccccc}
\vertical{task} & & \vertical {finetuning set} & \vertical{mean} & \vertical{bathtub} & \vertical{bed} & \vertical{book shelf}
& \vertical{box} & \vertical{chair} & \vertical{counter} & \vertical{desk} &
\vertical{door} & \vertical{dresser} & \vertical{garbage bin} & \vertical{lamp}
& \vertical{monitor} & \vertical{night stand} & \vertical{pillow} &
\vertical{sink} & \vertical{sofa} & \vertical{table} & \vertical{television} &
\vertical{toilet} \\ \hline
\multirow{4}{*}{\boxAP} 
& \cite{guptaECCV14} & train & 35.9 & 39.5 & 69.4 & 32.8 & 1.3 & 41.9 & 44.3 & 13.3 & 21.2 & 31.4 & 35.8 & 35.8 & 50.1 & 31.4 & 39.0 & 42.4 & 50.1 & 23.5 & 33.3 & 46.4\tabularnewline
& \cite{guptaECCV14} + Region Features & train & {39.3} & \textbf{50.0} & 70.6 & 34.9 & 3.0 & 45.2 & \textbf{48.7} & \textbf{15.2} & 23.5 & 32.6 & 48.3 & 34.9 & 50.2 & 32.2 & \textbf{44.2} & 43.1 & \textbf{54.9} & 23.4 & 41.5 & 49.9\tabularnewline
& \cite{guptaECCV14} & trainval & {38.8} & 36.4 & 70.8 & 35.1 & 3.6 & 47.3 & 46.8 & 14.9 & 23.3 & 38.6 & 43.9 & \textbf{37.6} & \textbf{52.7} & 40.7 & 42.4 & \textbf{43.5} & 51.6 & 22.0 & 38.0 & 47.7\tabularnewline
& \cite{guptaECCV14} + Region Features & trainval & \textbf{41.2} & 39.4 & \textbf{73.6} & \textbf{38.4} & \textbf{5.9} & \textbf{50.1} & 47.3 & 14.6 & \textbf{24.4} & \textbf{42.9} & \textbf{51.5} & 36.2 & 52.1 & \textbf{41.5} & 42.9 & 42.6 & 54.6 & \textbf{25.4} & \textbf{48.6} & \textbf{50.2}\tabularnewline
\hline 
\multirow{3}{*}{\regionAP} 
& \cite{guptaECCV14} (Random Forests) & train & {32.1} & 18.9 & \textbf{66.1} & 10.2 & 1.5 & 35.5 & 32.8 & \textbf{10.2} & \textbf{22.8} & 33.7 & 38.3 & \textbf{35.5} & 53.3 & \textbf{42.7} & 31.5 & 34.4 & 40.7 & 14.3 & 37.4 & 50.3\tabularnewline
& \cite{guptaECCV14} + Region Features & train & {34.0} & 33.8 & 64.4 & 9.8 & 2.3 & 36.6 & 41.3 & 9.7 & 20.4 & 30.9 & 47.4 & 26.6 & 51.6 & 27.5 & \textbf{42.1} & 37.1 & 44.8 & 14.7 & 42.7 & 62.6\tabularnewline
& \cite{guptaECCV14} + Region Features & trainval & \textbf{37.5} & \textbf{42.0} & 65.1 & \textbf{12.7} & \textbf{5.1} & \textbf{42.0} & \textbf{42.1} & 9.5 & 20.5 & \textbf{38.0} & \textbf{50.3} & 32.8 & \textbf{54.5} & 38.2 & 42.0 & \textbf{39.4} & \textbf{46.6} & \textbf{14.8} & \textbf{48.0} & \textbf{68.4}\tabularnewline
\end{tabular}}
\end{center}
\tablelabel{obj-det-inst-seg}
\end{table*}

\setlength{\extrarowheight}{1pt}
\begin{table*} \caption{\textbf{\textit{Test} set results for 3D detection on
\nyu}: We report the 3D detection AP \cite{songECCV14}. We use the evaluation
code from \cite{songECCV14}. `3D all' refers to the setting with all object
instances where as `3D clean' refers to the setting when instances with heavy
occlusion and missing depth are considered difficult and not used for
evaluation \cite{songECCV14}. See \secref{detection-3d} for details.} 
\begin{center}
\setlength{\tabcolsep}{10pt}
\scalebox{0.7}{
  \begin{tabular}{cl|c|ccccc}
  \multicolumn{8}{c}{3D all} \\
  &  & mean & bed & chair & sofa & table & toilet\\
  \noalign{\hrule height 1.2pt}
  % \multirow{5}{*}{3D all} 
  & Song and Xiao \cite{songECCV14} & 
  39.6 & 33.5 & 29.0 & 34.5 & \notextbf{33.8} & 67.3\\
  & Our (3D Box on instance segmentation from Gupta \etal \cite{guptaECCV14}) & 
  48.4 & \notextbf{74.7} & 18.6 & 50.3 & 28.6 & 69.7\\ 
  & Our (3D Box around estimated model) & 
  \notextbf{58.5} & 73.4 & \notextbf{44.2} & \notextbf{57.2} & 33.4 & \notextbf{84.5}\\
  \noalign{\hrule height .5pt}
  & Our [no RGB] (3D Box on instance segmentation from Gupta \etal \cite{guptaECCV14}) & 
  46.5 & \notextbf{71.0} & 18.2 & 49.6 & 30.4 & 63.4\\ 
  & Our [no RGB] (3D Box around estimated model) & 
  \notextbf{57.6} & 72.7 & \notextbf{47.5} & \notextbf{54.6} & 40.6 & \notextbf{72.7}\\ 
  \multicolumn{8}{c}{} \\
  \multicolumn{8}{c}{} \\
  \multicolumn{8}{c}{3D clean} \\
  &  & mean & bed & chair & sofa & table & toilet\\
  \noalign{\hrule height 1.2pt}
  % \multirow{5}{*}{3D clean} 
  & Song and Xiao \cite{songECCV14} & 
  64.6 & 71.2 & \notextbf{78.7} & 41.0 & \notextbf{42.8} & 89.1\\
  & Our (3D Box on instance segmentation from Gupta \etal \cite{guptaECCV14}) & 
  66.1 & \notextbf{90.9} & 45.9 & 68.2 & 25.5 & \notextbf{100.0}\\
  & Our (3D Box around estimated model) & 
  \notextbf{71.1} & 82.9 & 72.5 & \notextbf{75.3} & 24.6 & \notextbf{100.0}\\ 
  \noalign{\hrule height .5pt}
  & Our [no RGB] (3D Box on instance segmentation from Gupta \etal \cite{guptaECCV14}) & 
  62.3 & \notextbf{86.9} & 43.6 & 57.4 & 26.6 & \notextbf{ 96.7}\\
  & Our [no RGB] (3D Box around estimated model) & 
  \notextbf{70.7} & 84.9 & 75.7 & \notextbf{62.8} & 33.7 & \notextbf{ 96.7}\\ 
  \end{tabular}}
\end{center}
\tablelabel{detection-3d}
\end{table*}

\begin{figure*}
\centering
\begin{tabular}{ccccc}
\insertA{0.18}{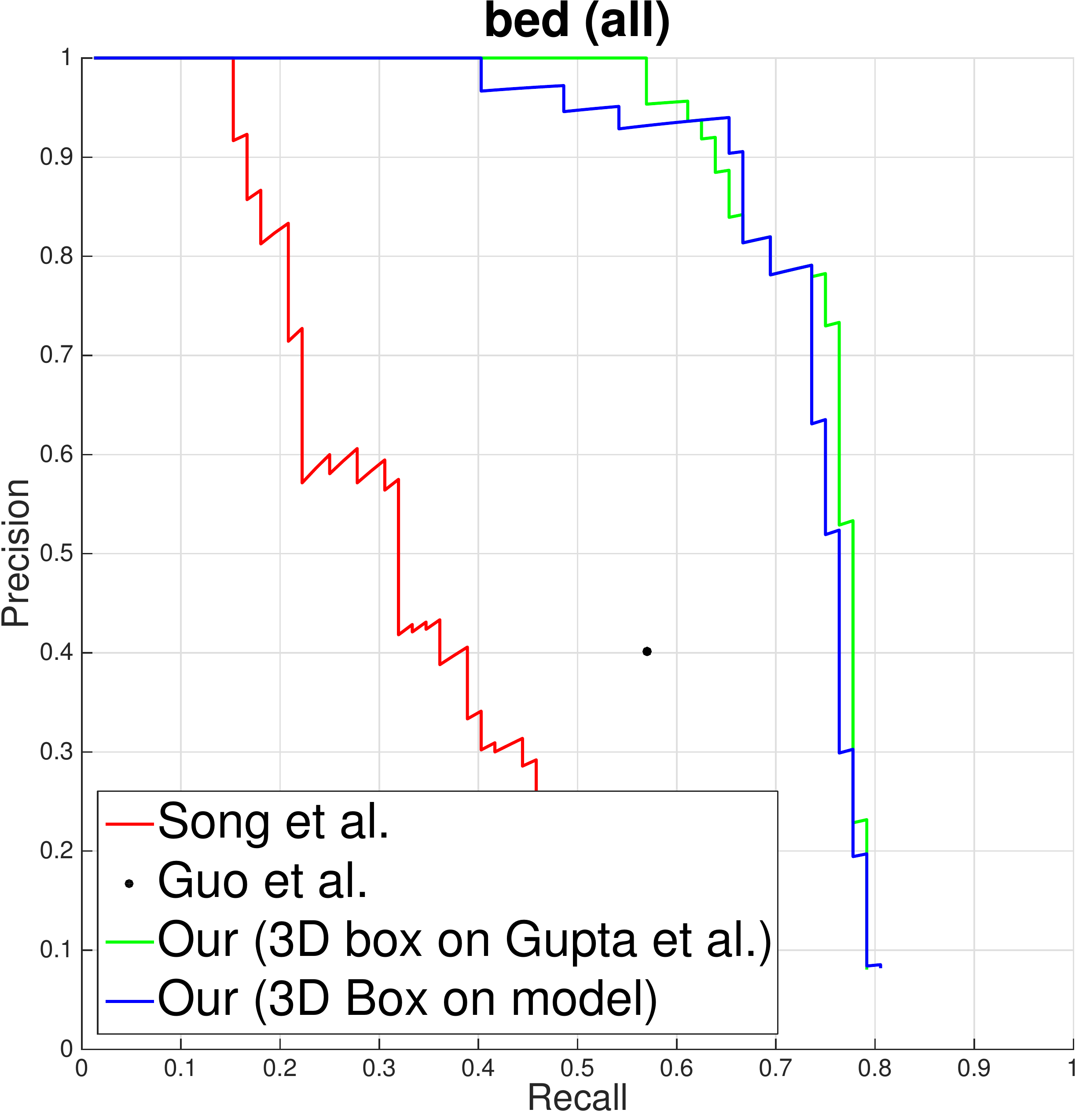} &
\insertA{0.18}{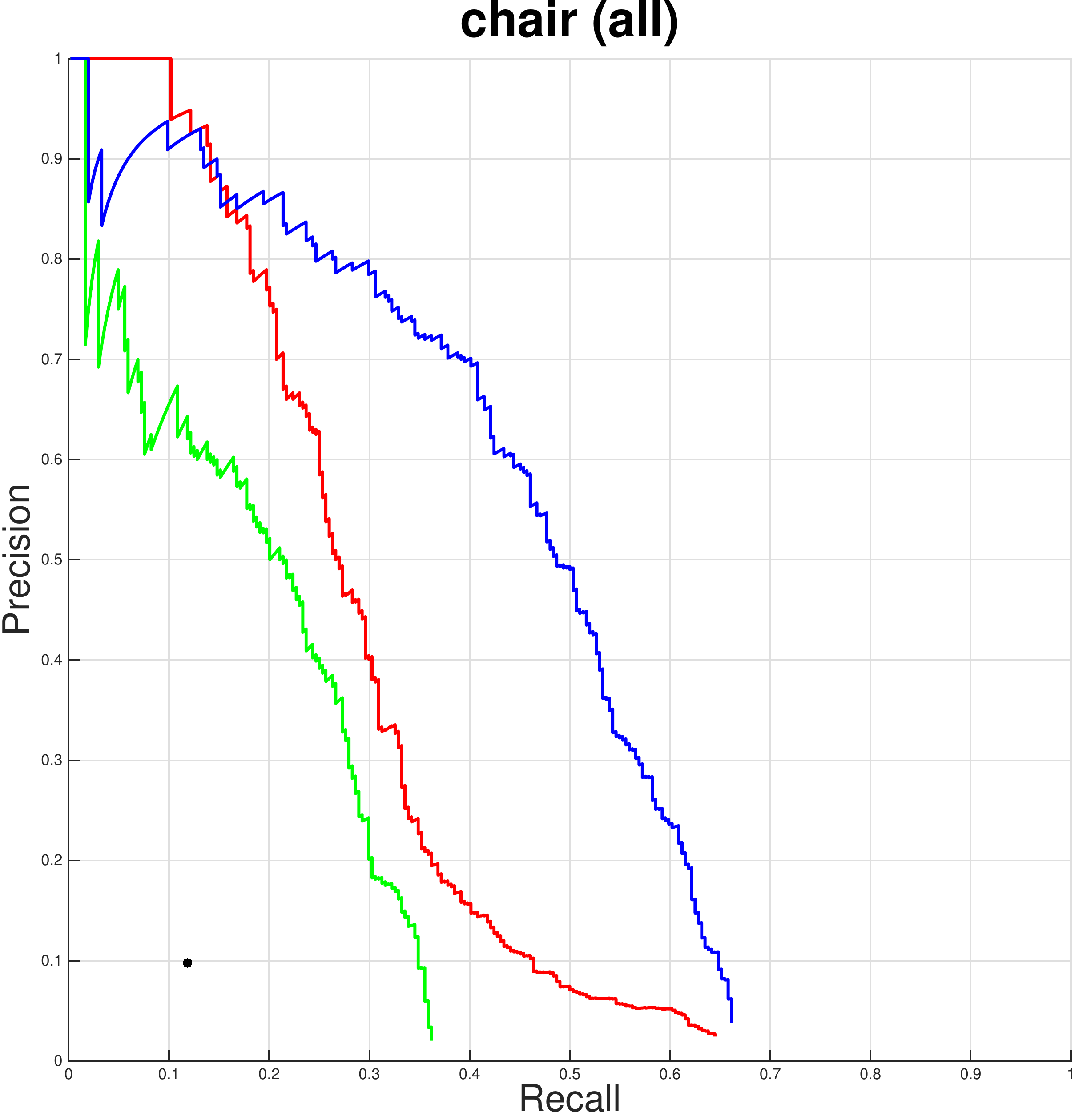} &
\insertA{0.18}{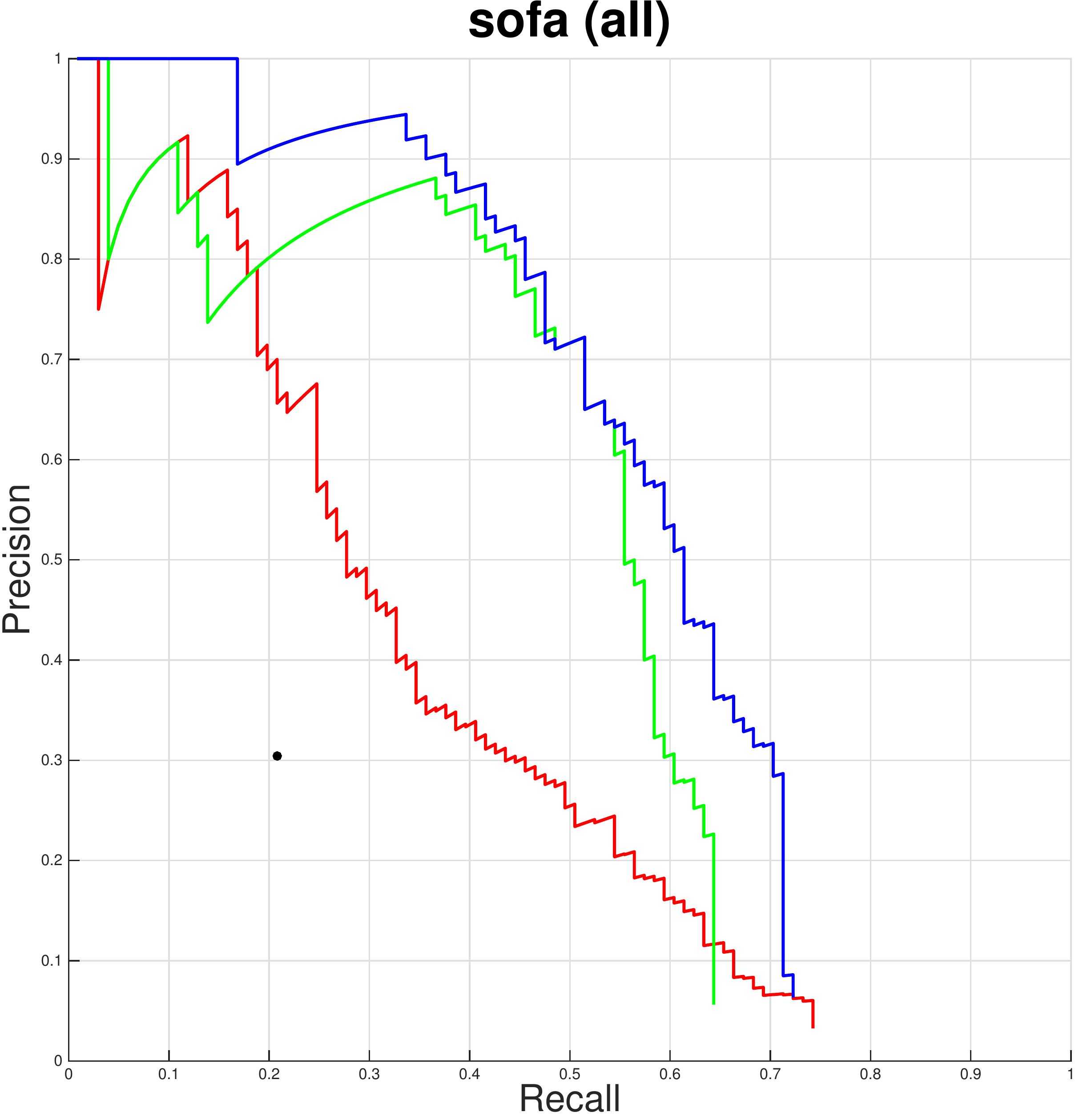} &
\insertA{0.18}{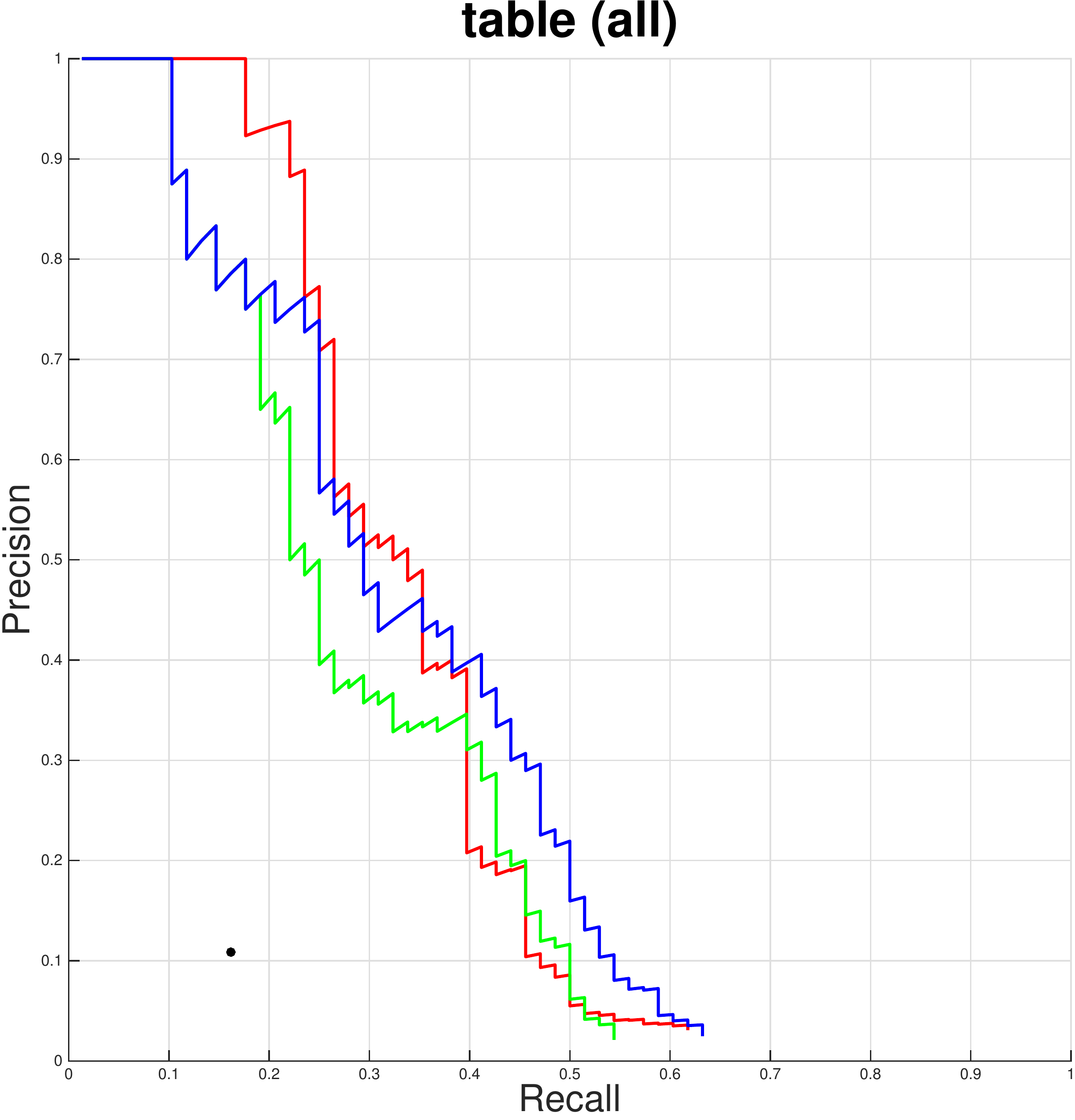} &
\insertA{0.18}{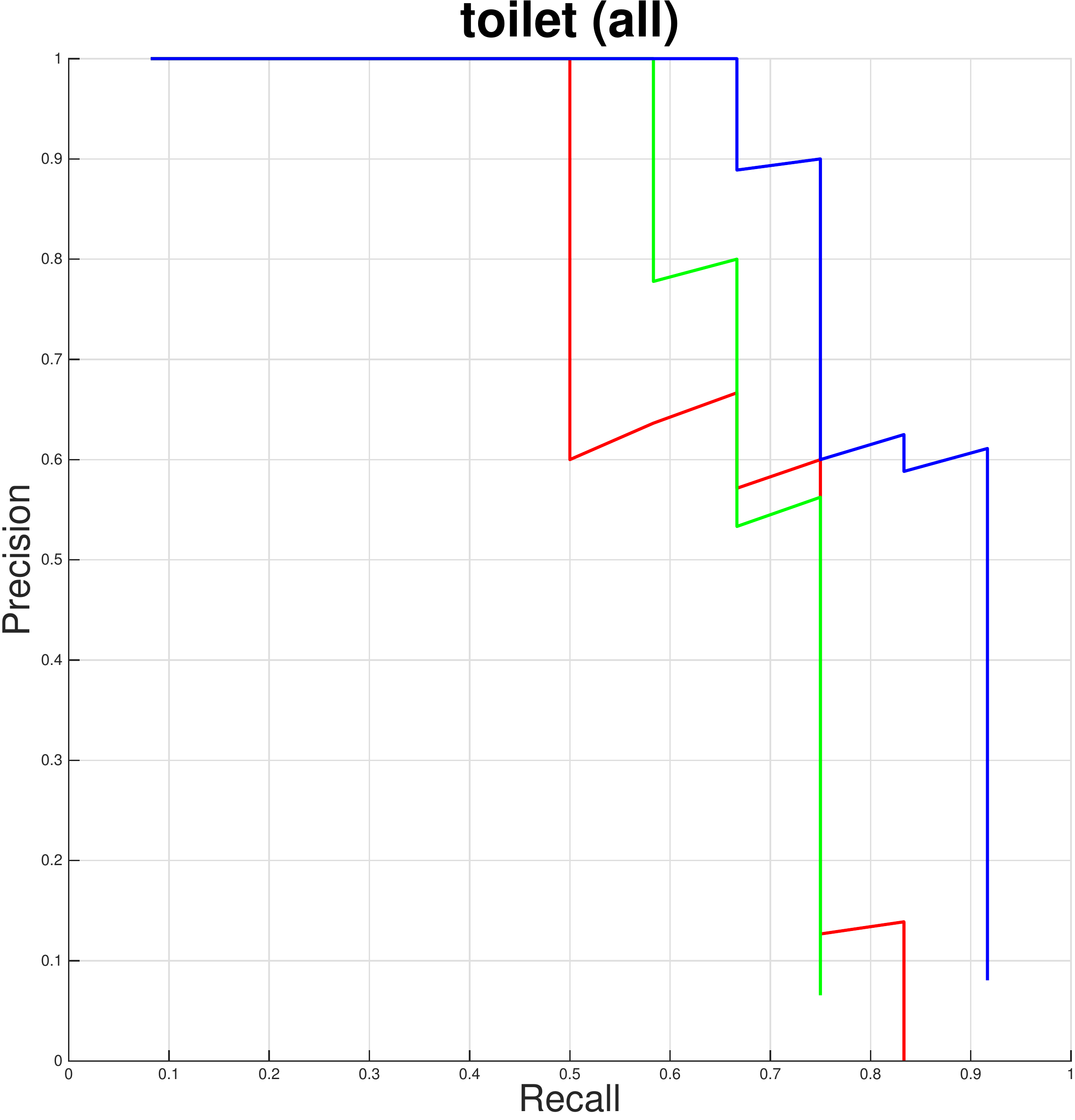} \\
\insertA{0.18}{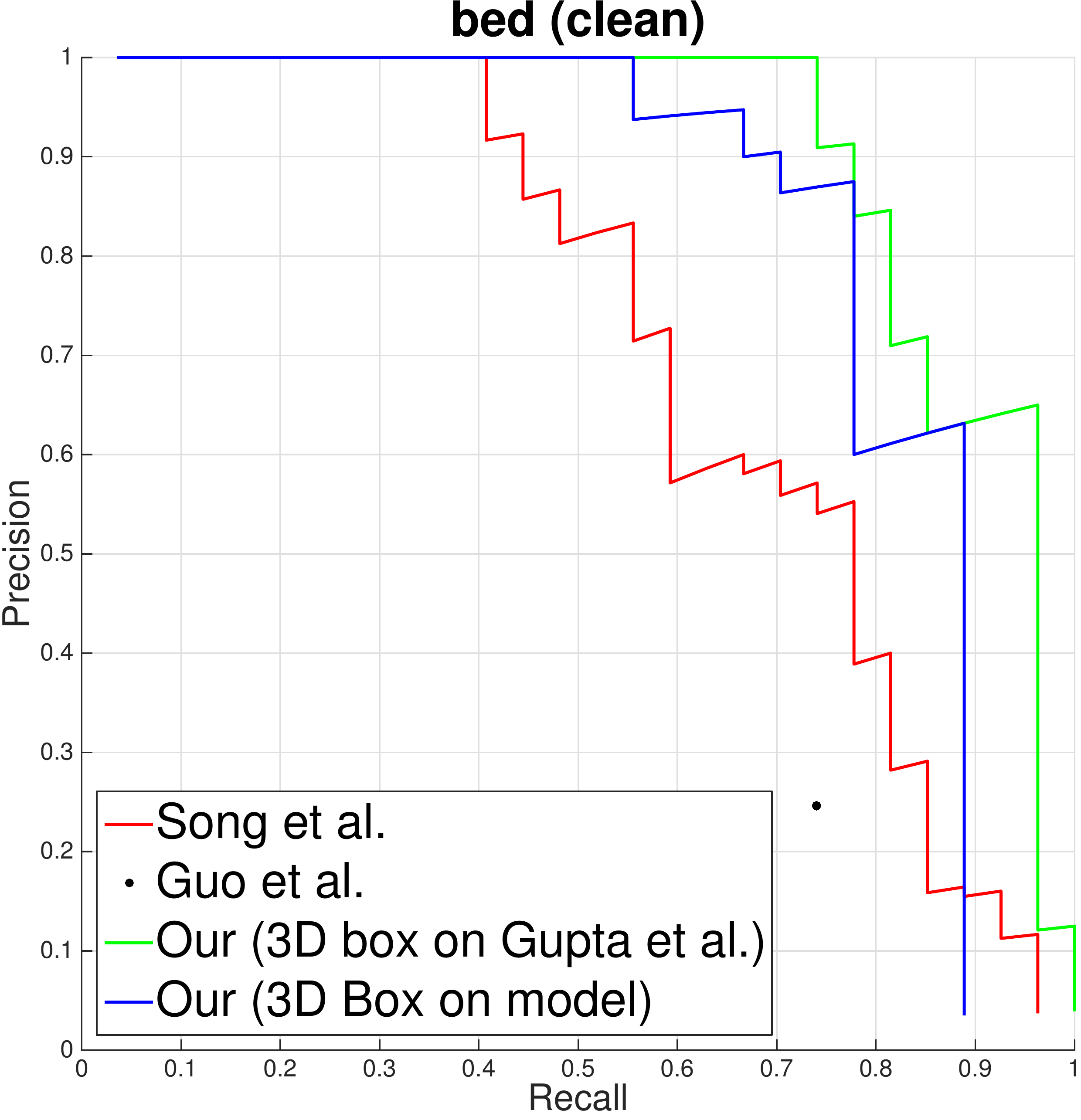} &
\insertA{0.18}{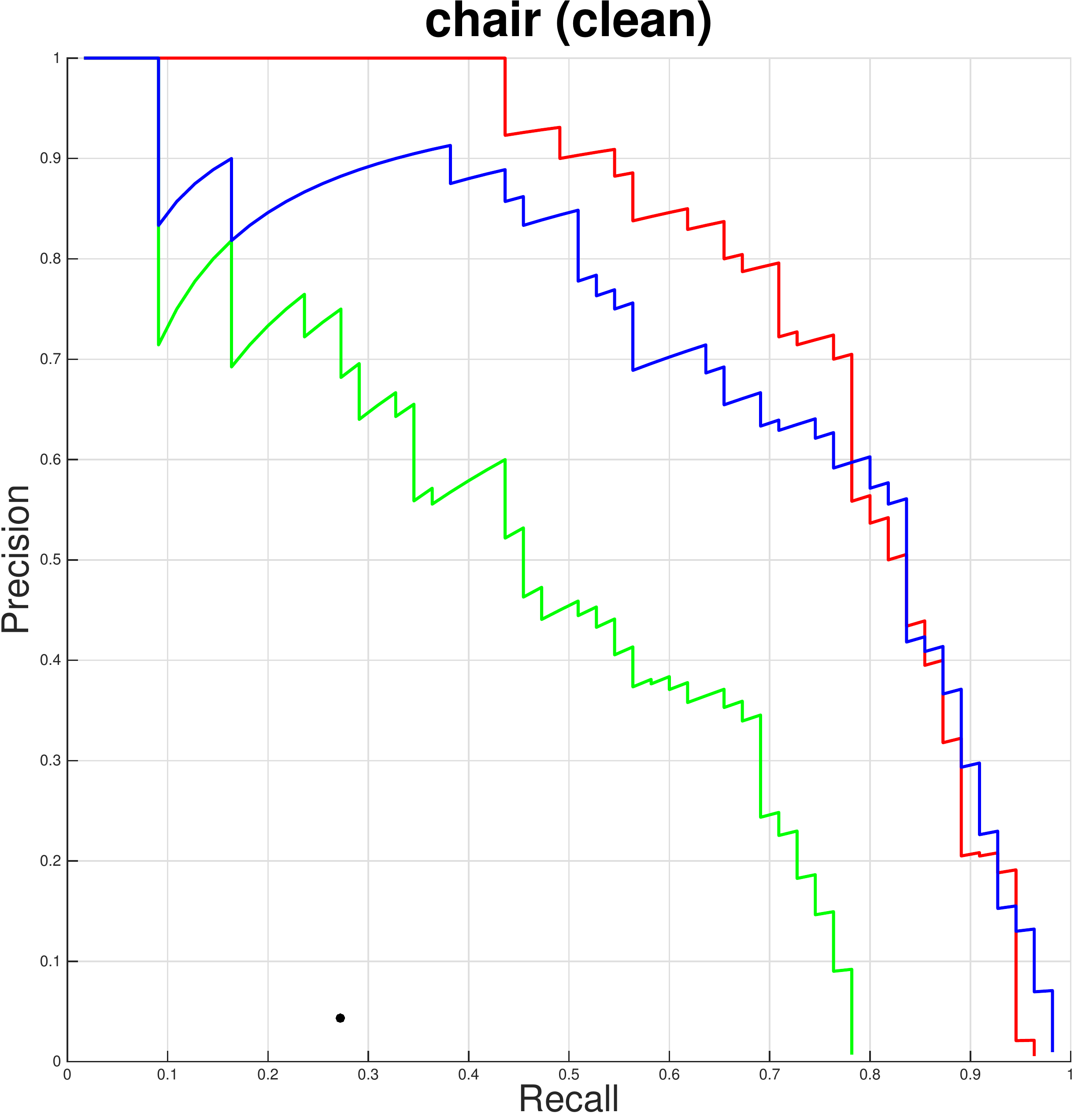} &
\insertA{0.18}{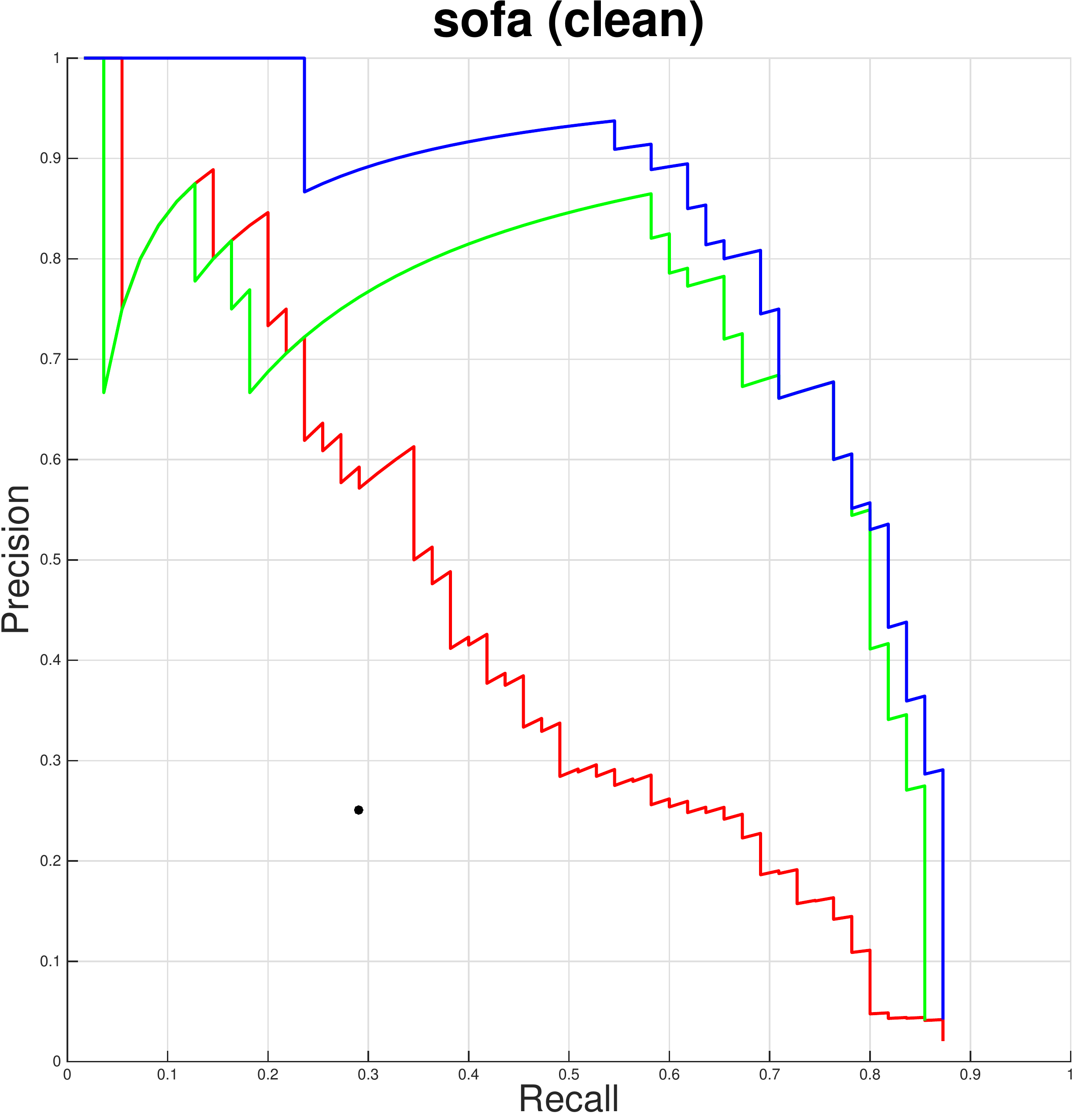} &
\insertA{0.18}{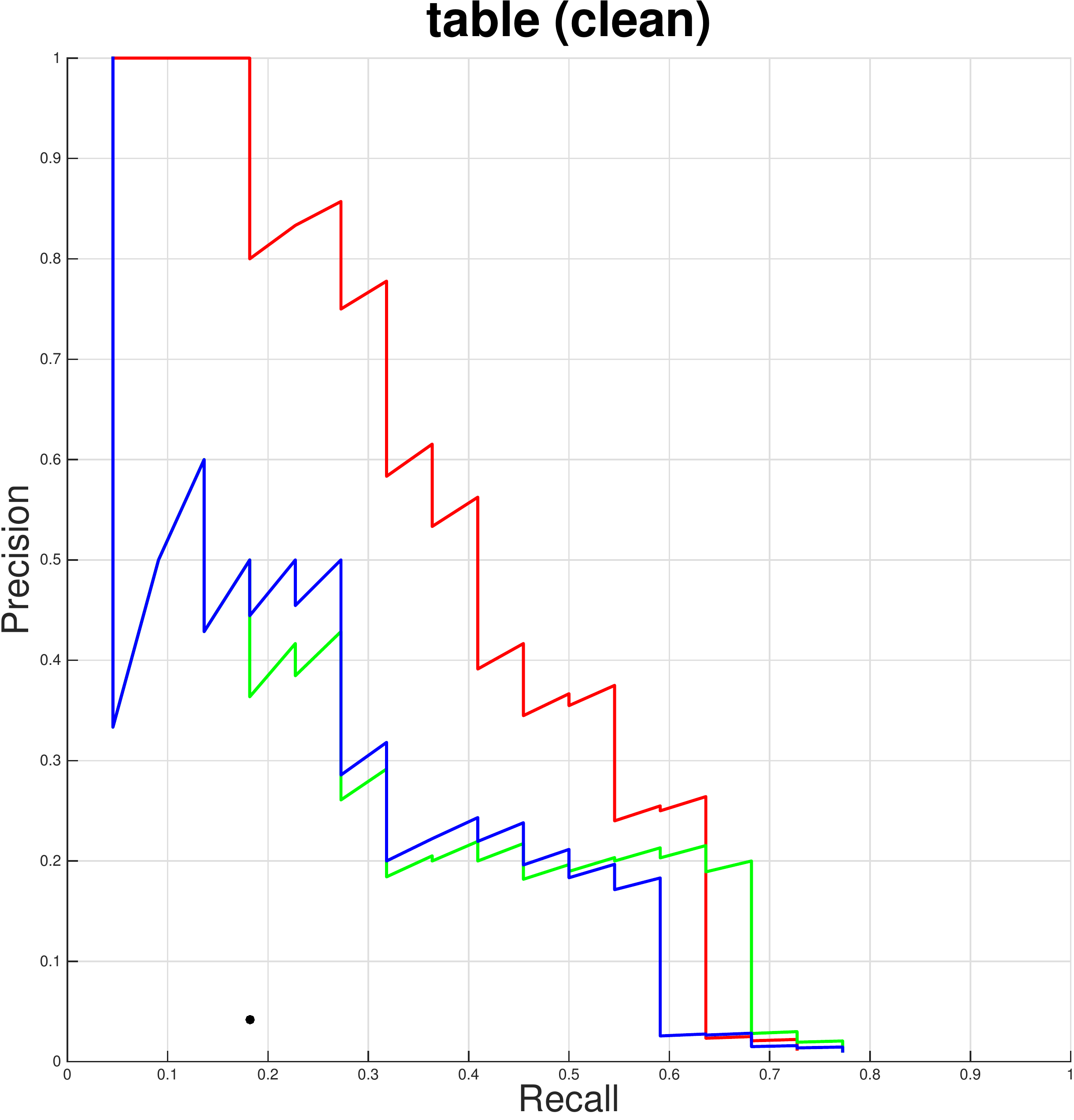} &
\insertA{0.18}{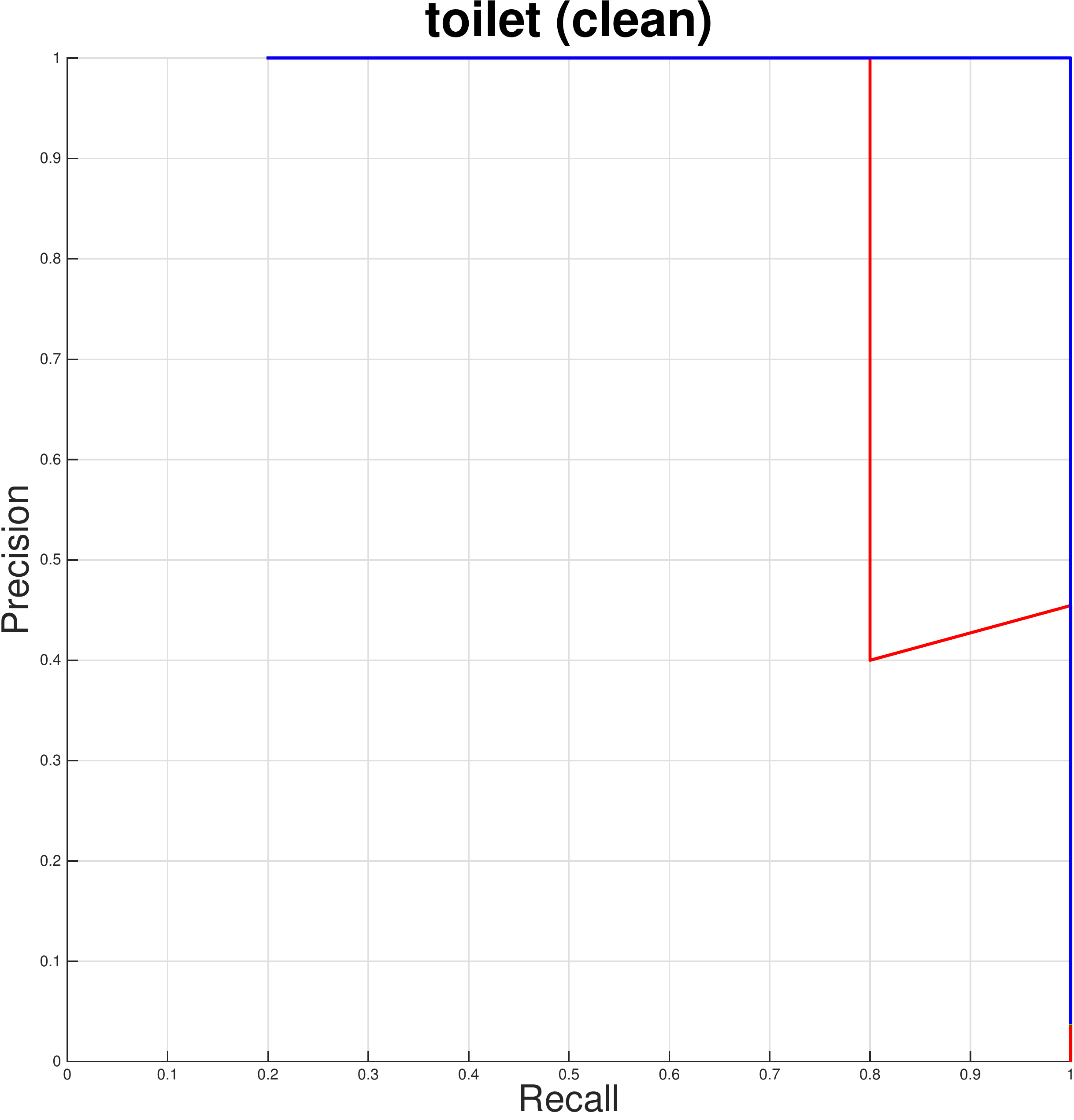}
\end{tabular}
\caption{\textbf{Precision Recall Plots for 3D Detection}: Variation in \modelAP as we change number of
scales, number of models and number of pose hypothesis we search over in our
model alignment stage.}
\figlabel{detection-3d-pr-plots}
\end{figure*}

\paragraph{Measuring Performance}
% \seclabel{coarse-pose-eval} 
To measure performance, we work with ground truth boxes, and consider the
distribution of the angular error in the top view. In particular, we plot the
angular error $\delta_{\theta}$ on the X-axis and the accuracy (the fraction of
data which incurs less than $\delta_{\theta}$ error) on the Y-axis. Note that we
plot this graph for small ranges of $\delta_{\theta}$ ($0^\circ$ to $45^\circ$)
as accuracy in the high error ranges is useless from the perspective of our
model alignment, where an initialization very far from the actual pose
often fails to align well. What is more desirable is a high
$\text{top}_{k}$ accuracy (fraction of instances which are within
$\delta_{\theta}$ of the $\text{top}_{k}$ predictions of the model). The
rationale is that the model alignment step can use multiple hypothesis and
pick the best among them. 

\paragraph{Evaluating on synthetic data} \figref{coarse-pose-synthetic} shows 
the performance on the synthetic data. This synthetic testing data is obtained
the same way as the synthetic training data, except that it comes from a
distinct set of models as compared to the training set. 
We experimented with the number of models in the training set and looking at the
error when considering $top_1$ and $top_2$ pose estimates and see expected
trends: more models help and that there is a large increase in recall when
considering two hypothesis as opposed to one.

\paragraph{Evaluating on real data} We now proceed to test our trained model on
real data. Here we work with the annotations from Guo and Hoiem
\cite{guoICCV13}. Guo and Hoiem annotate the \nyu dataset with 3D CAD models for
the following 6 categories: chair, bed, sofa, table, desk and book shelf. To
obtain interpretable results we work with categories which have a clearly
defined pose: chair, sofa and bed (we would have liked to also work with
bookshelf, but it is not among the 10 categories which are pose aligned in
ModelNet \cite{wuARXIV14}). In \figref{coarse-pose-real} we plot the same curves
as for synthetic data. The top row plots the $\text{top}_{1}$ accuracy and the
second row plots $\text{top}_{2}$ accuracy. Note that there is a large number of
objects which have missing depth data (for instance 30\% of chairs have more
than 50\% missing depth pixels), hence we plot these curves only for instances
with less than 50\% depth pixels missing. We compare against other
algorithms. We experimented with the HHA network from \cite{guptaECCV14} with
and without fine-tuning for this task, training a shallow network from random
initialization using HHA images and normal images. All these experiments are
done by training on the real data, and we see that we are able to outperform
these variants by training on clean synthetic data. 

\subsection{Model Fitting} 
An input to our model fitting procedure is an initial pixel support for the
object to fit to the model. We first describe and evaluate the instance
segmentation input we are using, describe how we can accurately lift 2D output
to 3D. To illustrate this we compare against \cite{songECCV14} and
\cite{guoThesis} for the task of 3D detection \cite{songECCV14}. Next, given
the lack of metrics to evaluate 3D model placement, we describe a metric and
the experiments that we used to make design choices for our model alignment
algorithm. Finally, we show examples of our output. 

\subsubsection{Object Detection and Instance Segmentation}
\seclabel{instance-segmentation} We note that the instance segmentation system
proposed in \cite{guptaECCV14} does not use the \cnn to compute features on the
bottom-up region but works with bounding boxes and later refines their support
using a random forest. We experimented with features computed on the masked
region in addition to features on the box as proposed by Hariharan \etal
\cite{hariharanECCV14}, and observe that these additional features improve
performance for bounding box detection as well as instance segmentation, thus
achieving state-of-the-art performance on these tasks
(\tableref{obj-det-inst-seg}). \boxAP goes up from 35.9\% to 39.3\%, \regionAP
improves from 32.1\% \cite{guptaECCV14} to 34.0\%. Moreover, Gupta \etal
\cite{guptaECCV14} only finetuned the model on 381 training image, \boxAP and
\regionAP both improve further when finetuning over the 795 trainval images
(row 4 and row 7 in \tableref{obj-det-inst-seg}). 

We work with these final instance segmentations for this work.  Of course, one
could refine these regions \cite{guptaECCV14,hariharanECCV14} to obtain even
better instance segmentations, but we chose to work with this output to
minimize the number of times we train on the same data. 

\subsubsection{3D Detection} 
\seclabel{detection-3d}
We next illustrate the richness of our approach by studying the task of 3D
detection.  Note that our method outputs a model aligned with objects in the
image. A trivial side-product of our output is a 3D bounding box (obtained by
putting a box around the inferred 3D model). We use this 3D bounding box as our
output for the 3D detection task and compare to the method from Song and Xiao
\cite{songECCV14} which was specifically designed and trained for this task. 

We study this task in the setting proposed by Song and Xiao in
\cite{songECCV14}. Song and Xiao work with the images from the \nyu dataset but
create different splits for different categories and study two different tasks:
a `clean' task where they remove instances which are heavily occluded or have
missing depth, and an `all' task in which they consider all instances. Given
they use non-standard splits which are different from the standard dataset
splits that we use, we simply evaluate on the intersection of their test set
for the category and the standard test set for the dataset. 

In addition we also compare to a simple baseline using the instance
segmentation from \cite{guptaECCV14} as described in
\secref{instance-segmentation} for 3D detection. We use a simple heuristic
here: we put a tight fitting box around the 3D points in the inferred instance
segmentation. We first determine the extent of the box in the top view by
searching over the orientation of the rectangular box such that its area is
minimized. We next set the bottom of the box to rest on the floor and estimate
the height as the maximum height of the points in the instance segmentation.
All these operations are done using percentiles ($\delta$ and $100 - \delta$,
with $\delta = 2$) to be robust to outliers. 

We report the performance in \tableref{detection-3d} and show the Precision
Recall curves in \figref{detection-3d-pr-plots}. We observe that this simple
strategy of fitting a box around the inferred instance segmentation (denoted as
`Our (3D Box on instance segmentation from Gupta \etal \cite{guptaECCV14})' in
\tableref{detection-3d}) already works better than the method proposed in
\cite{songECCV14} which was specifically designed for this task. At the same
time, this method is fast (40 seconds CPU + 30 seconds on a GPU) and scales
well with number of categories, as compared to 25 minutes per categories per
image for \cite{songECCV14}. This result illustrates that starting with
efficient and well established 2D reasoning (since \cite{guptaECCV14} does 2D
reasoning they are more readily able to leverage rich features for \rgb images) to
prune out large parts of the search space is not only more efficient but also
more accurate than 3D reasoning from the get go for such tasks.

Finally, a 3D box around our final output (denoted `Our (3D Box around
estimated model)') outperforms both \cite{songECCV14} and the baseline of
putting a 3D bounding box around the instance segmentation output, thus
illustrating the efficacy and utility of the methods proposed in the paper.
There is a large improvement over the baseline in performance for non-box like
objects, chair, sofa and toilet. The improvement for chair is particularly large
(18.6\% to 44.2\% in the `all' setting). This is because chairs are often
heavily occluded (\eg chair occluded behind a table) and the box around the
visible extent is systematically an underestimate of the actual amodal box. We
also note that our performance for tables is only comparable to \cite{songECCV14}. This
is because there is a mismatch between the definition of table as used by our
object detectors (which are based on ground truth from
\cite{silbermanECCV12,guptaECCV14}) and ground truth used in the 3D detection
benchmark from \cite{songECCV14} (the mismatch comes from coffee-tables and
desks being inconsistently marked as tables). 

Guo and Hoiem \cite{guoThesis} also aligns 3D CAD models to objects in the
image. We also compare to their work on this 3D detection task. We take the
scenes produced by the algorithm from \cite{guoThesis}, compute tight 3D
bounding boxes around their detected objects and benchmark them in the same
setup as described above to obtain a point on the Precision Recall plot as
shown in \figref{detection-3d-pr-plots} for categories that both works study:
bed, chair, table and sofa. This comparison is also largely favorable to our
method. 

Lastly, we also report performance of our system when only using the depth
image for object detection, pose estimation and model placement steps (denoted
`Our [no RGB] (3D Box on instance segmentation from Gupta et al.
\cite{guptaECCV14})' and `Our [no RGB] (3D Box around estimated model)') (the
bottom-up region generation step still uses the \rgb image, we do not expect
this to impact this result significantly). It is interesting to see that this
version of our system is better than the full version for some categories. We
believe this is because \rgb information allows our full system to detect
objects with missing depth with high scores which go on to become high scoring
false positives when the model placement step fails given the absence of enough
depth data. On average this ablated version of our system performs comparably
to our final system (57.6\% versus 58.5\% in the `3D all' setting), and
continues to outperform the algorithm from Song and Xiao \cite{songECCV14}.
This is consistent with the observation that \regionAP for these two systems
across these five categories is also fairly similar (mean \regionAP without RGB
features across these five categories: 44.5\% compared to average \regionAP
full system: 47.4\%).

\renewcommand{\arraystretch}{1.4}
\begin{table*}[t] 
\parbox{.6\linewidth}{
\begin{center}
  \setlength{\tabcolsep}{4.75pt}
  \caption{\textbf{Control experiments for model placement on \nyu \textit{val}
  set}: We report the \modelAP for the three different setting we tested our
  algorithm on: using ground truth object segmentation masks, using latent
  positive segmentation masks and using the detection output from the instance
  segmentation from \cite{guptaECCV14}.  We report performance on two different
  values for threshold $t_{agree}$. See \secref{model-align-eval} for details.}
  \scalebox{0.70}{
    \begin{tabular}{c|cc|ccc|ccc}
      & \multicolumn{2}{c|}{ground truth segm} & \multicolumn{3}{c|}{latent positive
      setting} & \multicolumn{3}{c}{detection setting}\tabularnewline
      \hline 
           & 0.5, 5 & 0.5, 5 & 0.5, 5 & 0.5, 5 & \regionAP & 0.5, 5 & 0.5, 5 & \regionAP \tabularnewline
       $t_{agree}$ & 7 & $\infty$ & 7 & $\infty$ & upper & 7 & $\infty$ & upper\tabularnewline
           & & & & & bound & & & bound \tabularnewline
       \hline 
             bathtub &   57.4 &   76.8 &   55.3 &   83.3 &   94.7 &    6.7 &   19.4 &   25.7 \\
                 bed &   42.3 &   87.3 &   28.8 &   86.0 &   96.1 &   25.8 &   63.2 &   57.0 \\
               chair &   45.3 &   74.1 &   29.0 &   56.9 &   70.1 &   11.8 &   25.2 &   30.4 \\
                desk &   33.9 &   67.4 &   20.3 &   40.9 &   55.7 &    3.0 &    4.0 &    6.2 \\
             dresser &   82.7 &   92.0 &   76.1 &   96.0 &  100.0 &   13.3 &   21.1 &   21.1 \\
             monitor &   31.4 &   39.8 &   18.4 &   20.8 &   41.3 &   12.5 &   12.5 &   26.8 \\
         night-stand &   62.5 &   77.6 &   51.3 &   65.2 &   87.9 &   18.9 &   21.6 &   25.5 \\
                sofa &   45.1 &   85.0 &   28.5 &   72.0 &   92.4 &   10.5 &   30.4 &   37.7 \\
               table &   18.8 &   52.2 &   15.8 &   34.3 &   46.8 &    5.5 &   11.9 &   13.3 \\
              toilet &   66.0 &  100.0 &   46.0 &   86.0 &  100.0 &   35.9 &   72.4 &   73.2 \\
       \hline
       \textbf{mean} & \textbf{  48.5} & \textbf{  75.2} & \textbf{  37.0} & \textbf{  64.1} & \textbf{  78.5} & \textbf{  14.4} & \textbf{  28.2} & \textbf{  31.7} \\ 
     \end{tabular}
     \tablelabel{modelap}
  }
\end{center}} \qquad \qquad
\parbox{.3\linewidth}{
\begin{center}
  \setlength{\tabcolsep}{4.75pt}
  \caption{\textbf{{Results for model placement on \nyu \textit{test} set}}: We
  report the \modelAP in the detection setting. See \secref{model-align-eval}
  for details.} \vspace{1.2cm} 
  \scalebox{0.70}{
    \begin{tabular}{c|ccc}
    & \multicolumn{3}{c}{detection setting}\tabularnewline
    \hline 
         & 0.5, 5 & 0.5, 5 & \regionAP \tabularnewline
     $t_{agree}$ & 7 & $\infty$ & upper\tabularnewline
         & & & bound \tabularnewline
     \hline 
           bathtub &    7.9 &   50.4 &   42.0 \\
               bed &   31.8 &   68.7 &   65.0 \\
             chair &   14.7 &   35.6 &   42.9 \\
              desk &    4.1 &   10.8 &   12.0 \\
           dresser &   26.3 &   35.0 &   36.1 \\
           monitor &    5.7 &    7.4 &   11.4 \\
       night-stand &   28.1 &   33.7 &   34.8 \\
              sofa &   21.8 &   48.5 &   47.4 \\
             table &    5.6 &   12.3 &   15.0 \\
            toilet &   41.8 &   68.4 &   68.4 \\
     \hline
     \textbf{mean} & \textbf{  18.8} & \textbf{  37.1} & \textbf{  37.5} \\
     \end{tabular}
     \tablelabel{modelap-test}
  }
\end{center}}
\end{table*}

\begin{figure*}
\centering
\begin{tabular}{ccc}
\insertB{0.2}{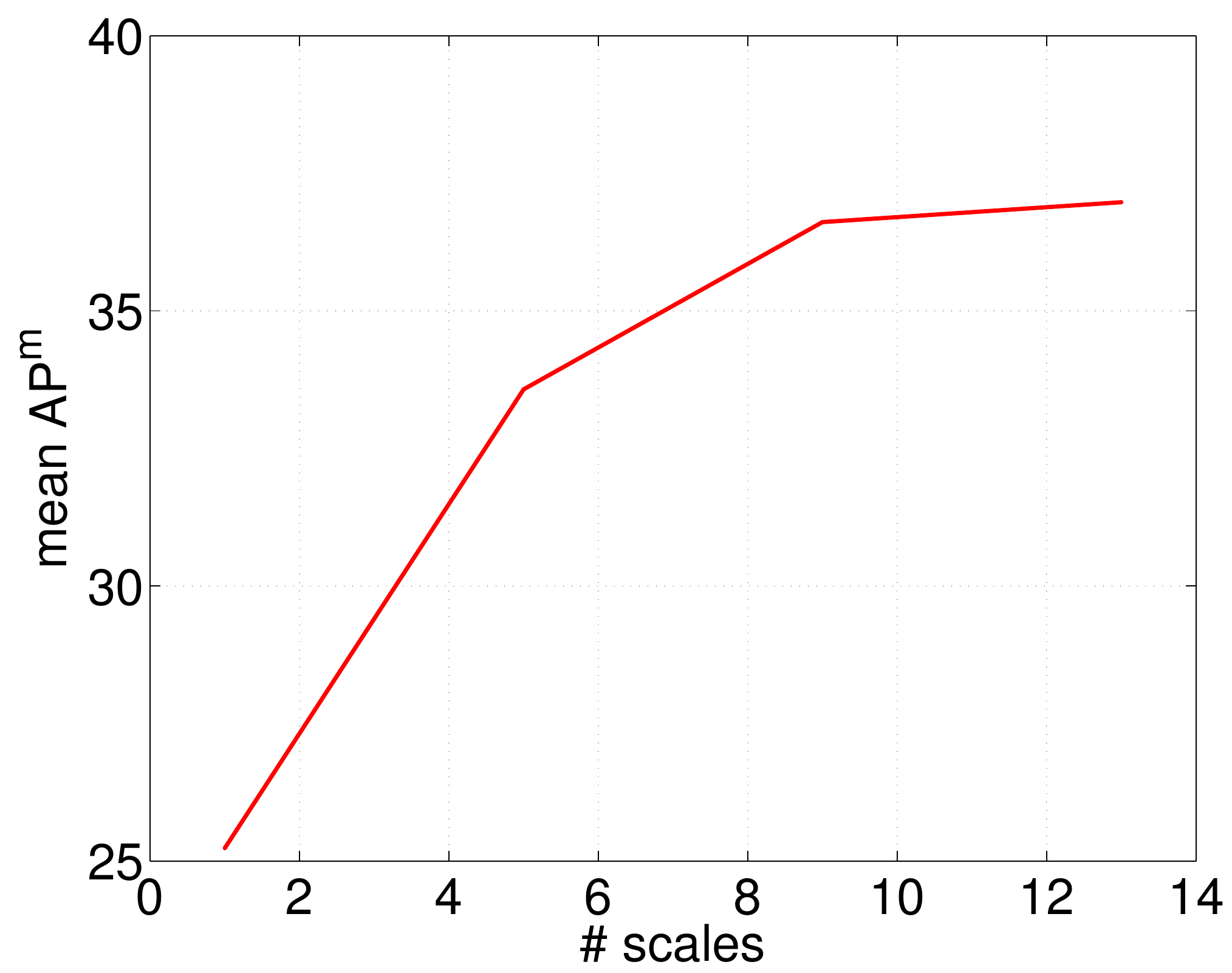} &
\insertB{0.2}{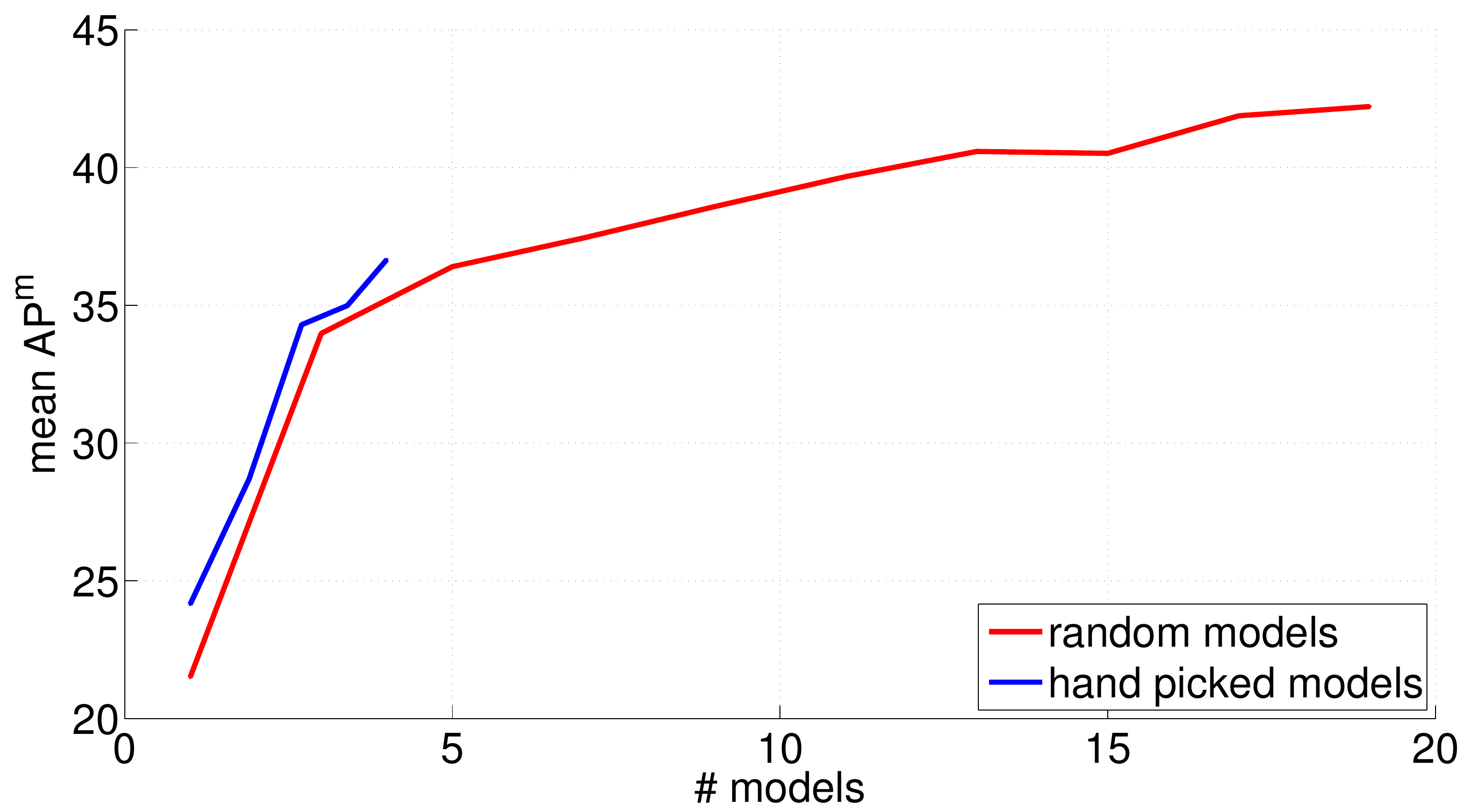} &
\insertB{0.2}{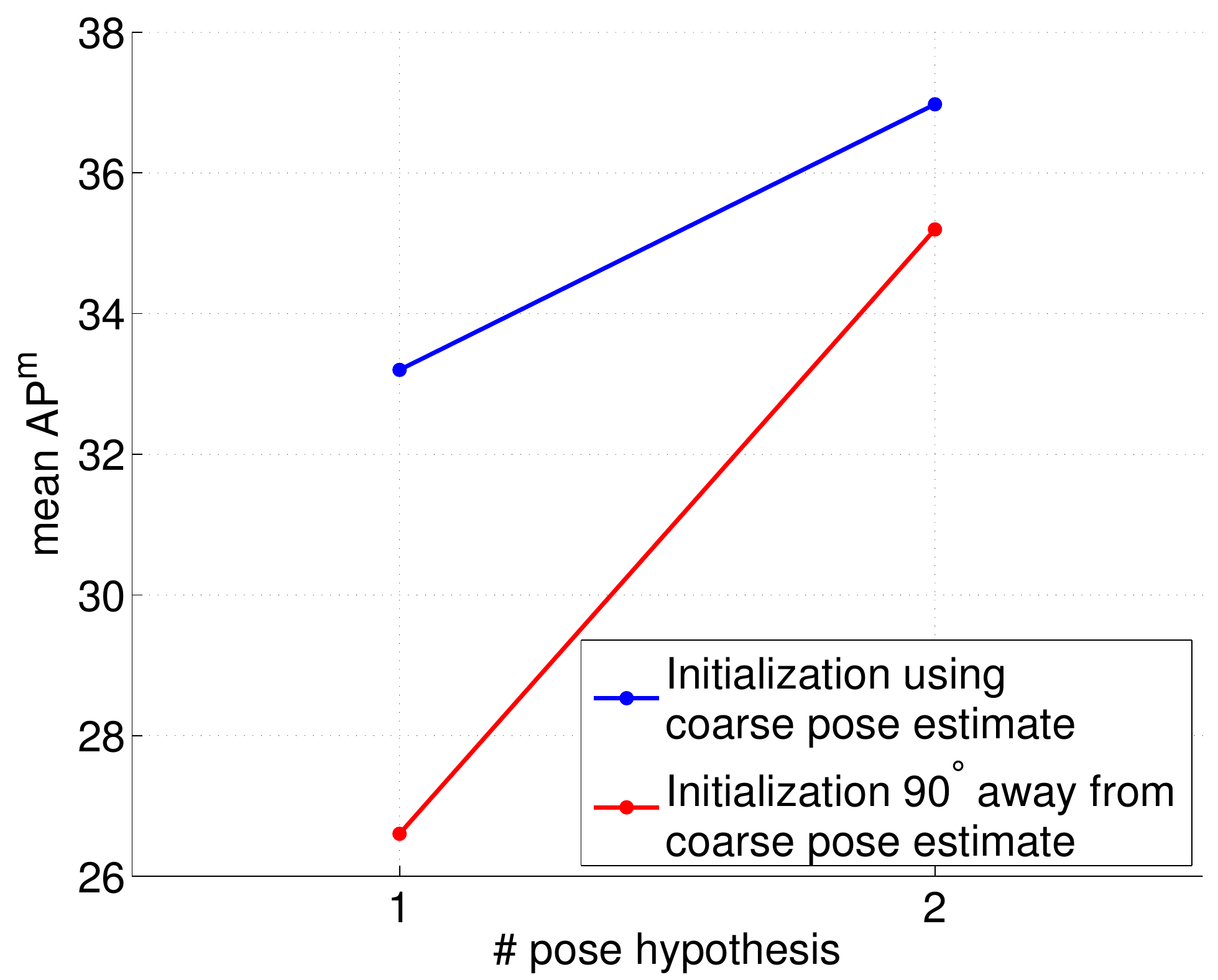} 
\end{tabular}
\caption{\textbf{Control Experiment}: Variation in \modelAP as we change number of
scales, number of models and number of pose hypothesis we search over in our
model alignment stage.}
\figlabel{control}
\end{figure*}

\subsubsection{Model Alignment Performance}
\seclabel{model-align-eval}
\paragraph{Measuring Performance} 
Given that the output of our algorithm is a 3D model placed in the scene, it is not
immediately obvious how to evaluate performance. One might think of evaluating
individual tasks such as pose estimation, sub-type classification, key point
prediction or instance segmentation, but doing these independently does not
measure the performance of the task we are considering, which is beyond each of
these individual tasks. Moreover, for many categories we are considering there may
not be a consistent definition of pose (\eg table), or key points (\eg sofa), or
sub-types (\eg chair). 

Thus, to surpass the limitation of these metrics to measure performance at our
task of placing 3D models in the scene, we propose a new metric which directly
evaluates the fit of the inferred model with the observed depth image. We assume
that there is a fixed library of 3D models $\mathcal{L}$, and a given algorithm
$\mathcal{A}$ has to pick one of these models, and place it appropriately in the
scene. We assume we have category level instance segmentations annotations for
the categories we are studying.

Our proposed metric is a generalization of the Average Precision, the standard
metric for evaluating detection and segmentation \cite{hariharanECCV14}.
Instead of just using the image level intersection over union of the predicted
box (in case of \boxAP) or region (in case of \regionAP) with the ground truth,
we also enforce the constraint that the prediction must agree with the
depth values observed in the image. In particular, we modify the way
intersection between a prediction and a ground truth instance in computed. We
render the model from the library $\mathcal{L}$ as proposed by the algorithm
$\mathcal{A}$ to obtain a depth map and a segmentation mask. We then do
occlusion checking with the given image to exclude pixels that are definitely
occluded (based on a threshold $t_{occlusion}$). This gives us the visible part
of the object $P_{visible}$. We then compute the intersection $I$ between the
output and the ground truth $G$ by counting the number of pixels which are
contained in both $P_{visible}$ and $G$, but in addition also agree on their
depth values by being within a distance threshold of $t_{agree}$ with each
other. Union $U$ is computed by counting the number of pixels in the ground
truth $G$ and the visible extent of the object $P_{visible}$ as $|G \cup
P_{visible}|$. If this $\frac{I}{U}$, is larger than $t_{iou}$ then this
prediction is considered to explain the data well, otherwise not. With this
modified definition of overlap, we plot a precision recall curve and measure the
area under it as measure of the performance of the algorithm $\mathcal{A}$. We
denote this average precision as \modelAP. To account for the inherent noise in
the sensor we operate with disparity as opposed to the depth value, and set
thresholds $t_{occluded}$ and $t_{agree}$ on disparity. Doing this allows for
larger error in far away objects as opposed to close by objects.  While this
behavior may not be desirable, it is unavoidable given the noise in the input
depth image behaves similarly. 

\paragraph{Evaluation} We evaluate our algorithm in 3 different settings: first
using ground truth segmentations, second using high scoring instance
segmentations from Gupta \etal \cite{guptaECCV14} that overlap with the ground
truth by more than 50\% (denoted as `latent positive setting'), and third a
completely unconstrained setting using only the instance segmentation output
without any ground truth (denoted as `detection setting'). \tableref{modelap}
summarizes results in these settings on the \textit{val} set. 

We use an $t_{iou}$ of 0.5 to count a true positive, $t_{occlusion}$ of 5
disparity units, and report performance at two different values of $t_{agree}$ 7
and $\infty$. An error of 7 disparity units corresponds to a 20 cm error at 3 meters.
A $t_{agree}$ of $\infty$ corresponds to \regionAP subject to the constraint that the
segmentation must come from the rendering of a 3D model.

We see that even when working with ground truth segmentations, estimating and
placing a 3D model to explain the segment is a hard task. We obtain a (model
average precision) \modelAP of 48.5\% in this setting. Even when evaluating at
$t_{agree}$ of $\infty$, we only get a performance of 75.2\% which is
indicative of the variety of our 3D model library and accuracy of our pose
estimator.

In the second setting, we take the highest scoring detection which
overlaps with more than 50\% with the ground truth mask. Note that this
setup decouples the performance of the detector from the performance of the model
placement algorithm while at the same time exposing the model placement
algorithm with noisier segmentation masks. Under
this setting, the \regionAP upper bound is 78.5\% which means that only as many
percentage of regions have a bottom-up region which overlaps with more than 0.5
with the ground truth mask, indicating the recall of the region
proposal generator that we are using \cite{guptaECCV14}. In this setting the
performance at $t_{agree} = \infty$ is 64.1\% and at $t_{agree} = 7$ is
37.0\%. This shows that our model alignment is fairly robust to
segmentation errors and we only see a small drop in performance from 48.5\% to
37.0\% when moving from ground truth setting to latent positive setting.

In the setting when we work with detections (using no ground truth information
at all), we observe a \regionAP upper bound of 31.7\% (which are comparable to
\regionAP reported in \tableref{obj-det-inst-seg} but slightly different
because a) we ignore pixels with missing depth values in computing this metric
and b) these are on the validation set). In this setting we observe a
performance of 14.4\% for $t_{agree}$ of 7 and 28.2\% for $t_{agree}$ of
$\infty$. We also report \modelAP on the \textit{test} set in the detection
setting in \tableref{modelap-test}. 

\paragraph{Control Experiments} We do additional control experiments to study
the affect of the number of scales, the number of models, difference in hand
picking models versus randomly picking models, number of pose hypothesis, and
the importance of initialization for the model alignment stage. These
experiments are summarized in \figref{control} and discussed below.

As expected, performance improves as we search over more scales (but saturates
around 10 scales) (\figref{control} left). The performance increases as we use
more models. Hand picking models so that they capture different modes of
variation is better than picking models randomly, and that performance does not
seem to saturate as we keep increasing the number of models we use during model
alignment step (\figref{control} center), although this comes at
proportionately larger computation time. Finally, using two pose hypothesis is
better than using a single hypothesis. The model alignment stage is indeed
sensitive to initialization and works better when used with the pose estimate
from \secref{coarse-pose}. This difference is more pronounced when using a single pose
hypothesis (33\% using our pose estimate versus 27\% when not using it, \figref{control} right).

\paragraph{Qualitative Visualizations} Finally, we provide qualitative
visualizations of the output of our method in \figref{output-vis} for chair,
bed, sofa, and toilets categories. We also show images where multiple objects have
been replaced with 3D models in \figref{output-vis-multiple}. 

\begin{figure*}
\centering
\renewcommand{\arraystretch}{0.5}
\setlength{\tabcolsep}{1.0pt}
\begin{tabular}{cc}
\insertA{0.49}{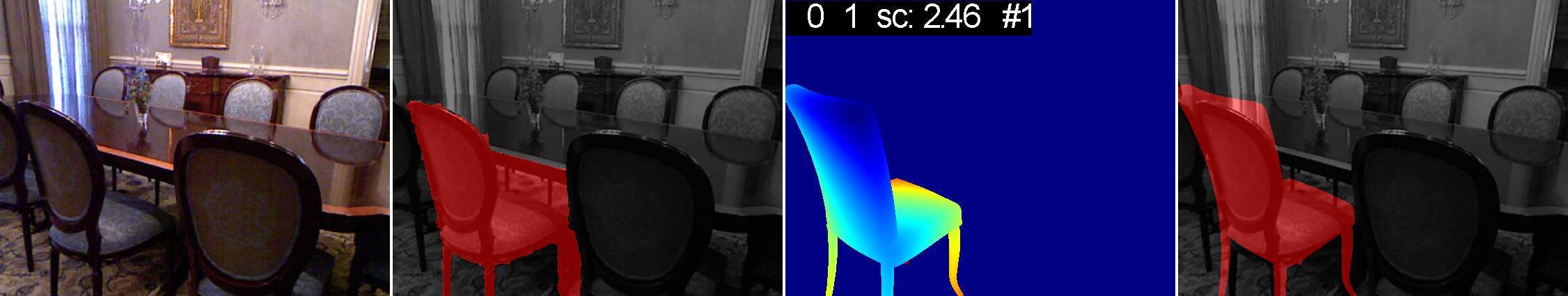} & \insertA{0.49}{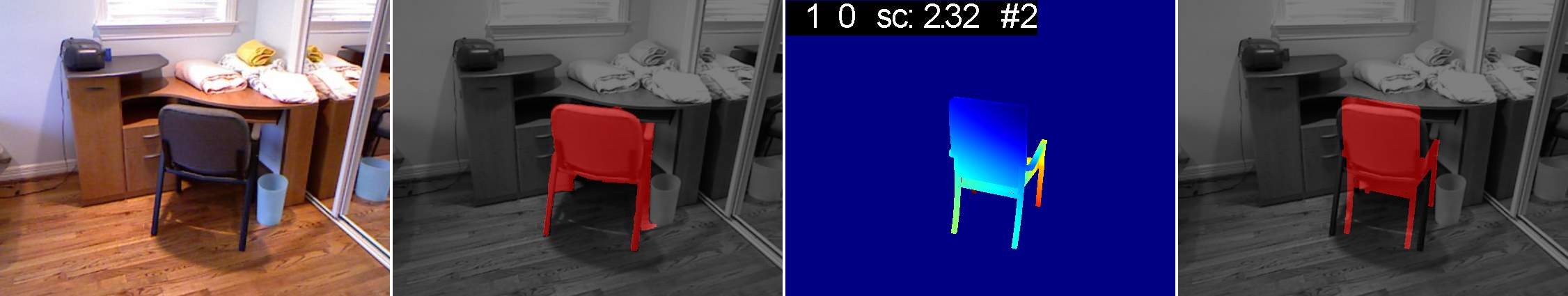} \\
\insertA{0.49}{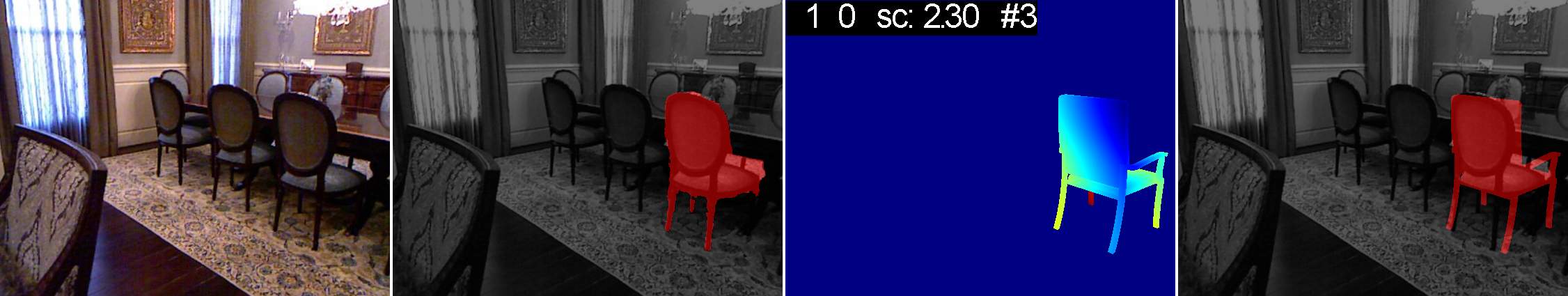} & \insertA{0.49}{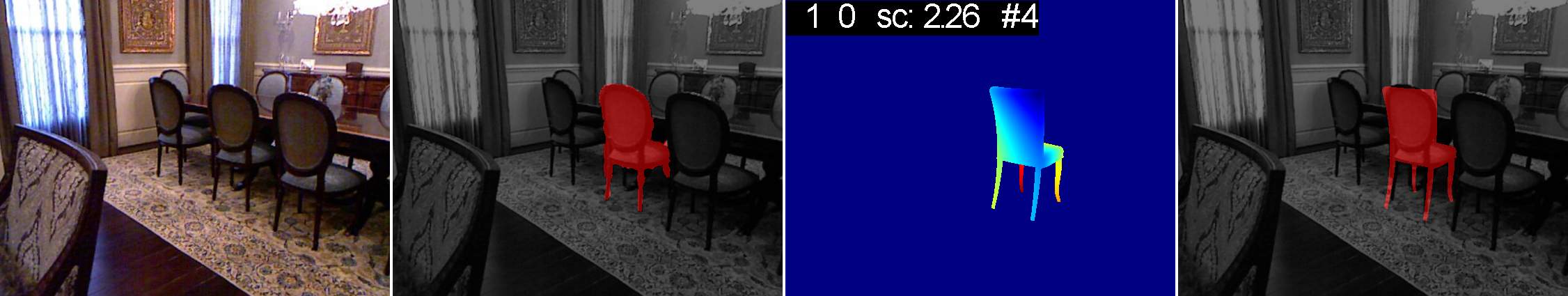} \\
\insertA{0.49}{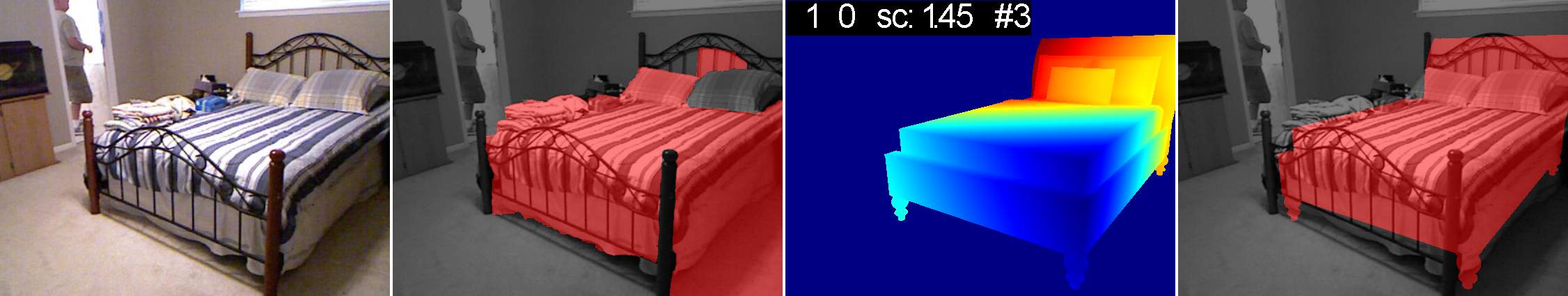} & \insertA{0.49}{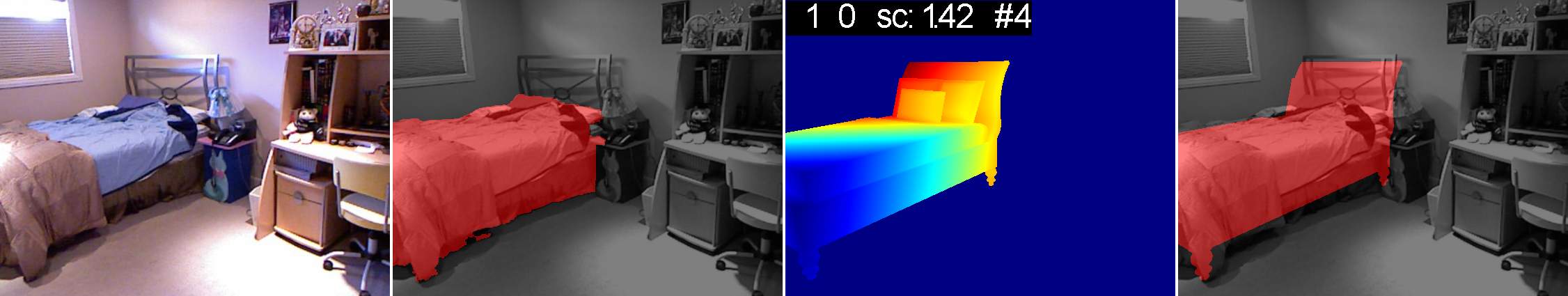} \\
\insertA{0.49}{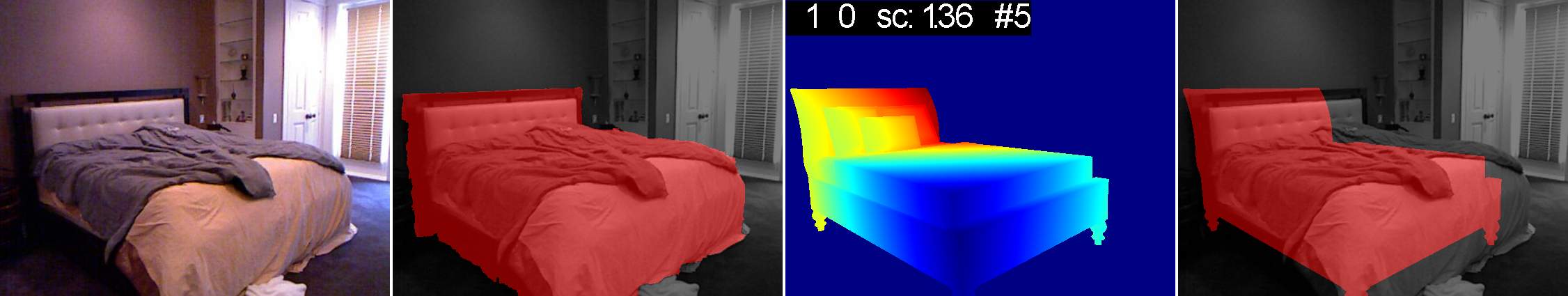} & \insertA{0.49}{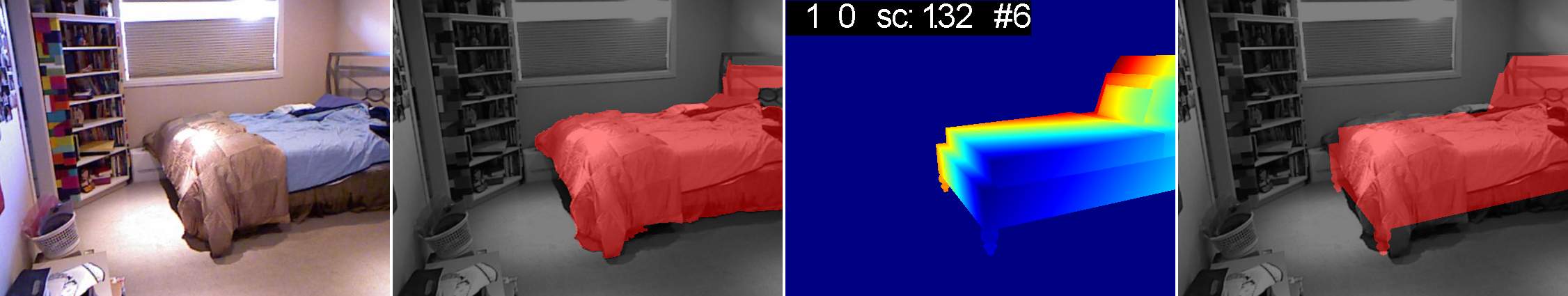} \\
\insertA{0.49}{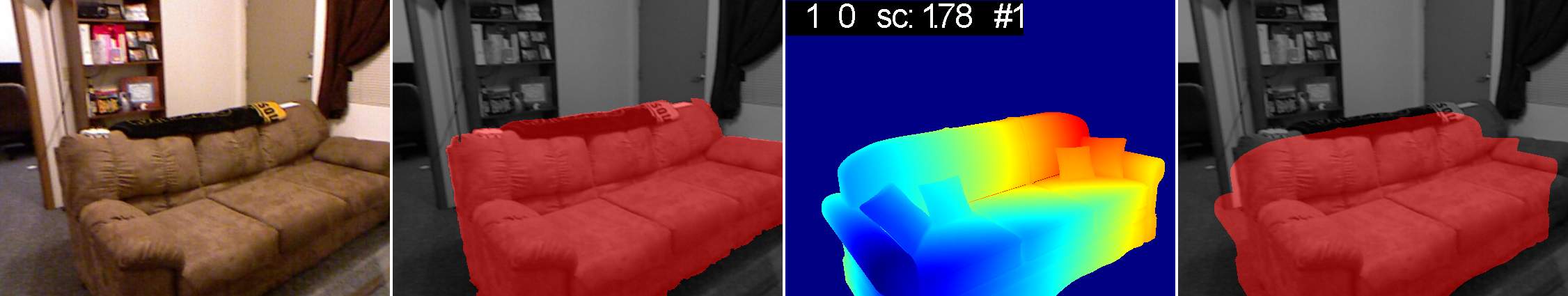} & \insertA{0.49}{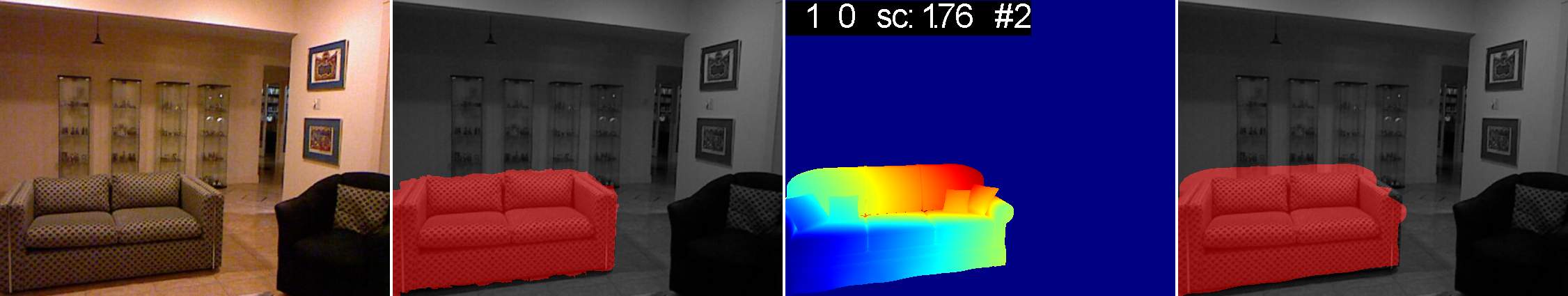} \\
\insertA{0.49}{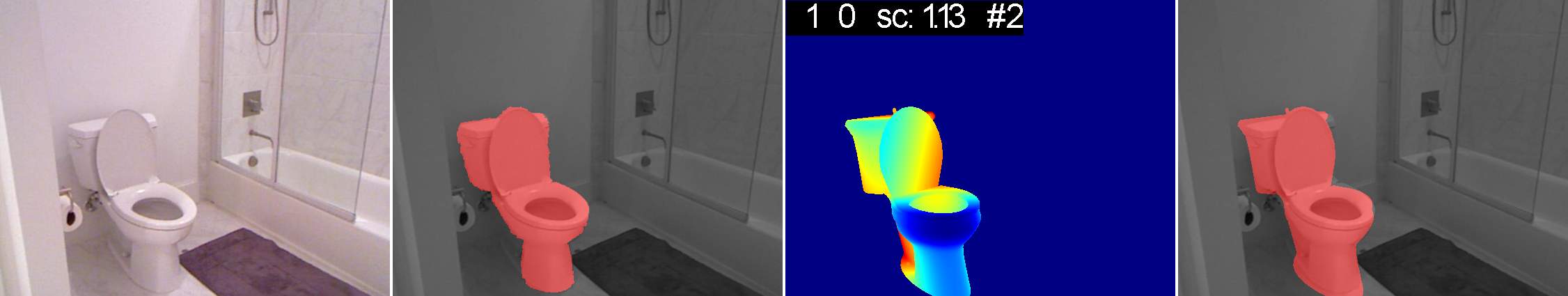} & \insertA{0.49}{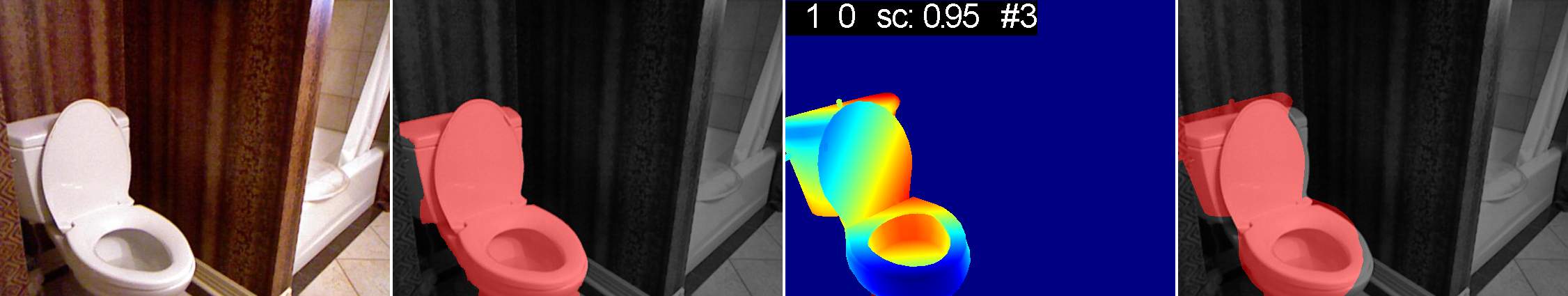} \\
\insertA{0.49}{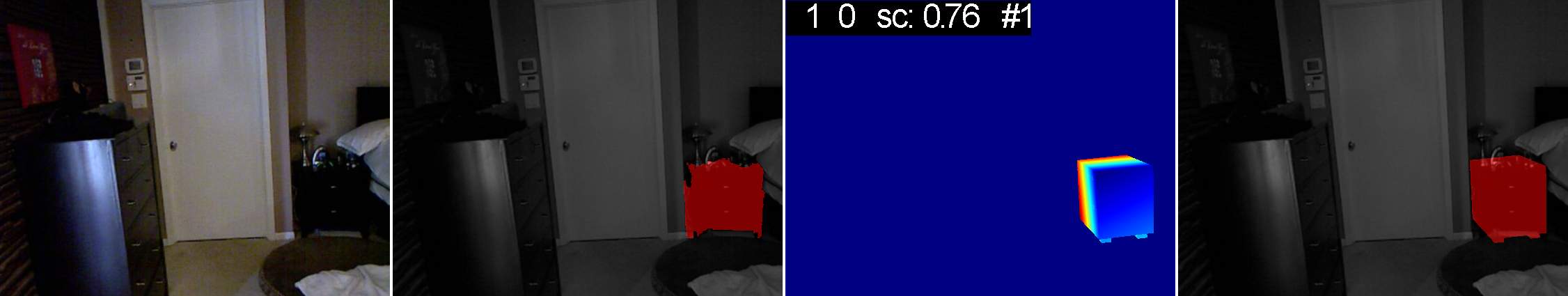} & \insertA{0.49}{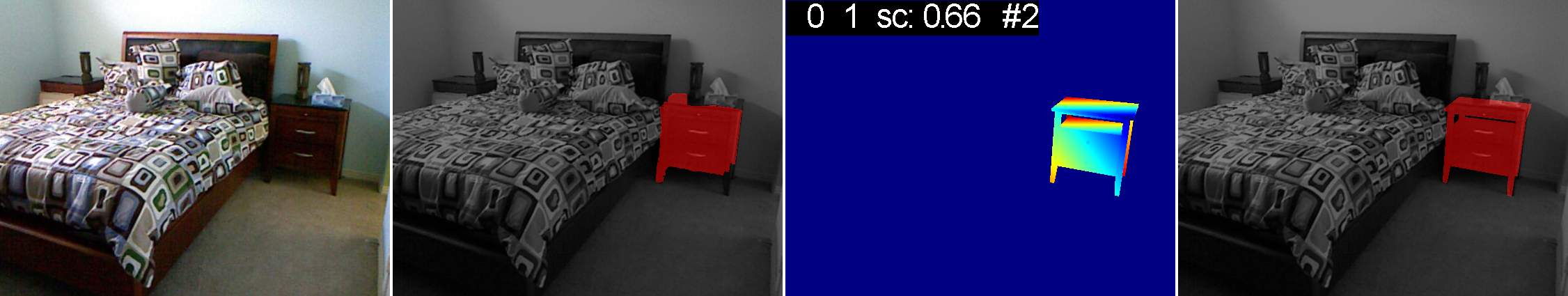} \\
\end{tabular}
\caption{\textbf{Visualizations of the output on the \textit{test} set}: We
show instance segmentation mask, rendered model, and rendered model overlaid on
image for high scoring detections for chairs, beds, sofas, toilets and
night-stands.} \figlabel{output-vis}
\end{figure*}

\begin{figure*}
\centering
\renewcommand{\arraystretch}{0.5}
\setlength{\tabcolsep}{1.0pt}
\begin{tabular}{ccc}
\insertA{0.330}{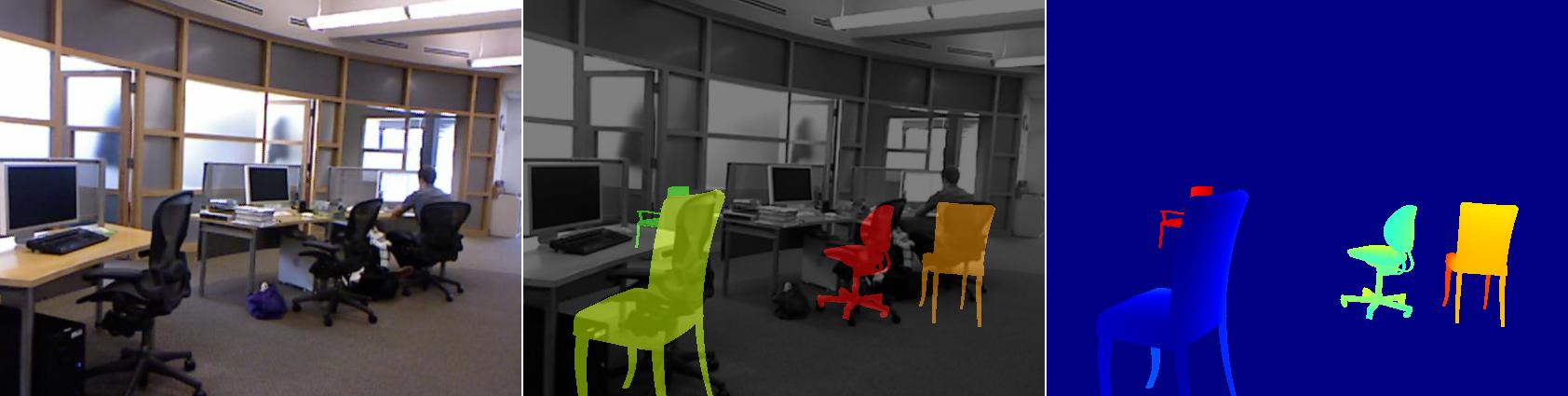} &
\insertA{0.330}{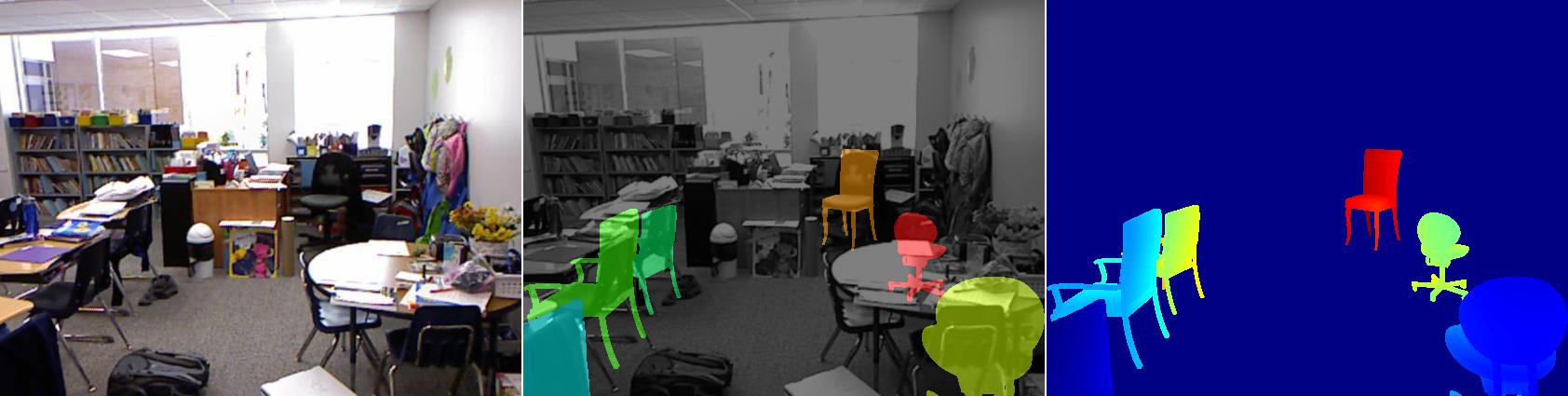} &
\insertA{0.330}{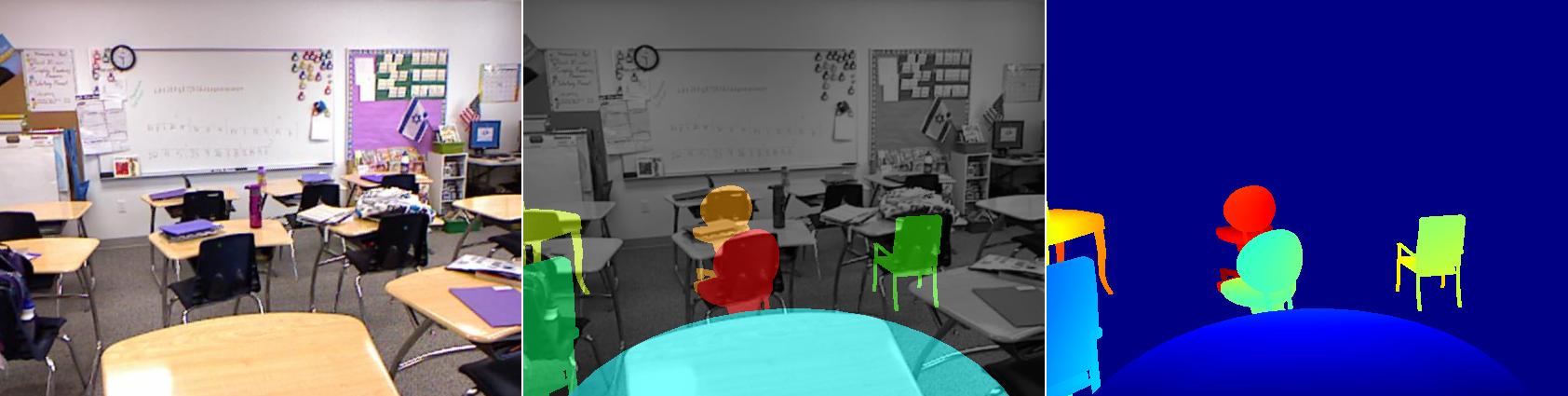} \\
\insertA{0.330}{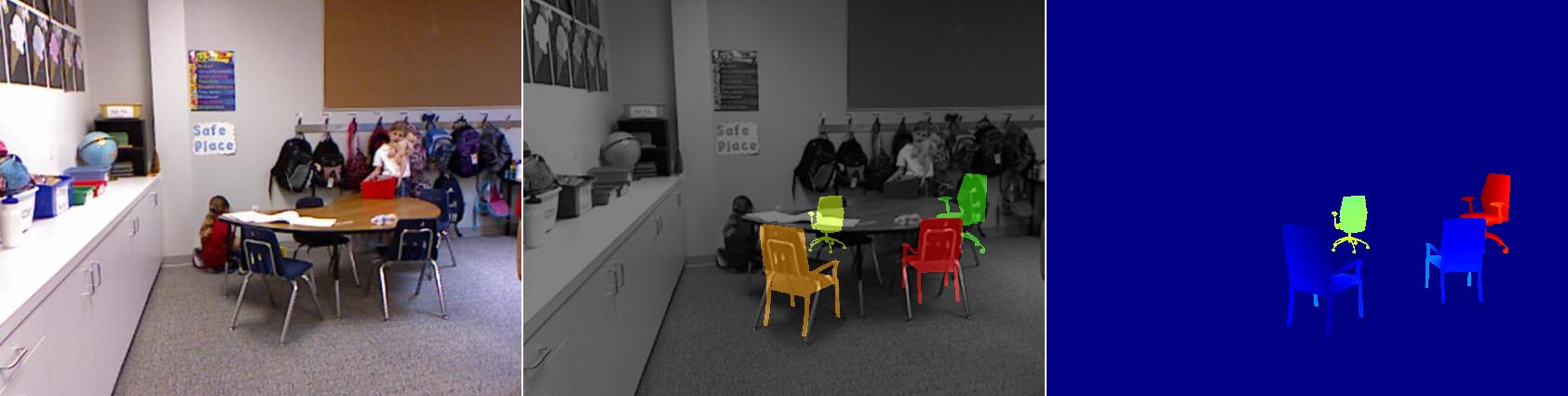} &
\insertA{0.330}{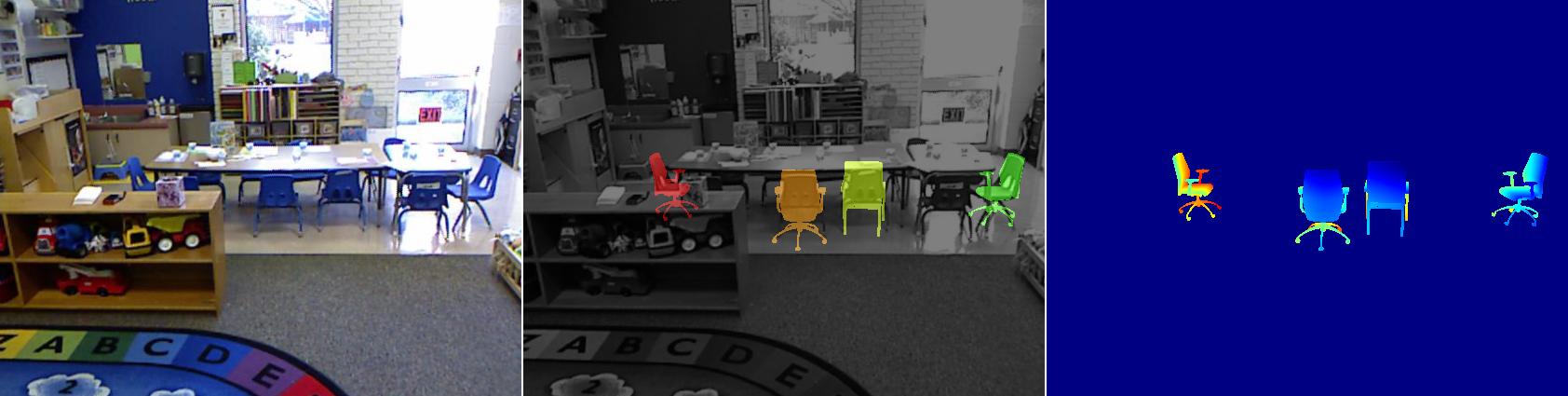} &
\insertA{0.330}{images-full-v3/img_5335.jpg} \\
\insertA{0.330}{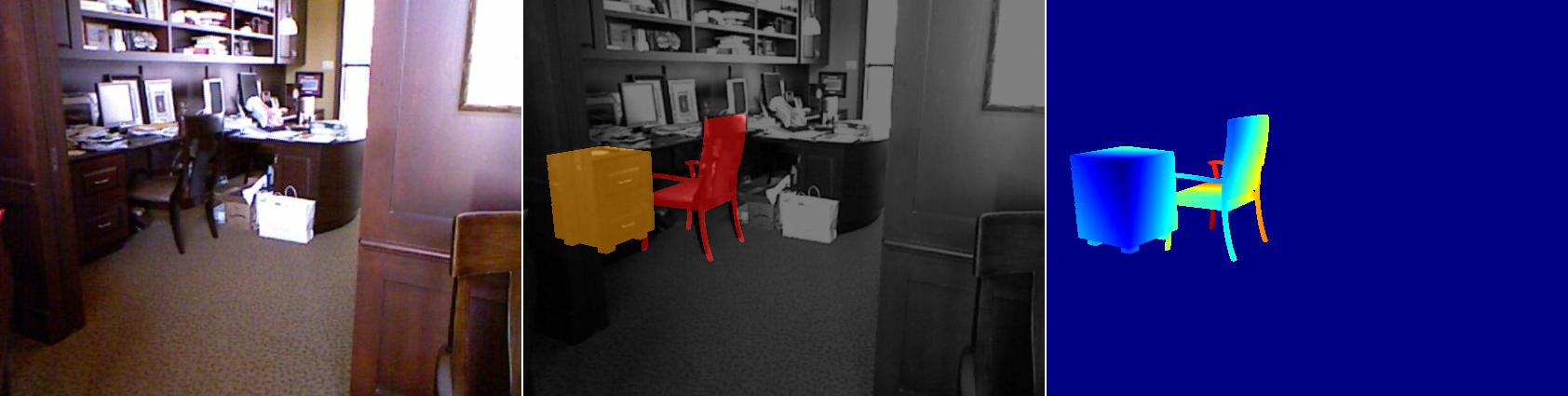} &
\insertA{0.330}{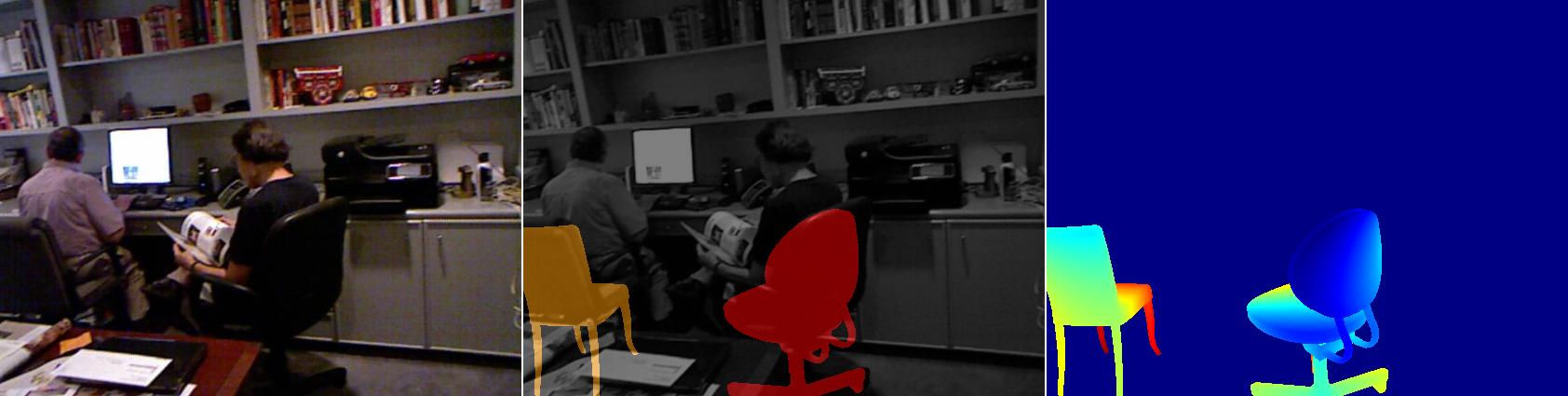} &
\insertA{0.330}{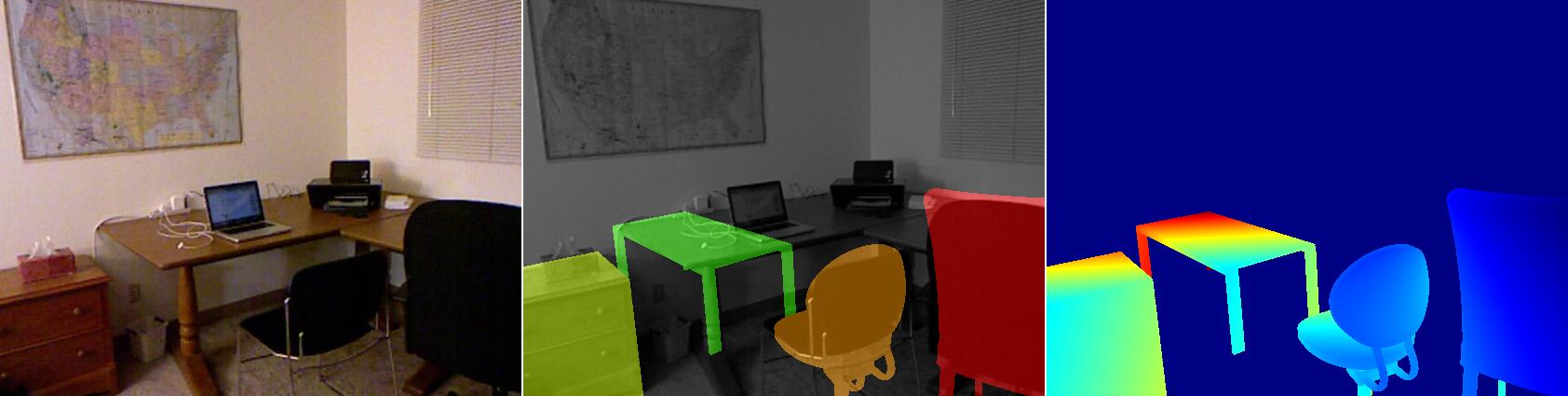} \\
\insertA{0.330}{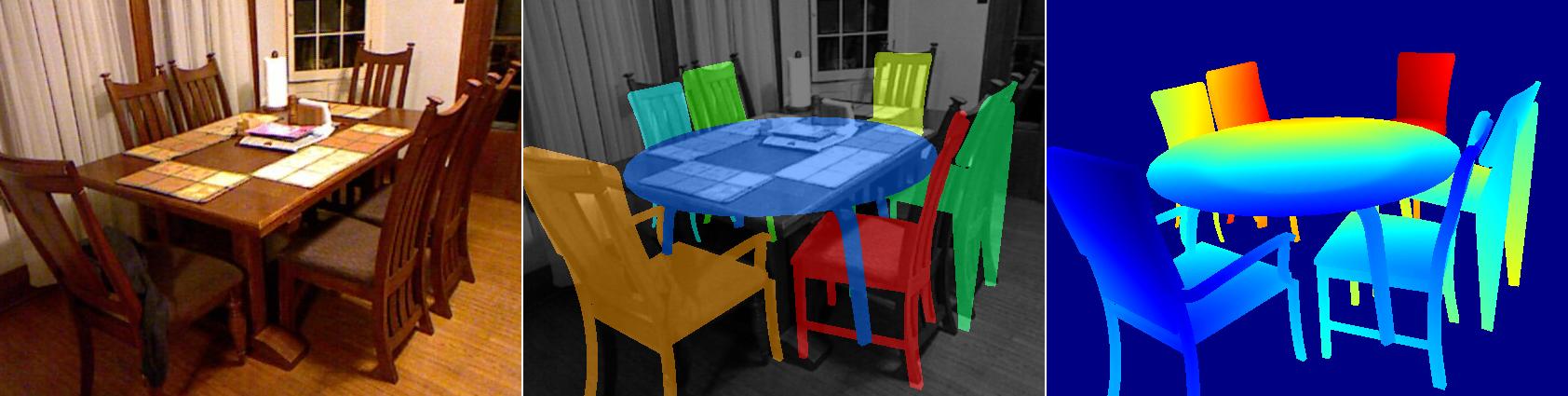} &
\insertA{0.330}{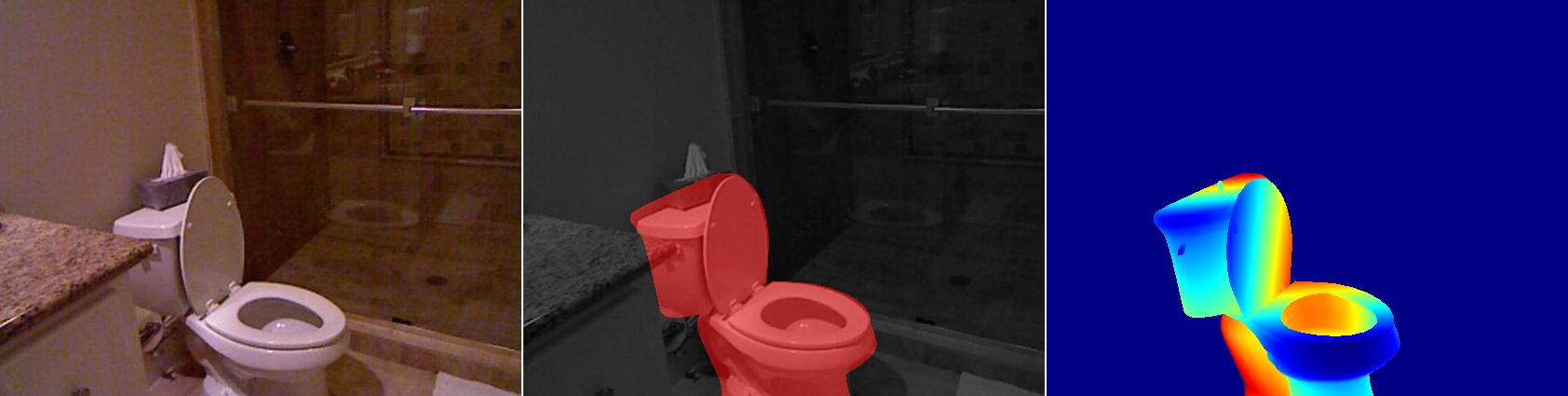} &
\insertA{0.330}{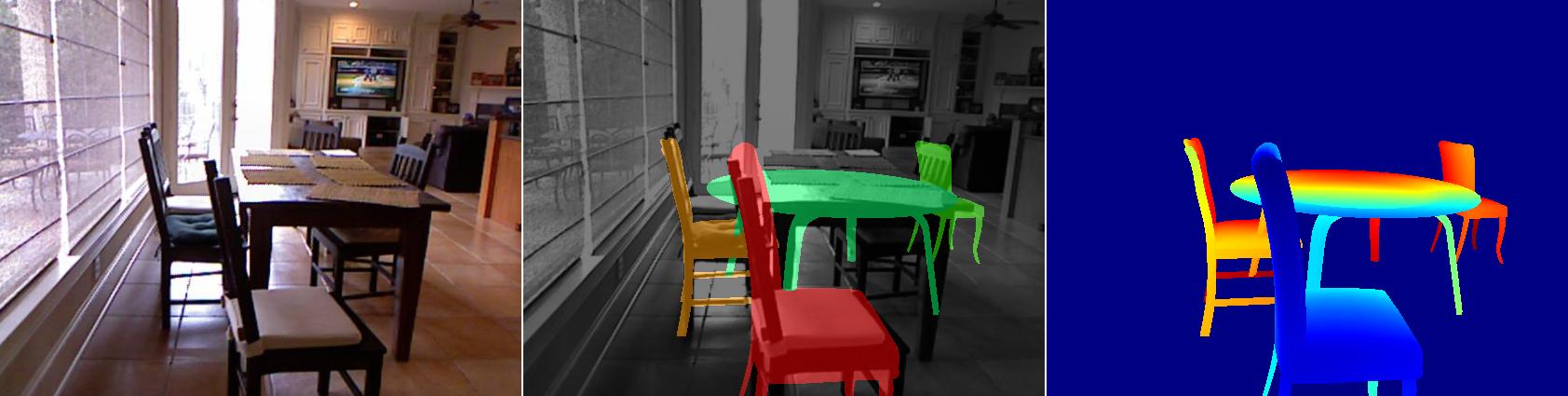} \\
\insertA{0.330}{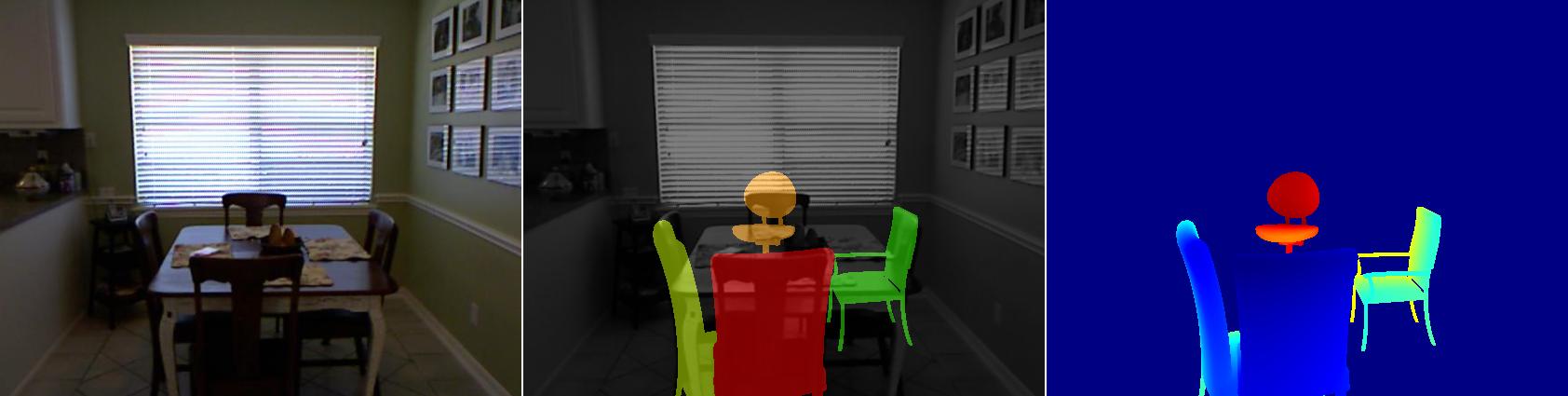} &
\insertA{0.330}{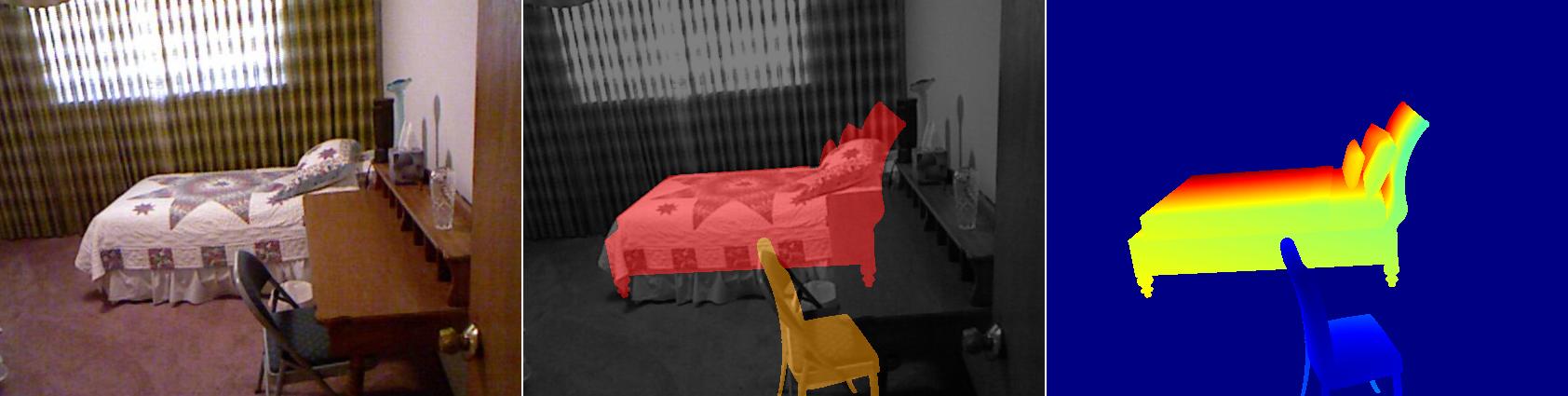} &
\insertA{0.330}{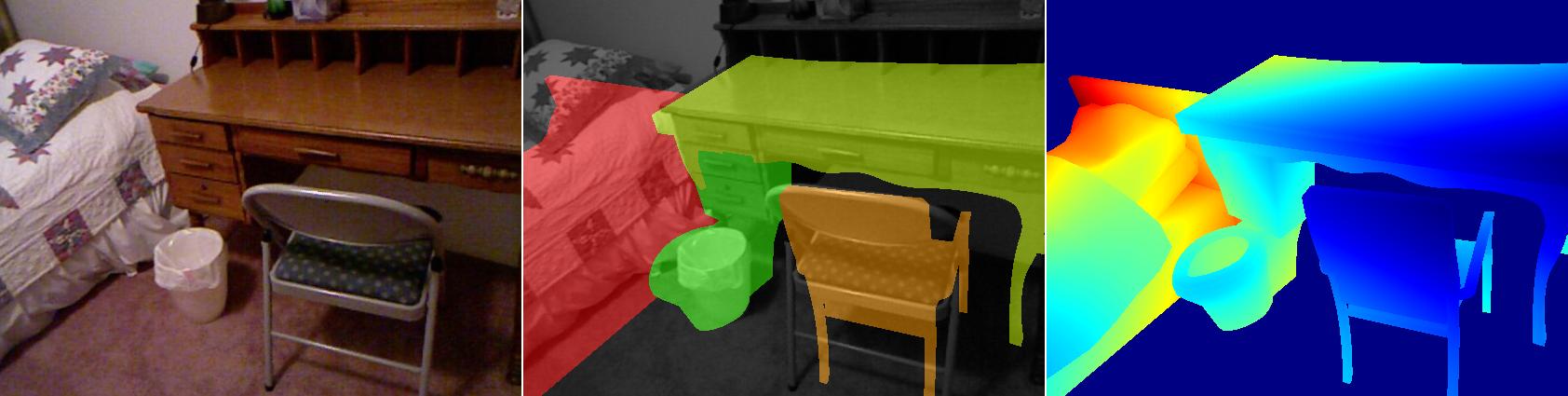} \\
\insertA{0.330}{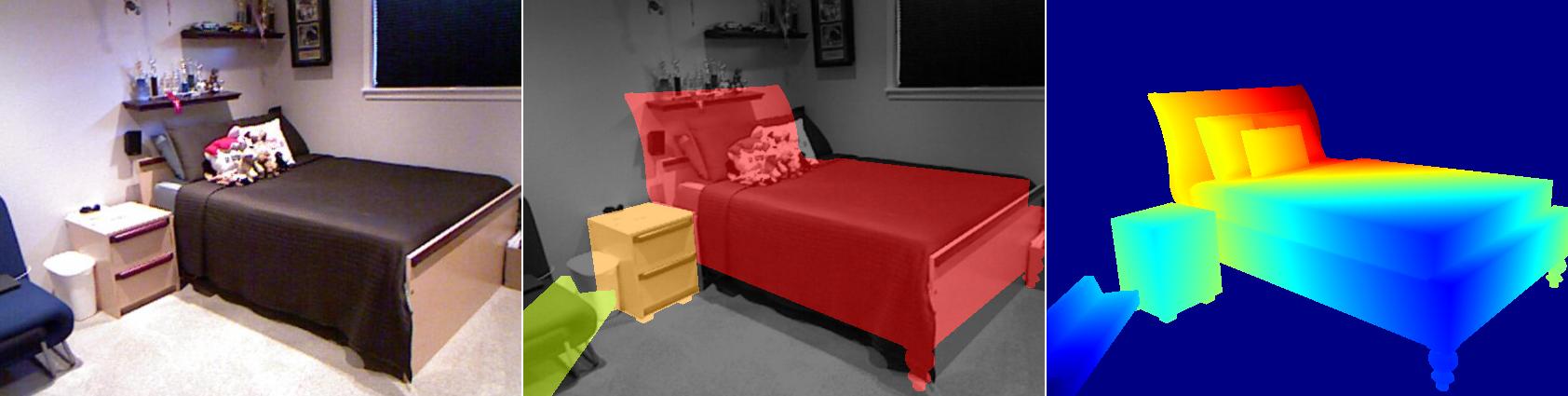} &
\insertA{0.330}{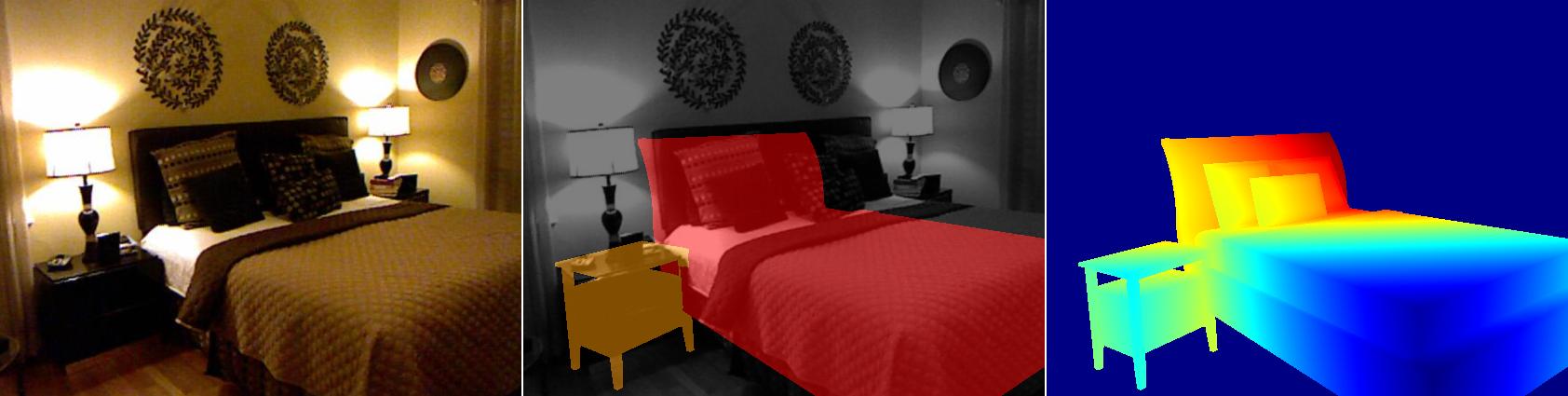} &
\insertA{0.330}{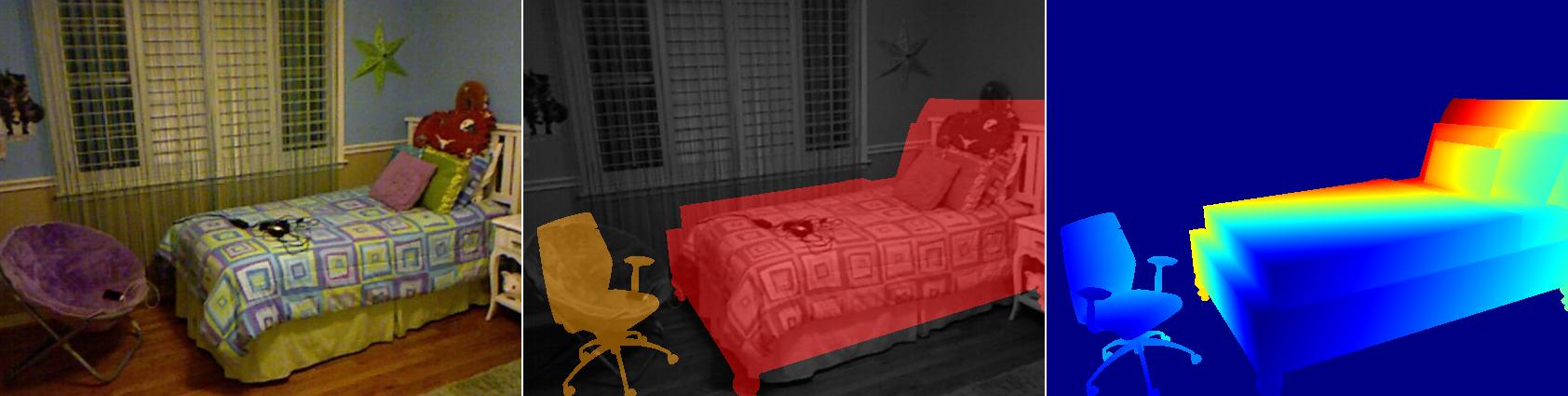} \\
\insertA{0.330}{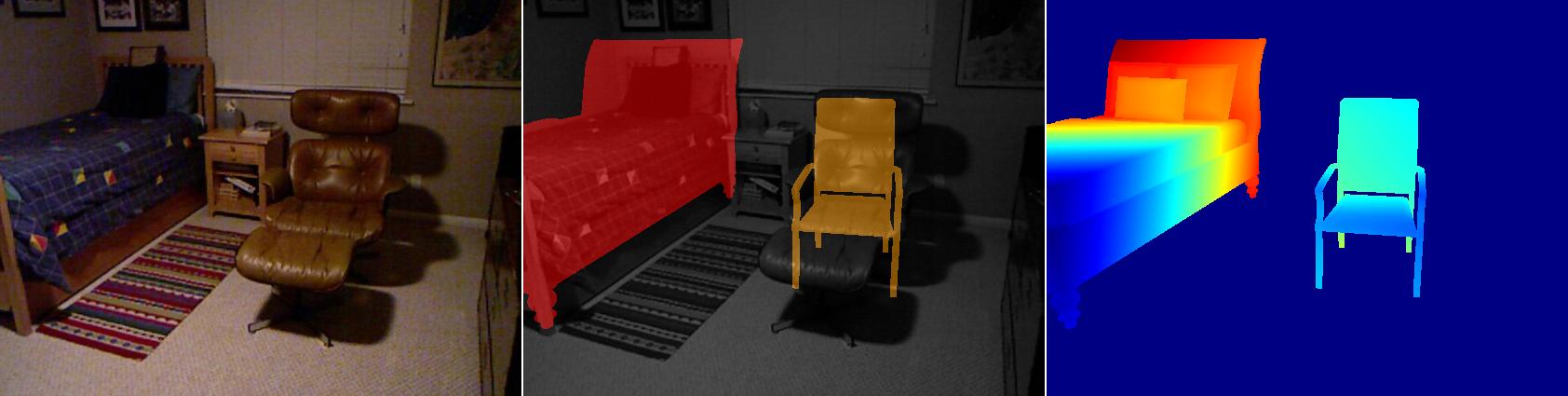} &
\insertA{0.330}{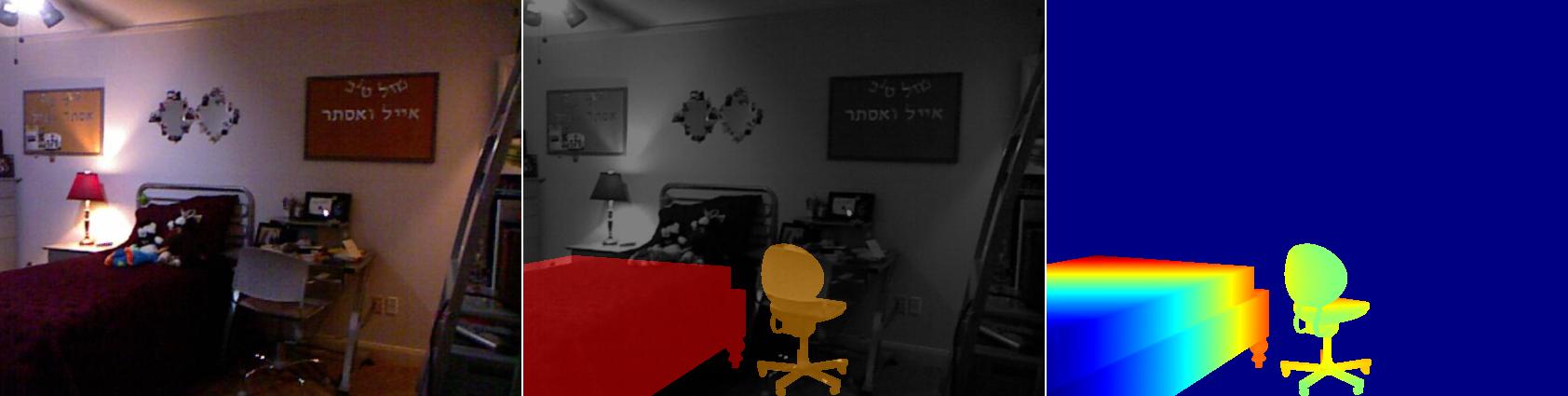} &
\insertA{0.330}{images-full-v3/img_6032.jpg} \\
\insertA{0.330}{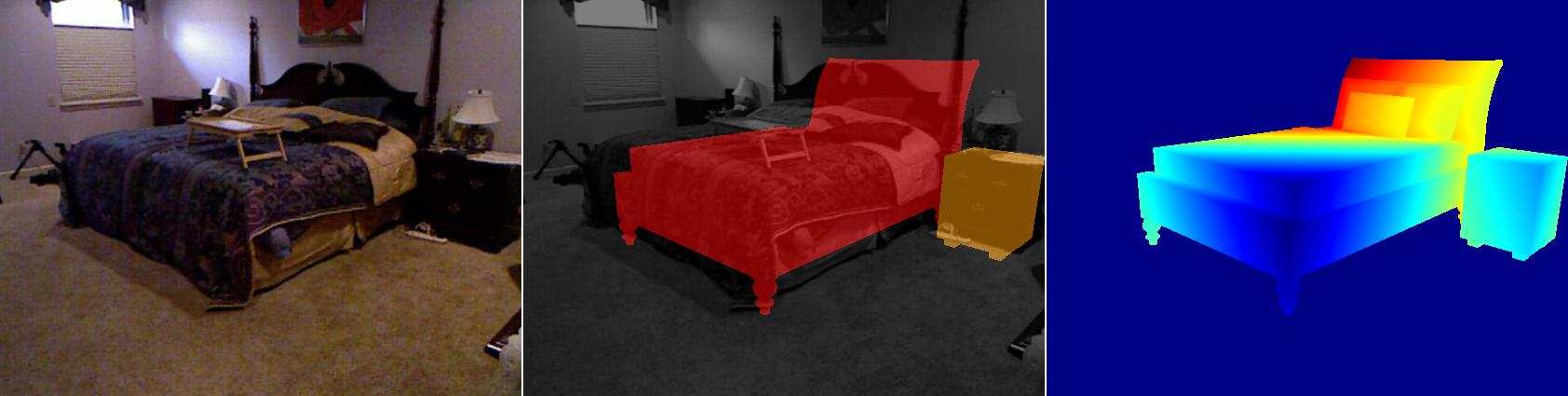} &
\insertA{0.330}{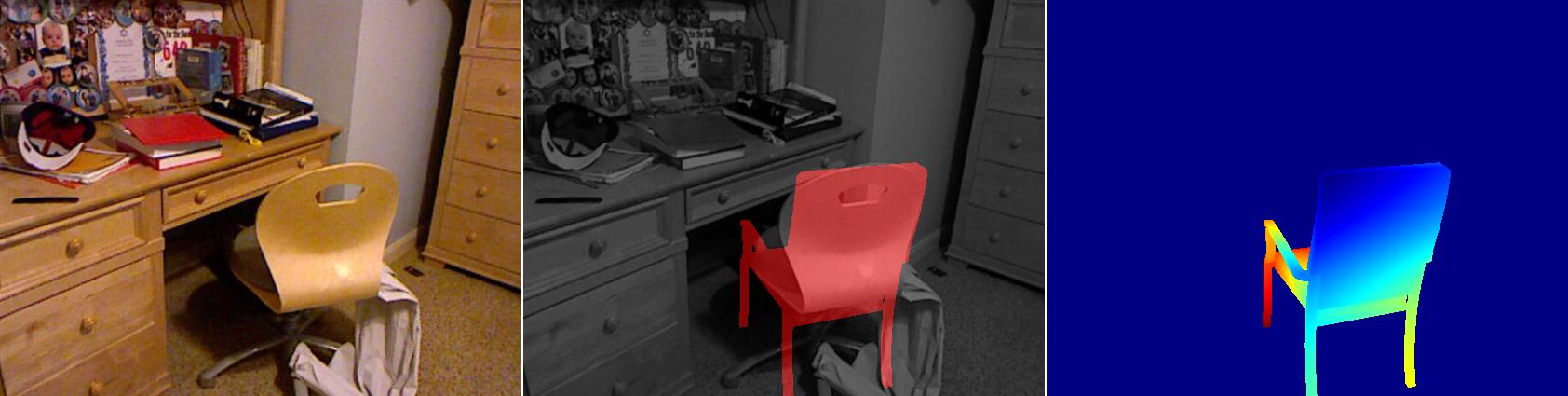} &
\insertA{0.330}{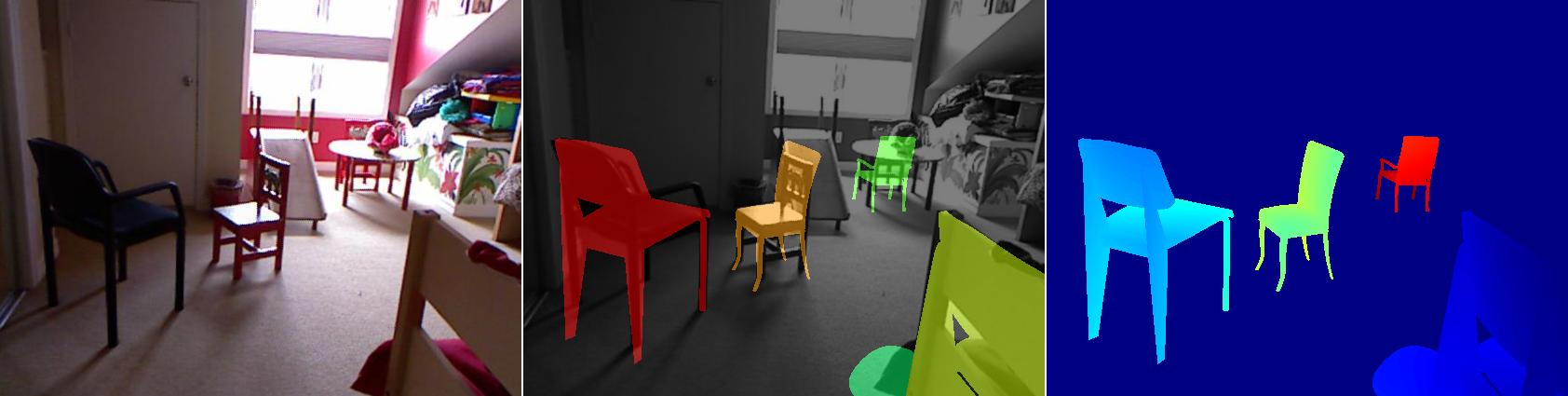} \\
\insertA{0.330}{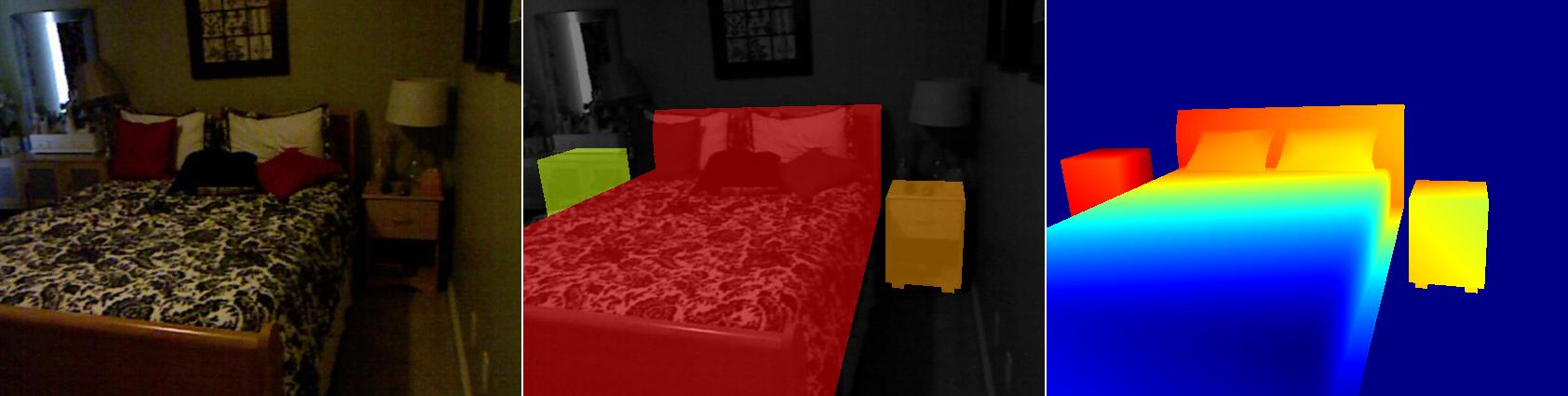} &
\insertA{0.330}{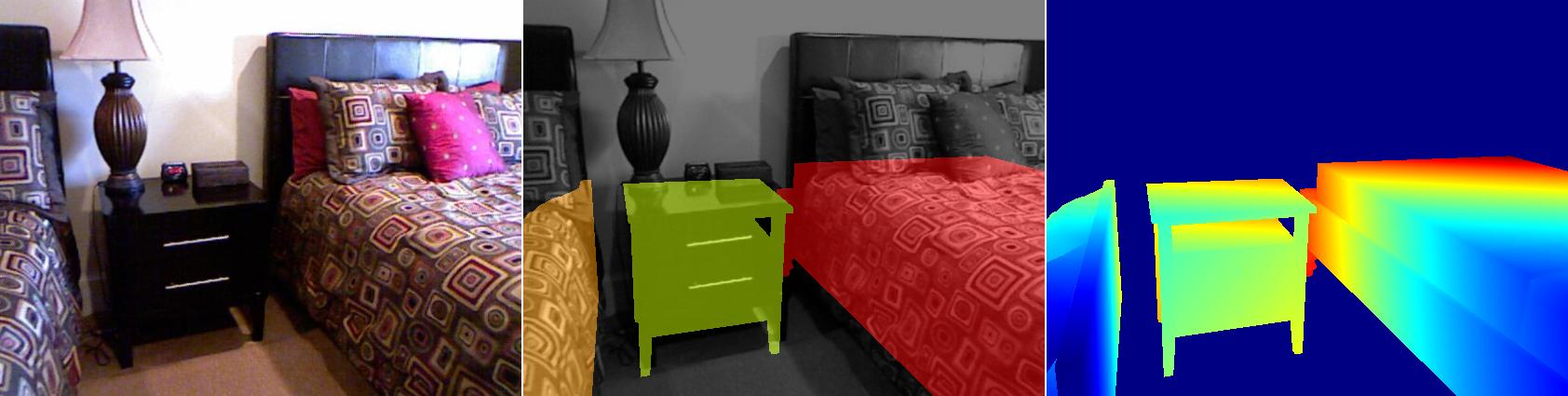} &
\insertA{0.330}{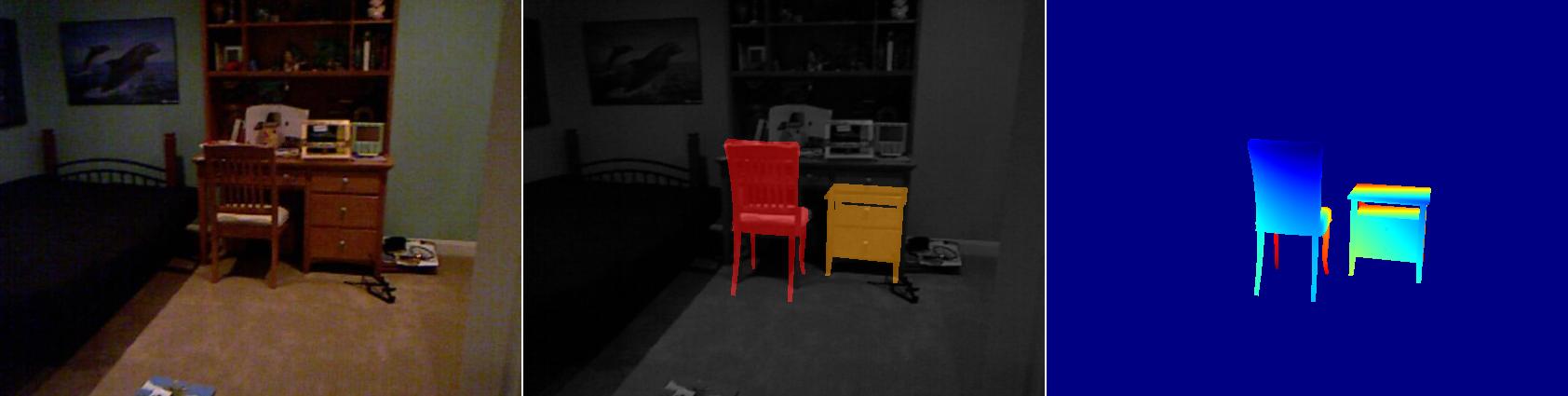} \\
\insertA{0.330}{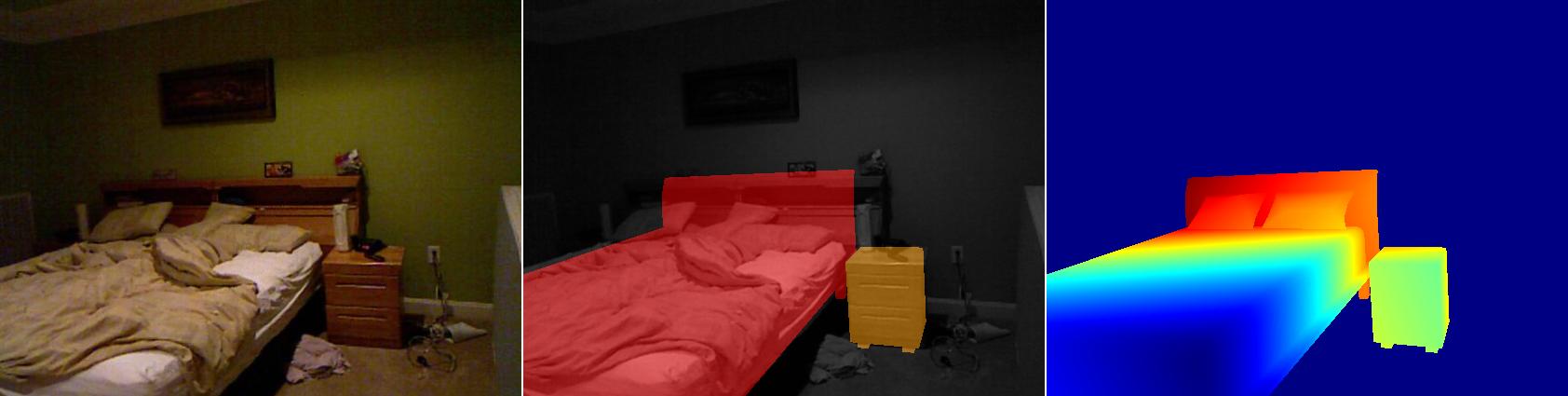} &
\insertA{0.330}{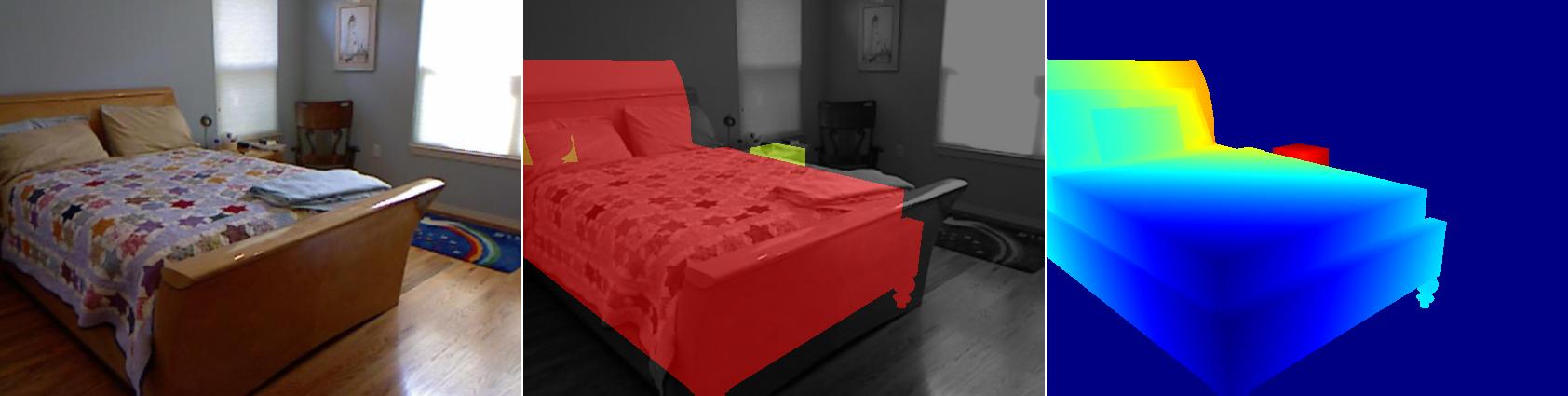} &
\insertA{0.330}{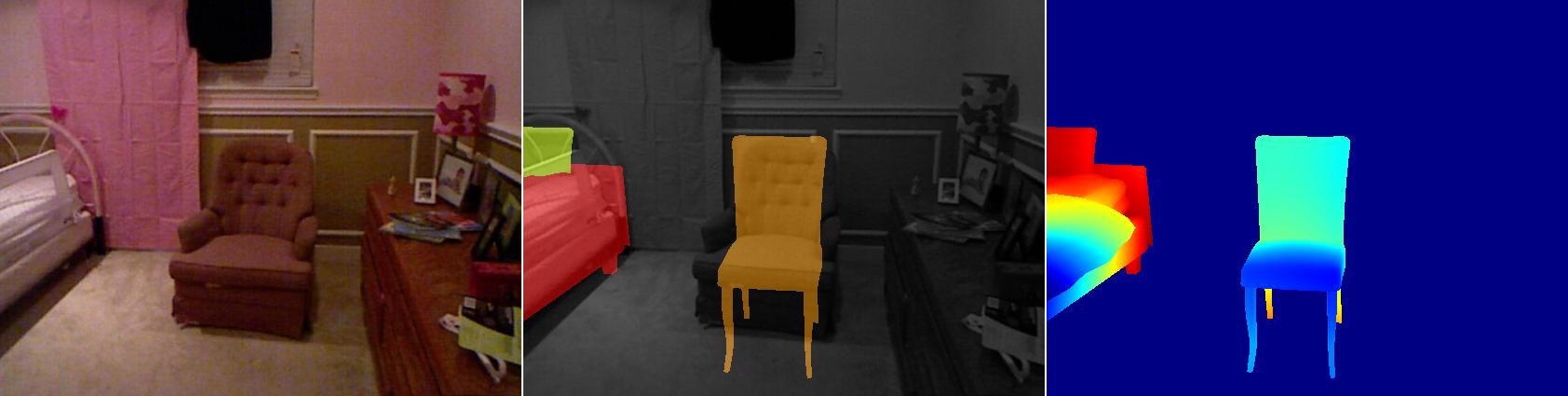} \\
\insertA{0.330}{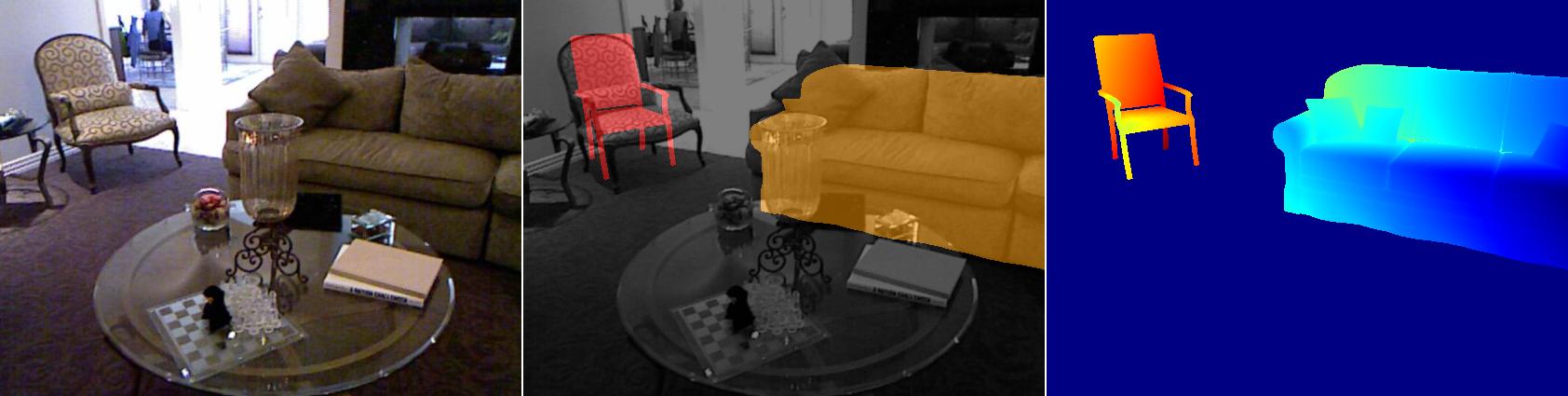} &
\insertA{0.330}{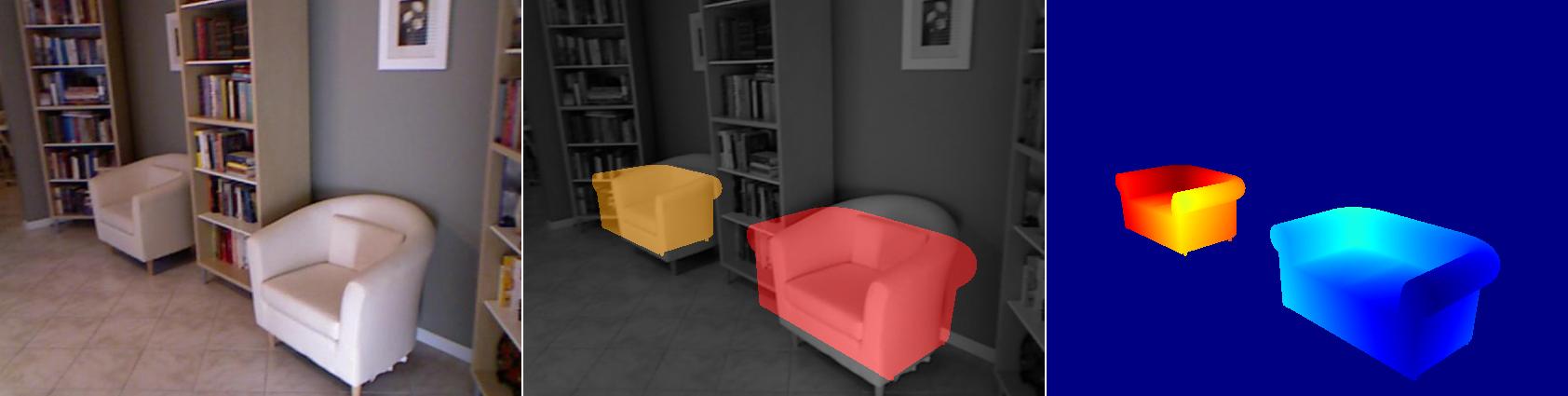} & 
\insertA{0.330}{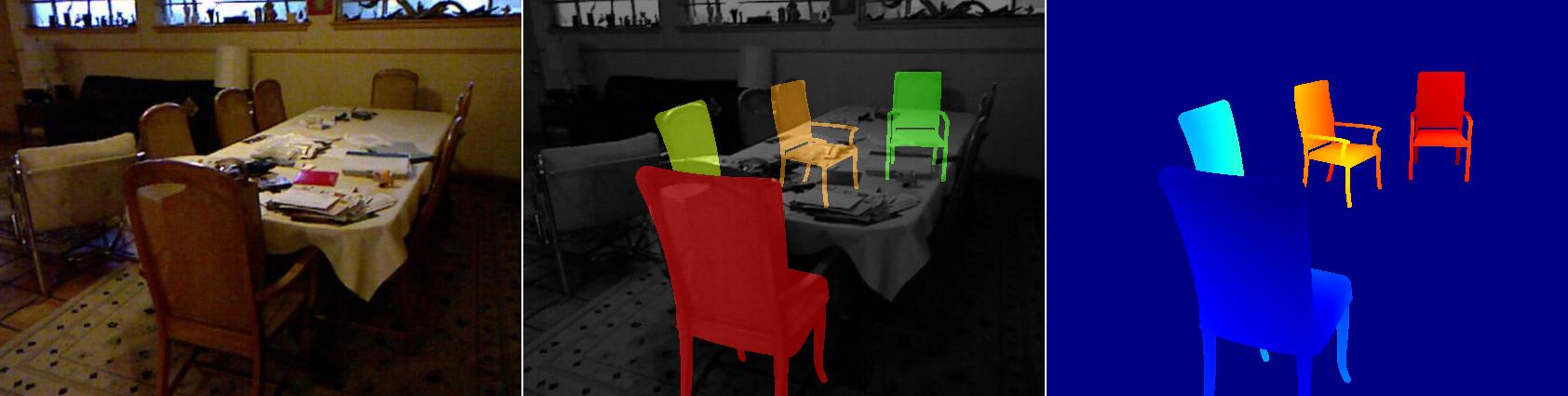} \\
\insertA{0.330}{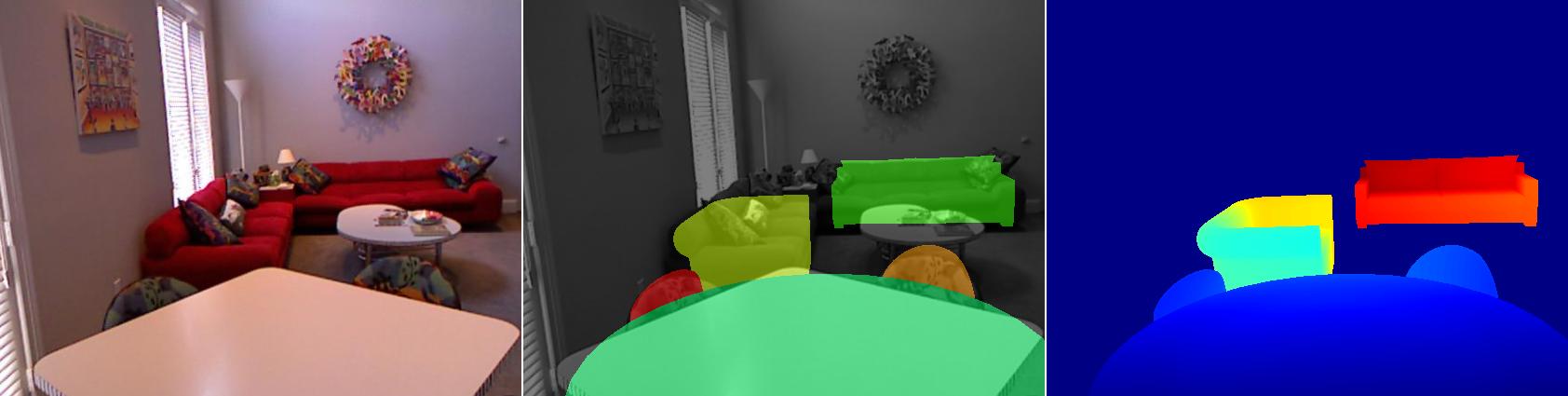} &
\insertA{0.330}{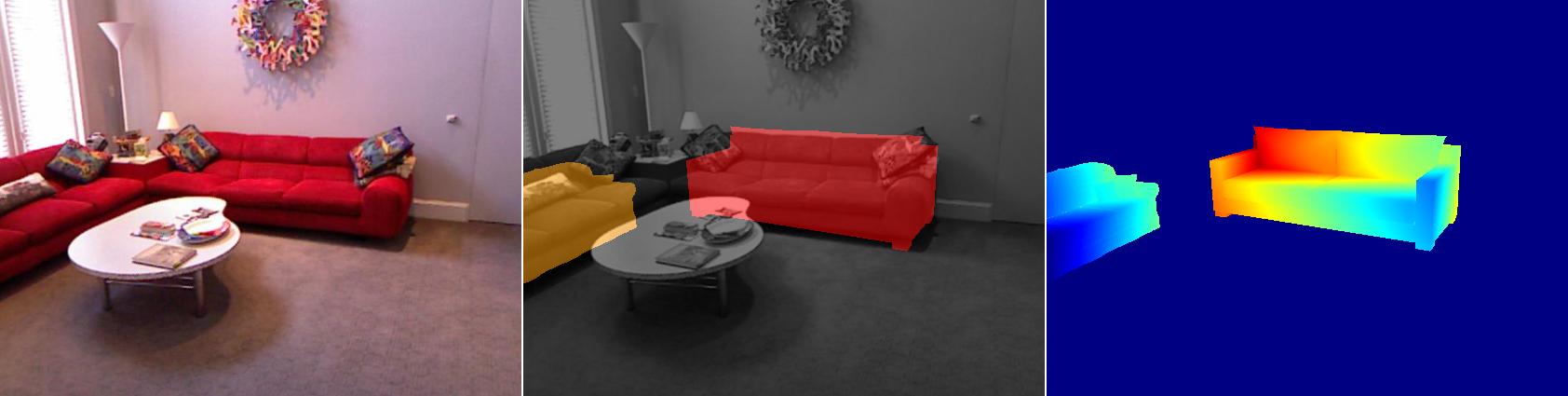} & 
\insertA{0.330}{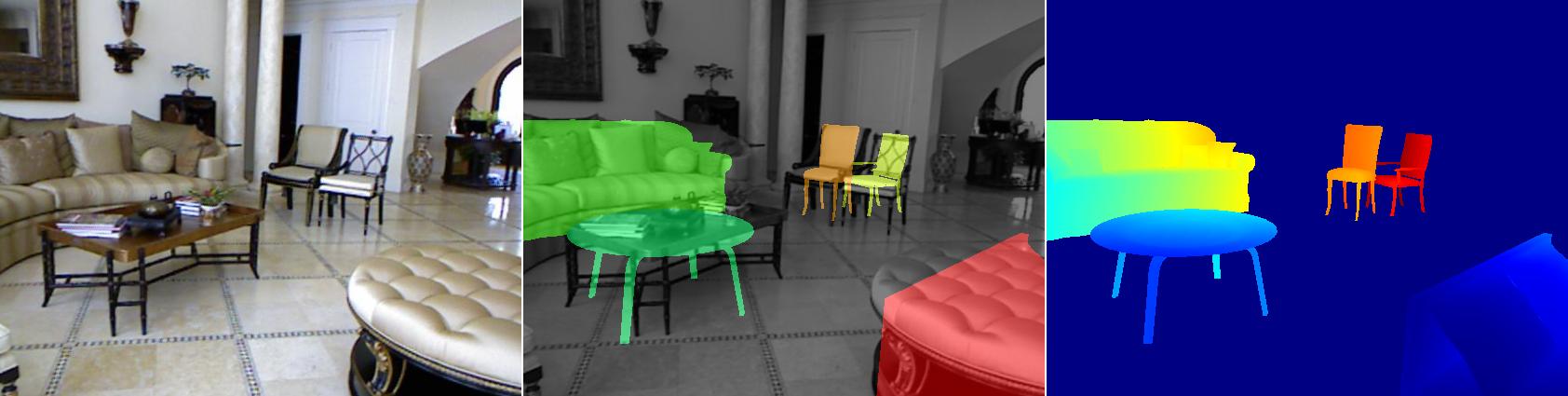} \\
\insertA{0.330}{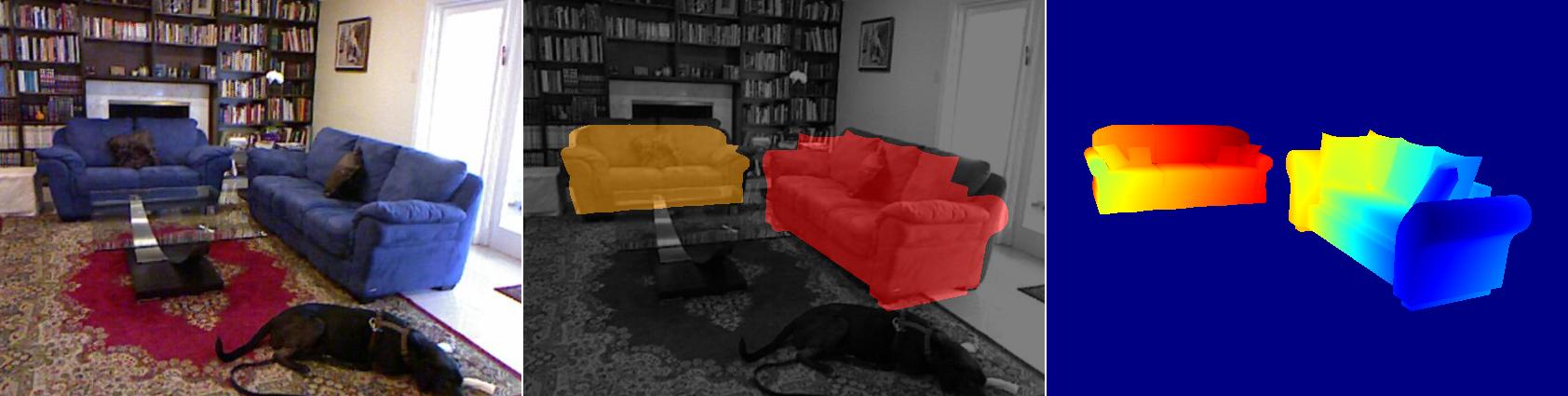} &
\insertA{0.330}{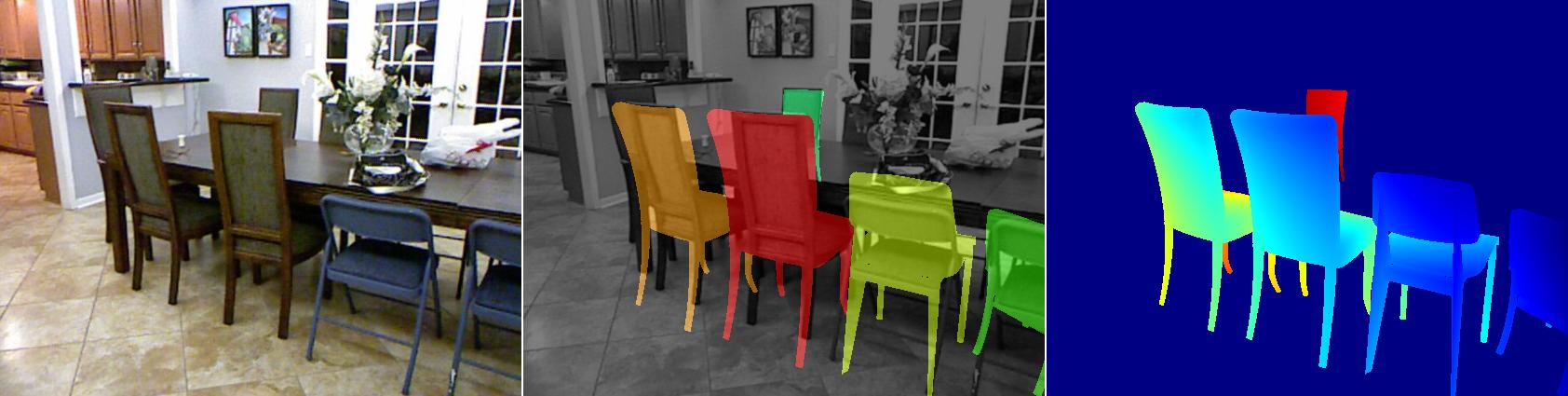} & 
\insertA{0.330}{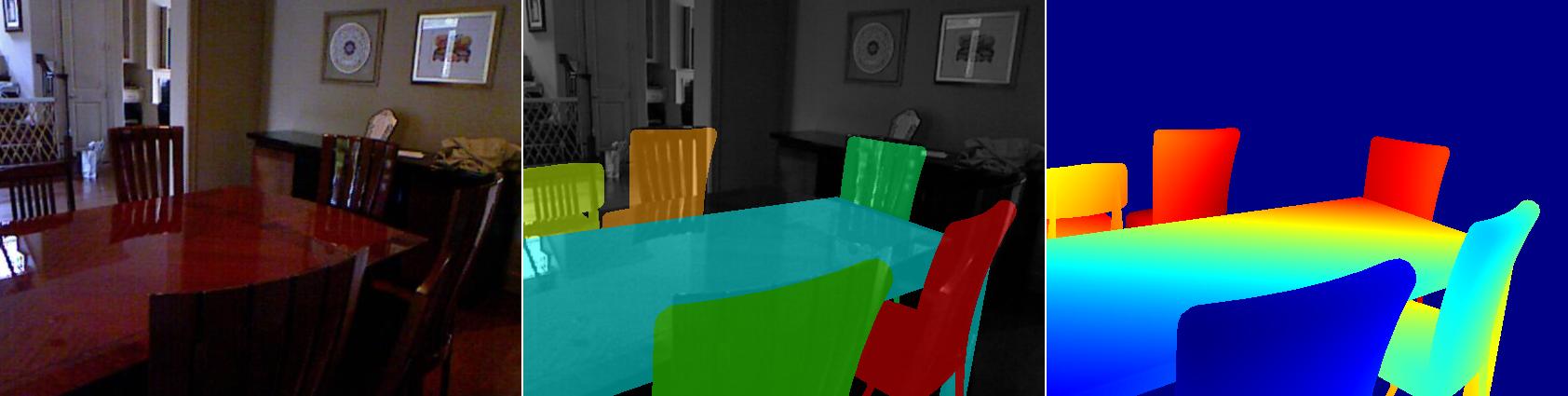} \\
\insertA{0.330}{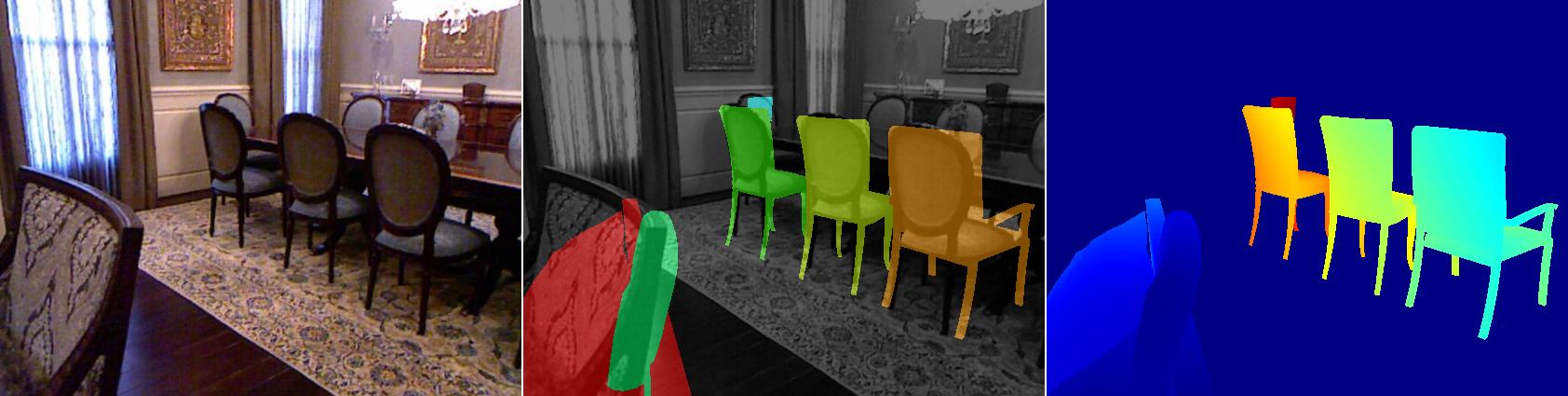} &
\insertA{0.330}{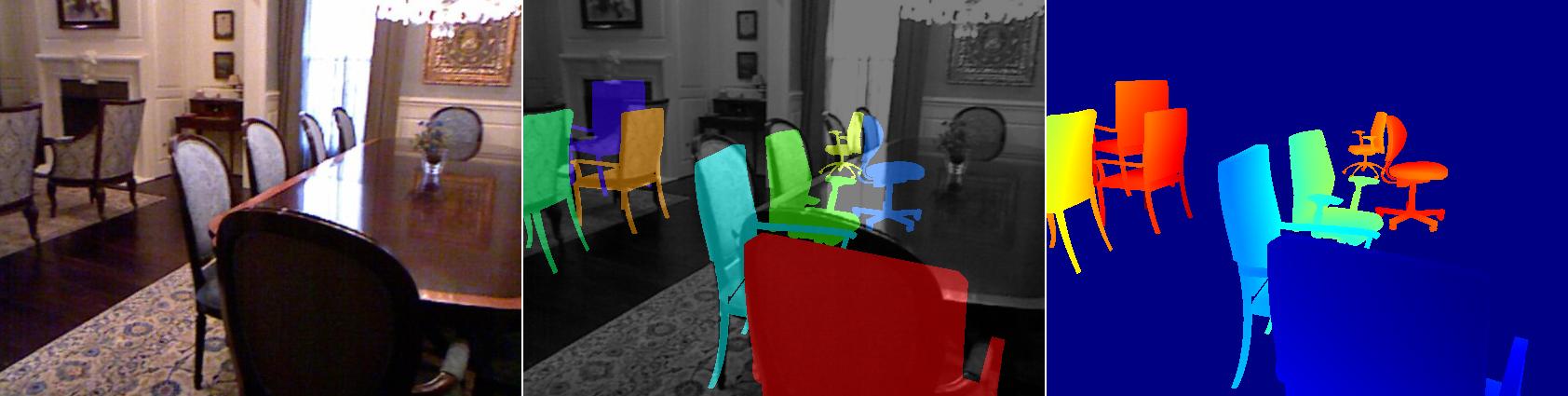} & 
\insertA{0.330}{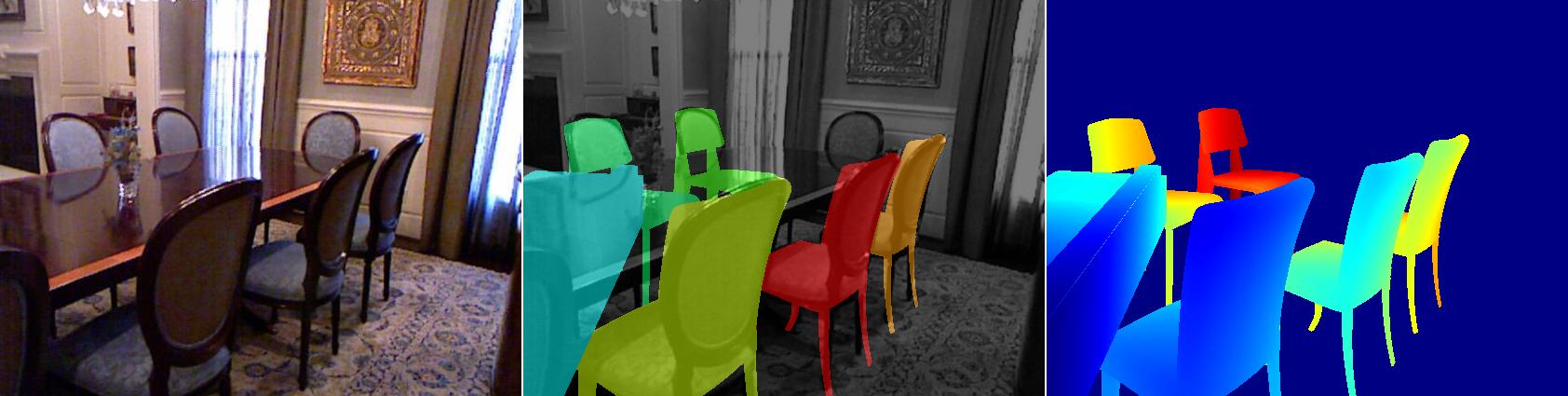} \\
\end{tabular}
\caption{\textbf{Visualizations of the output on the \textit{test} set}: We
show images with multiple objects replaced with corresponding 3D CAD models. We
show the image, models overlaid onto the image and the depth map for models
placed in the scene. Depth maps are visualized using the `jet' colormap, far
away points are red and and close by points are blue.}
\figlabel{output-vis-multiple}
\end{figure*}

\paragraph{Acknowledgements: }
The authors would like to thank Shubham Tulsiani for his valuable comments.
This work was supported by ONR SMARTS MURI N00014-09-1-1051, ONR MURI
N00014-10-10933, and a Berkeley Graduate Fellowship. We gratefully acknowledge
NVIDIA corporation for the donation of Tesla and Titan GPUs used for this research.

{\small
\bibliographystyle{ieee}
\bibliography{refs-rbg}
}
\end{document}